%% file: main.tex
\documentclass[10pt]{article} 
\usepackage[accepted]{tmlr}

\def\month{06}   
\def\year{2026}  
\def\openreview{\url{https://openreview.net/forum?id=3pfsQcEtNC}}

\usepackage[utf8]{inputenc} 
\usepackage[T1]{fontenc}    
\usepackage{hyperref}       
\usepackage{url}            
\usepackage{booktabs}       
\usepackage{amsfonts}       
\usepackage{nicefrac}       
\usepackage{microtype}      
\usepackage{xcolor}         

\usepackage{subcaption}
\usepackage{wrapfig}
\usepackage{booktabs}
\usepackage{graphicx}
\newcommand{\ours}[1]{\textsc{CreativityPrism}}
\usepackage{makecell}
\usepackage{hyperref}
\usepackage{booktabs}
\usepackage{multirow}
\usepackage{geometry}
\usepackage{siunitx}
\usepackage[dvipsnames]{xcolor}
\usepackage[table]{xcolor}
\usepackage{xspace}
\usepackage{soul}
\usepackage{amsmath} 
\usepackage{tabularx}
\usepackage[most]{tcolorbox}
\tcbset{mystyle/.style={
  colback=blue!5!white,
  colframe=white!90!black,
  fonttitle=\bfseries,
  coltitle=black,
  boxrule=0.5pt,
  arc=2mm,
  left=6pt,
  right=6pt,
  top=6pt,
  bottom=6pt,
  before skip=.5\baselineskip,
    after  skip=.5\baselineskip   
}}

\usepackage{arydshln}
\usepackage{float}
\usepackage{amssymb}
\usepackage[most]{tcolorbox}

\newcommand{\brainemoji}{\raisebox{-2pt}{\includegraphics[width=1em]{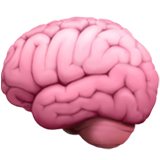}}}
\newcommand{\writingemoji}{\raisebox{-2pt}{\includegraphics[width=1em]{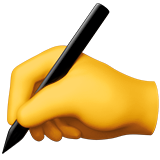}}}
\newcommand{\laptopemoji}{\raisebox{-2pt}{\includegraphics[width=1em]{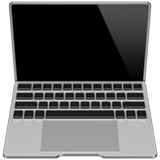}}}

\title{CreativityPrism: A Cross-Domain Evaluation \\ Framework for Large Language Model Creativity}

\input{authors}
\definecolor{mysalmon}{RGB}{250,205,207}
\definecolor{mygreen}{RGB}{203,255,207}
\definecolor{mypurple}{RGB}{203,210,255}

\definecolor{green-correct}{RGB}{38,127,65}
\definecolor{blue-llmoutput}{RGB}{59,111,231}
\definecolor{grey-typical}{RGB}{150,150,150}
\begin{document}

\maketitle
\input{sections/0-abstract}

\input{sections/1-intro}

\input{sections/2-related-work}
\input{sections/3-creativity-taxonomy}
\input{sections/4-experiments}

\input{sections/5-analysis}

\input{sections/6-conclusion}
\bibliographystyle{tmlr}
\bibliography{reference}

\input{sections/appendix}
\end{document}

%% file: authors.tex
\author{\name{Zhaoyi Joey Hou}\textsuperscript{1\textdagger},
    \name{Bowei Alvin Zhang}\textsuperscript{2},
    \name{Yining Lu}\textsuperscript{3},
    \name{Bhiman Kumar Baghel}\textsuperscript{1},
    \name{Anneliese Brei}\textsuperscript{4},
    \name{Ximing Lu}\textsuperscript{5},
    \name{Meng Jiang}\textsuperscript{3},
    \name{Faeze Brahman}\textsuperscript{6},
    \name{Snigdha Chaturvedi}\textsuperscript{4},
    \name{Haw-Shiuan Chang}\textsuperscript{7},
    \name{Daniel Khashabi}\textsuperscript{2},
    \name{Xiang Lorraine Li}\textsuperscript{1\textdagger}\\
    \addr{\textsuperscript{1}University of Pittsburgh,
        \textsuperscript{2}Johns Hopkins University,
        \textsuperscript{3}University of Notre Dame,\\
        \textsuperscript{4}University of North Carolina at Chapel Hill, 
        \textsuperscript{5}University of Washington,\\
        \textsuperscript{6}Allen Institute for Artificial Intelligence,  
        \textsuperscript{7}University of Massachusetts Amherst,\\
        \textsuperscript{\textdagger}Correspondence to {joey.hou@pitt.edu}, {xianglli@pitt.edu}}}

%% file: sections/0-abstract.tex
\begin{abstract}
Creativity is often seen as a hallmark of human intelligence. 
While large language models(LLMs) are increasingly perceived as generating creative text, there is still no \textit{cross-domain} and \textit{scalable} framework to evaluate their creativity across diverse scenarios. 
Existing methods of LLM creativity evaluation either heavily rely on humans, limiting speed and scalability, or are fragmented across different domains and different definitions of creativity. 
To address this gap, we propose \ours{}\footnote{Code and data: \url{https://joeyhou.github.io/CreativityPrism/}}, an evaluation and analysis framework that consolidates eight tasks from three domains: divergent thinking, creative writing, and logical reasoning, into a taxonomy of creativity that emphasizes three dimensions: quality, novelty, and diversity of LLM generations. The framework is designed to be scalable with reliable automatic evaluation judges that have been validated against human annotations.
We evaluate 17 state-of-the-art (SoTA) LLMs on \ours{} and find that while frontier-scale LLMs dominate creative writing and logical reasoning tasks by a .10 (or 15\%) lead over locally-deployable open models, they offer no significant advantage in divergent thinking, a domain much less explored in existing post-training regimes. Our analysis also shows that high performance in one creative dimension or domain rarely generalizes to others; specifically, novelty metrics often show weak or negative correlations with other metrics. This fragmentation confirms that a cross-domain, multi-dimensional framework like \ours{} is essential for any meaningful assessment of LLM creativity.
\end{abstract}

%% file: sections/1-intro.tex
\section{Introduction} 

Creativity, the capacity to generate novel and valuable ideas or solutions ~\citep{holyoak2005cambridge, Boden1994-uh, Finke1992-ke}, is a core human cognitive ability. It appears in many domains: crafting stories with surprising plot twists ~\citep{Ismayilzada2024-ug, cs4}, producing groundbreaking scientific discoveries~\citep{Hu2002-nn, llm_for_research_idea}, solving problems under constraints ~\citep{neocoder, assessing_creativity_llms_math}, or even expressing humor in everyday life ~\citep{He2019-mq, think_outside_of_the_box}. Its multifaceted nature has prompted extensive study in psychology and cognitive science, with efforts to capture creativity through both qualitative and quantitative approaches ~\citep{Guilford2012-kh, Olson2021-xq, Alabbasi2022-jz, Sternberg1991-qe}. 

Recently, with the rapid rise of general-purpose large language models(LLMs), interest has grown in probing their creativity ~\citep{assessing_understanding_creativity_llm, push_gpt_creativity_aut, Chakrabarty2024-fd, neocoder, cs4}. But as with human creativity, machine creativity spans diverse and expansive contexts, making it difficult to define, formalize, and, above all, measure.
Concretely, LLM creativity evaluation faces two challenges: \textbf{distinct definitions of creativity} across different domains and difficulty of \textbf{scalable, automatic evaluation} due to the convoluted nature of creativity. 
 
The first challenge stems from current research in machine creativity being scattered across different domains and focusing on narrow or singular dimensions. For example, the Divergent Association Task(DAT) ~\citep{Chen2023-tb, Bellemare-Pepin2024-yc} and the Creative Short Story Task ~\citep{Ismayilzada2024-ug} emphasize lexical diversity; the Alternative Uses Test(AUT) ~\citep{push_gpt_creativity_aut, Organisciak2023-pb} solely focuses on unconventional ideas of using daily items, overlooking the pragmatics of those solutions; CreativeMath ~\citep{assessing_creativity_llms_math} and NeoCoder ~\citep{neocoder} only study math and coding problems, correspondingly. What makes comparison even harder is that these task-specific and domain-specific benchmarks only benchmark their own choices of LLMs, which vary from one another. Without a cross-domain evaluation that incorporates the evaluation of creativity from all those dimensions and covers a wide range of LLMs, it is hard to uncover a full picture of how well current state-of-the-art(SoTA) LLMs are doing when it comes to creativity. 

The second challenge arises from the subjective nature of creativity, which makes it hard to automatically evaluate LLM output and leads many existing benchmarks to rely heavily on human evaluation \citep{macgyver, Chakrabarty2024-fd}. While human judgment is often considered the gold standard for nuanced tasks, it presents significant hurdles for modern AI research: it is prohibitively expensive, difficult to replicate at scale, and requires a long turnaround time that cannot keep pace with the current field. With new LLMs and model iterations being released nearly every week, a reliance on manual grading creates a massive bottleneck that prevents the rapid, iterative testing required for progress. 

\begin{figure}[t!]
\centering
    \includegraphics[width=0.85\linewidth]{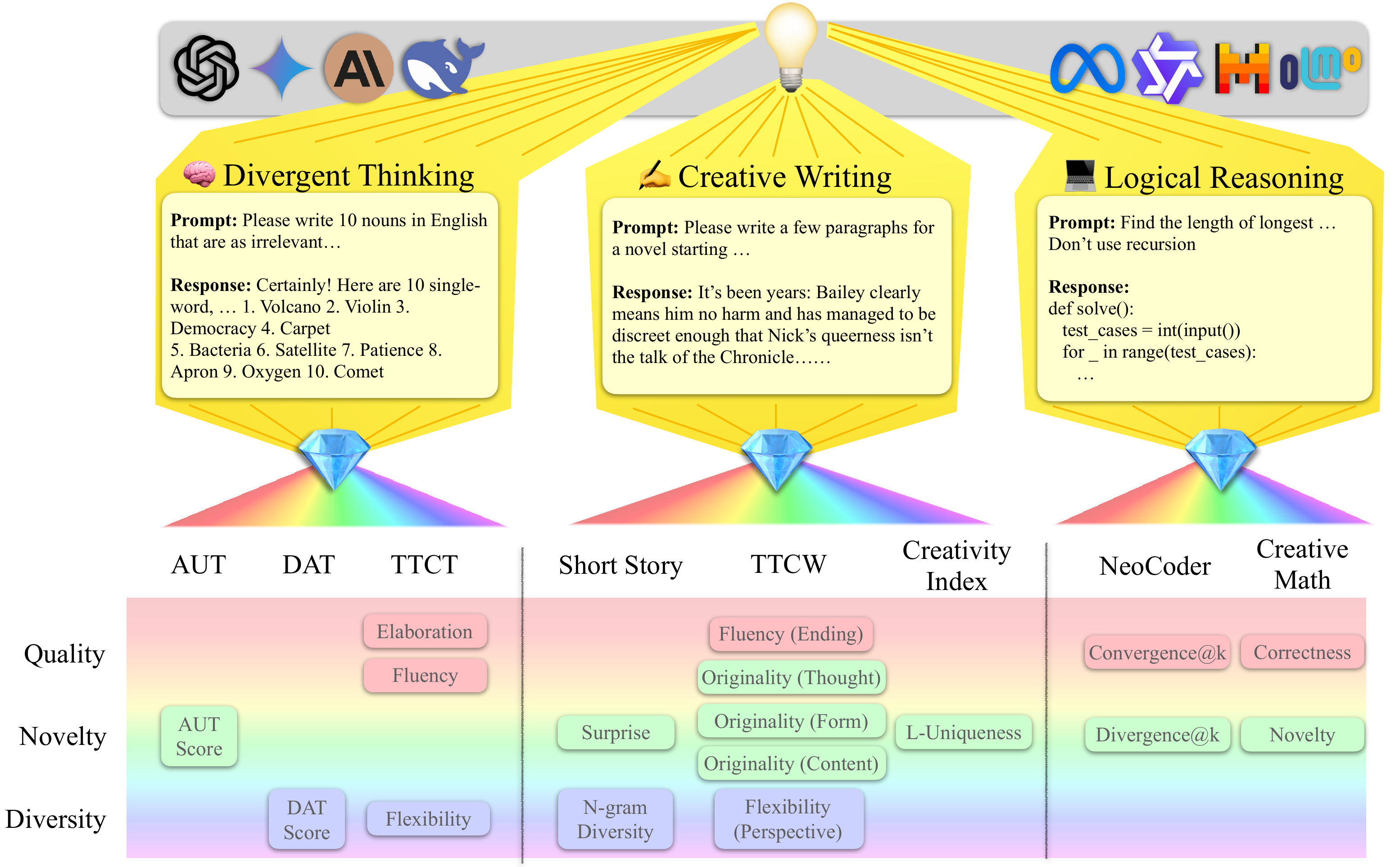}
    \caption{
    Overview of \ours{}. Each LLM is prompted to complete the tasks (example inputs in Table \ref{tab:task-descriptions}), and their outputs are evaluated using task-specific metrics. We also organize metrics into three dimensions of creativity:
    \textbf{\colorbox{mysalmon}{quality}}, \textbf{\colorbox{mygreen}{novelty}}, and \textbf{\colorbox{mypurple}{diversity}}. 
    }
    \vspace{-2em}
    \label{fig:main_fig}
\end{figure}

To this end, we propose \ours{}, a cross-domain and scalable evaluation framework of LLM creativity; it is cross-domain \textemdash{} consisting of eight tasks from three domains: divergent thinking, creative writing, and logical reasoning (i.e., mathematical reasoning and coding); it is also scalable \textemdash{} all evaluation metrics are automatic, ensuring easy benchmarking for any LLM. 
First, cross-domain: we extend beyond a simple combination of existing benchmarks from various domains by systematically categorizing existing task-specific metrics along the three dimensions, i.e., quality, novelty, and diversity, to facilitate a full-scale measurement of model creativity (Figure~\ref{fig:main_fig}). 
Every metric in \ours{} belongs to one of these three dimensions, and hence, evaluation results can be summarized into those three dimensions, providing a cross-domain and dimension-specific insight into LLMs' creativity performance. 
Second, scalability: we go beyond simple automatic evaluation with the LLM-as-a-Judge paradigm (\S \ref{sec-exp-llm-judge}). For every evaluation metric that requires LLM-as-a-Judge, we collect human judgments from well-trained researchers or domain experts, followed by Alternative Annotator Test ~\citep{alternative-annotator-test} to verify our LLM-Judge setup has a much closer alignment to high-quality human judgment among tasks in \ours{} than previous automatic evaluations. 
In this way, we ensure the metrics in \ours{} are well aligned with high-quality human judgment.
 
\setlength{\fboxsep}{0pt}

We then evaluate 17 SoTA LLMs on \ours{}, to answer the following research questions. \textbf{RQ1: Where is the performance gap across LLMs on creative tasks?} We found a notable performance gap between frontier-scale and locally-deployable open models, especially in logical reasoning tasks, followed by creative writing tasks. We also found the same performance gap to be most pronounced in the quality dimension. \textbf{RQ2: What are the correlations among LLMs' performance across different dimensions and domains?} In other words, would good performance in one creativity domain or dimension transfer to the other?  Results have shown that the models perform similarly in metrics from the same task or the same domain; for metrics from different domains, models perform similarly in diversity and quality dimensions, while performances in novelty dimensions are much less correlated. We believe this stems from the inherent difference in how novelty is defined in different tasks and domains. 

In short, our contribution can be summarized as follows:
\vspace{-1em}
\begin{itemize}
    \itemsep0em 
    \item We select, combine, and refactor the codebase of eight creativity tasks across three domains to propose an easy-to-use, cross-domain and scalable evaluation framework, \ours{};
    \item We propose the ``quality, novelty, and diversity'' taxonomy of creativity and categorize seventeen metrics into these dimensions for dimension-level LLM creativity evaluation;
    \item We conduct high-quality human annotation to verify human agreement for each metric and human-LLM-Judge agreement during the automatic evaluation process;
    \item We benchmark 17 SoTA LLMs on \ours{} and conduct extensive analysis on model performance across domains and dimensions.
\end{itemize}

%% file: sections/2-related-work.tex
\section{Related Work}
\paragraph{Human Creativity}
The definition of creativity has varied across different domains. In psychology, Torrance Tests of Creative Thinking (TTCT) \citep{Alabbasi2022-jz} considers creativity as a combination of originality, flexibility, fluency, and elaboration; \citet{runco2012standard} proposes a simpler taxonomy of originality and usefulness. In marketing, \citet{El-Murad2004-ue, Rosengren2020-ij} considers advertisement creativity as the combination of usefulness and originality; additionally, \citet{Smith2007-zs} adds flexibility, fluency, elaboration, synthesis, and artistic values. In terms of creativity evaluation, \citet{Said-Metwaly2017-eq} summarizes more than 100 existing works and concludes that creativity evaluation is an ``unsettled'' issue, with one of the key reasons being the lack of holistic, cross-domain evaluation, which also motivates our work. A common belief among these work is the balanced view of quality (e.g., elaboration, usefulness), novelty (e.g., originality, synthesis), and diversity (e.g., flexibility) when it comes to different dimensions of creativity. This view directly inspires our taxonomy of creativity metrics in \S \ref{sec-creativity-taxonomy}.

\vspace{-2mm}
\paragraph{Machine Creativity} 
Measurement of machine creativity has become increasingly popular with the rapid development of LLMs. Many researchers attempt to provide a cross-domain or cross-dimension view of machine creativity in which a variety of angles have been taken, such as panoramic survey \citep{creativity_ai_progresses_challenges}, computational models involved in machine creativity \citep{creativity_and_machine_learning_survey}, the efficacy of evaluation methods and metrics \citep{Lu2025-sn, He2025-ij}, the effect of LLM decoding strategy on output creativity \citep{nagarajan2025roll}, the tension between factuality and creativity \citep{banerjee2025does}, and multi-modal creativity \citep{Fang2025-im, xue2025crebench}. However, none of them attempt to validate, synthesize, and combine the evaluation metrics from a series of creativity tasks and benchmark existing SoTA LLMs as we do.

The community has also explored a wide range of domain-specific problems when it comes to LLM creativity, including logical-based problem-solving \citep{assessing_creativity_llms_math, neocoder, chen2025deepmath}, physical and commonsense reasoning \citep{Tian2024-py}, creative writing \citep{Gomez-Rodriguez2023-xp, Lu2024-ma, Chakrabarty2024-fd, Ismayilzada2024-ug, Tian2024-at, cs4, qiu-hu-2025-deep}, scientific discovery \citep{llm_for_research_idea, kumar-etal-2025-large, afzal2025beyond}, response diversity in question answering \citep{Yiming2025-xs, McLaughlin2024-jb}, and human-AI collaborative creative problem solving \citep{Chakrabarty2024-lj, boussioux2024crowdless, N-Lane2024-hp}. All these studies focus on LLM evaluation in a specific domain, each with its own evaluation philosophy and metrics, whereas our work aims at providing a cross-domain and cross-dimension evaluation of the LLM's output for tasks in a variety of domains.

\vspace{-2mm}

\paragraph{Automatic Creativity Evaluation}
Evaluating the creativity of machine-generated text has been a challenging task, and much of the work relies on human evaluation. However, due to the cost, human evaluation is challenging to scale and requires a lengthy wait time. To achieve scalable evaluation, researchers adopt two broad groups of evaluation methods: feature-based and generative-based. The former includes psycholinguistic features, such as arousal, valence score \citep{Mohammad2018-gc}, lexical features, such as lexical diversity \citep{Padmakumar2023-ry}, and text embedding distances \citep{Pennington2014-ea, Zhang2019-ab}. The latter is mainly LLM-as-a-judge \citep{judgebench, llm_as_a_judge_opportunities_challenges, llms_as_judges_comprehensive_survey_llm_based}. Recent work has shown promising potential of this method \citep{Zheng2023-pm} as well as quantitative ways to measure how well LLM-Judge aligns with human judgment \citep{alternative-annotator-test, han2025judge}. 
Our evaluation framework integrates both feature-based and generative evaluators. For metrics that rely on LLM judges, we systematically validate their output against human annotations using the Alternative Annotator Test (see \S\ref{sec:exp}).

%% file: sections/3-creativity-taxonomy.tex
\begin{table}[h]
    \centering
    \small
        \begin{tabularx}{\linewidth}{p{0.38\linewidth} X}
        \toprule
        \textbf{Task} & \textbf{Example}\\
        \midrule
        \brainemoji \hspace{0.1cm}\textbf{Alternative Uses Test (AUT)} \citep{push_gpt_creativity_aut}  & Create a list of creative alternative uses for a bottle. \\
        \brainemoji \hspace{0.1cm} \textbf{Divergent Association Task (DAT)} \citep{Chen2023-tb, Bellemare-Pepin2024-yc}  &  Please write 10 nouns in English that are as irrelevant from each other as possible, in all meanings and uses of the words.\footnotemark\\
        \brainemoji \hspace{0.1cm} \textbf{Torrance Tests of Creative Thinking (TTCT)} \citep{assessing_understanding_creativity_llm} & What might be the consequences if humans suddenly lost the ability to sleep?\\
        \midrule
        \writingemoji \hspace{0.1cm}\textbf{Torrance Test of Creative Writing (TTCW)} \citep{Chakrabarty2024-fd} & Write a New Yorker-style story given the plot below. Make sure it is at least 2000 words. Plot: A woman experiences a disorienting night in a maternity ward...; Story:  \\
        \writingemoji \hspace{0.1cm}\textbf{Creative Short Story} \citep{Ismayilzada2024-ug} & Come up with a novel and unique story that uses the required words in unconventional ways or settings. Use at most five sentences. The given words: petrol, diesel, and pump.  \\
        \writingemoji \hspace{0.1cm}\textbf{Creativity Index} \citep{Lu2024-ma} & Please write a few paragraphs for a novel starting with the following prompt: ``It’s been years: Bailey clearly means him no harm and has managed to...''  \\        
        \midrule
        \laptopemoji \hspace{0.1cm} \textbf{NeoCoder} \citep{neocoder} \newline & You are given a sequence of integers $a$ of length $2n$. You have to split them into $n$ pairs. Don’t use HashMap, while loop. \\
        \laptopemoji \hspace{0.1cm} \textbf{Creative Math} \citep{assessing_creativity_llms_math} & Question: What is the largest power of 2 that is a divisor of $13^4 - 11^4$? A.8 B.16 C.32 D.64 E.128; Reference Solutions 1: ... ; Reference Solutions 2: ...\\
        \bottomrule
        \end{tabularx} 
    \caption{Tasks in \ours{} with examples. \brainemoji: divergent thinking, \writingemoji: creative writing, \laptopemoji: logical reasoning. More details and examples can be found in Appendix \ref{app-task-details}.}
    \vspace{-1em}
    \label{tab:task-descriptions}
\end{table}
\footnotetext{Different versions of DAT prompts exist across psychology and computer science research; the exact prompt is in Appendix \ref{app-dat-details}.}

\section{\ours{}}
\label{sec-creativity-taxonomy}
\paragraph{Task Selection}
In designing \ours{}, we make sure it is both cross-domain and scalable. These two requirements are reflected in task selection: we examine all creativity-related tasks with publicly available data and executable, well-documented codebases; we also further ensure the reliability of tasks that require LLM-Judge by conducting additional human annotation with trained researchers and LLM-Judge quality test to filter out tasks where no well-aligned LLM-Judge is available (\S \ref{sec-exp-llm-judge}). Our task selection process leads to eight tasks and seventeen metrics, across three domains: \brainemoji~divergent thinking, \writingemoji~creative writing, and \laptopemoji~logical reasoning (Figure \ref{fig:main_fig}).
The divergent thinking domain consists of established psychology tasks, which were originally designed to assess human ability in coming up with diverse and alternative answers to given questions~\citep{push_gpt_creativity_aut, Chen2023-tb, Bellemare-Pepin2024-yc, assessing_understanding_creativity_llm}.
The creative writing domain includes tasks that require models to produce short written pieces ~\citep{Chakrabarty2024-fd, Ismayilzada2024-ug, Lu2024-ma}.
The logical reasoning domain includes coding ~\citep{neocoder} and math task ~\citep{assessing_creativity_llms_math} to evaluate models' ability to generate creative solutions under strict, explicit reasoning constraints.
Task format examples are in Table \ref{tab:task-descriptions}.

\begin{wrapfigure}{l}{0.4\textwidth}
    \includegraphics[width=1\linewidth]{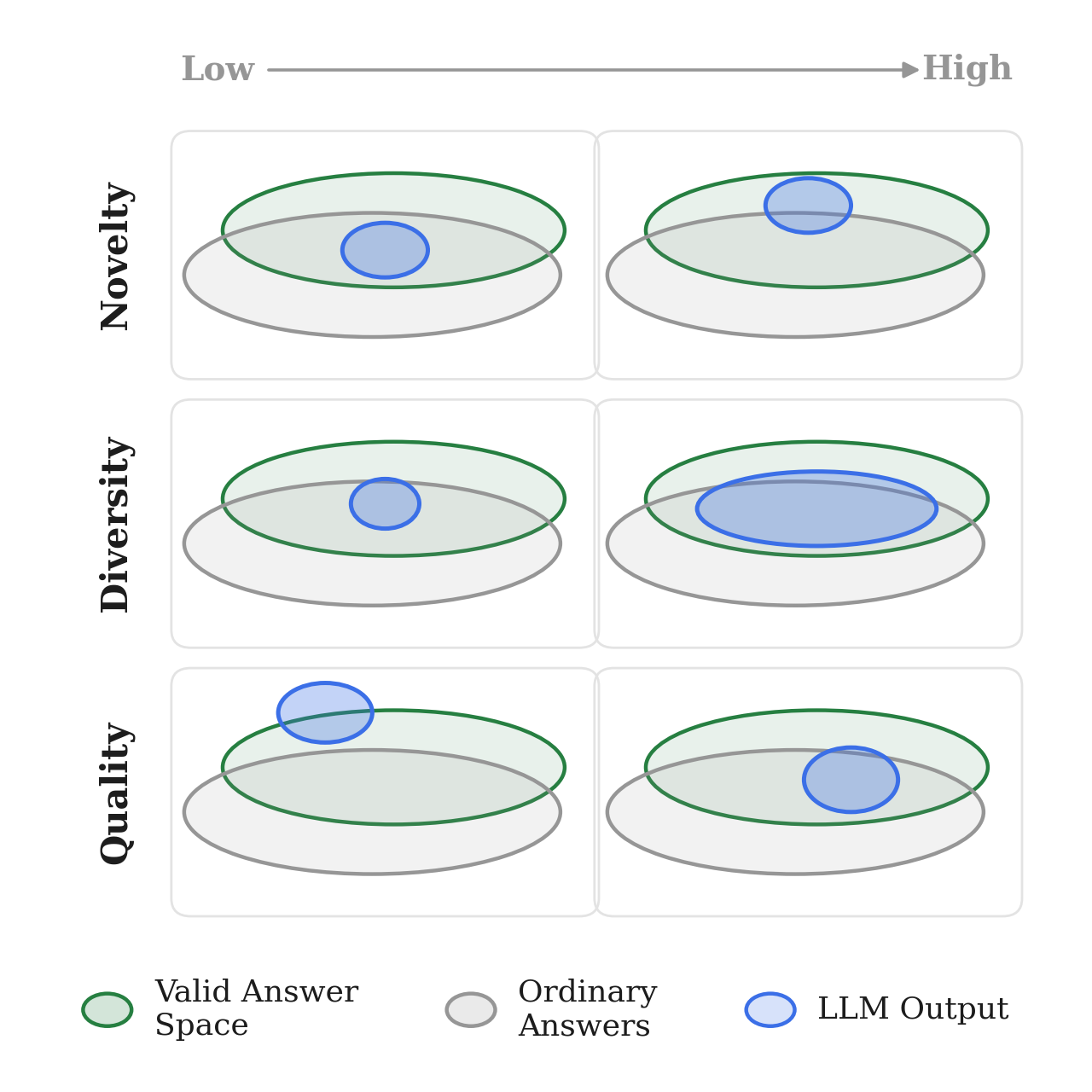} 
    \caption{Visual representation of three dimensions in \ours{}.}
    \label{fig:3-dim-vis}
    \vspace{-2em}
\end{wrapfigure}

\paragraph{Three Dimensions of Creativity}
We go beyond a simple combination of existing tasks by categorizing all metrics into three dimensions of creativity: \textit{quality}, \textit{novelty}, and \textit{diversity}. 
Our taxonomy is grounded in the most recent version of the TTCT verbal test, which operationalizes creativity along three dimensions: fluency, flexibility, and originality.\footnote{While the original TTCT \citep{torrance1974torrance} includes a fourth dimension, elaboration, the current version of the TTCT verbal test, as administered by major testing agencies \citep{ttct-current, ttct-training} consists of three dimensions only, a simplification also reflected in recent psychology literature \citep{Alabbasi2022-jz}.} These map directly onto our dimensions: fluency, the ability to produce coherent, well-formed responses, corresponds to \textit{quality}; flexibility, the ability to vary approaches and perspectives, corresponds to \textit{diversity}; and originality, the ability to produce responses that deviate from the commonplace, corresponds to \textit{novelty}\citep{ttct-current}. We are also different from the widely-cited binary taxonomy of usefulness and originality \citep{runco2012standard} by separating originality into novelty and diversity. We argue this distinction is particularly important for LLM evaluation: diversity captures breadth, i.e., how much a model varies its outputs across responses, while novelty captures depth, i.e., how much a single output deviates from existing or conventional solutions (Figure \ref{fig:3-dim-vis}). A model can score highly on one while failing on the other, e.g., producing many distinct but individually unoriginal responses (high diversity), or producing one highly novel response with no variation across prompts (high novelty). Combining diversity and novelty into one originality score would obscure such differences.

Besides theoretical groundings, we also refer to a visualized framing that motivates our choice of dimensions (Figure \ref{fig:3-dim-vis}). Consider three spaces: LLM's output distribution (\textcolor{blue-llmoutput}{\textbf{blue}}), valid answer space (\textcolor{green-correct}{\textbf{green}}), and ordinary answer space (\textcolor{grey-typical}{\textbf{grey}}). Each dimension captures a distinct geometric relationship among these three spaces.

\textit{\textbf{Quality}} measures how much of the LLM's output distribution falls within the valid answer space, i.e., how well the generated content fulfills the task's functionality. For example, in NeoCoder, the quality of generated code is measured by the success of execution and coding task completion; in TTCT, quality is measured by the amount of detail provided. We believe quality is critical for creativity because, without quality, random words and sentences would be very creative in terms of novelty and diversity, but do not convey any meaning.

\textit{\textbf{Novelty}} measures how far the LLM's output distribution extends beyond the ordinary answers, i.e., how rare the generated content is compared to existing or commonly seen content. For example, in NeoCoder and Creative Math, novelty involves coming up with solutions that are different from the reference solutions; in AUT, novelty involves different use of the tool compared to ordinary uses; in Creativity Index, novelty is measured by the normalized n-gram overlaps between model-generated text and the traceable part of the training corpus.

\textit{\textbf{Diversity}} measures the breadth of the LLM's output distribution within the valid answer space, i.e., how widely the model explores different valid responses across generations. For example, in TTCT, diversity is reflected in the flexibility score - the ability to produce varied solutions to given questions; for the DAT task, diversity involves the semantic difference among the LLM-generated nouns; similarly in \citet{Li2025-nv}, diversity is the number of semantically similar clusters an LLM generates in response to a given prompt.

These three relationships are geometrically independent: a model can expand its output distribution along one axis without moving along the others. This independence is what motivates treating them as separate dimensions rather than collapsing them.

\textbf{Different Domain Tasks within Same Dimension} As shown in Figure \ref{fig:main_fig}, there are tasks from different domains within the same dimension, e.g., ``N-gram Diversity'' from Creative Writing and ``DAT Score'' from Divergent Thinking are both in the \textit{Diversity} dimension. We make such choice because we believe \textit{Diversity} has different definition across different domains and so do other dimensions. Having heterogeneous metrics within one dimension is complementary to each other, making our evaluation framework cover a wider range of perspectives on creativity.

%% file: sections/4-experiments.tex
\section{Experiments}
\label{sec:exp}
To holistically evaluate machine creativity, we evaluated 17 SoTA LLMs across eight tasks, reporting both task-specific metrics and an aggregated creativity score using the three-dimensional framework.
In this section, we introduce the inference setups, where LLMs are prompted to generate creative responses according to corresponding task requirements (\S\ref{sec-exp-inference}); then we describe the evaluation process, including score aggregation and how we use LLM-as-a-Judge for evaluation (\S\ref{sec-exp-llm-judge}).

\subsection{Inference}
\label{sec-exp-inference}
All tasks in \ours{} undergo data processing according to the original task setup before running inference.
We include 17 LLMs in total, including frontier-scale LLMs \footnote{We use frontier-scale to refer to LLMs that are in practice only accessible via hosted API and not locally fine-tunable on research hardware. This includes both closed-weight LLMs (GPT, Claude, Gemini) and large open-weight LLMs from DeepSeek (DeepSeek-V3/R1); we group them together because they are comparable in practical accessibility despite being different in license. We contrast these with locally-deployable open models ($<$80B).} from GPT \citep{openai2024gpt4}, Claude \citep{anthropic2024claude3}, Gemini \citep{google2024gemini2, Kavukcuoglu2025-wr}, and DeepSeek family \citep{DeepSeek-AI2024-mv, DeepSeek-AI2025-wm}, and locally-deployable open models (referred to as ``open models'') from Mistral \citep{Jiang2024-uo, Jiang2023-xi}, Qwen \citep{Qwen2024-rw, Hui2024-qw}, OLMo \citep{Groeneveld2024-pt}, and Llama \citep{Grattafiori2024-yn} family. 
For frontier-scale models, we use API access from the corresponding company; for open models, we use \texttt{vLLM(v0.7.2)} \citep{kwon2023efficient} to run all experiments. 
We set the temperature to $0.75$ and $max\_tokens$ to $4096$ for all tasks, unless the original task requires a different setting.\footnote{Tasks with different decoding parameters: CreativityIndex ($temperature=1.0, max\_tokens=288$), Creative Math ($temperature=0.0, max\_tokens=2000$). For all tasks, outputs longer than pre-defined $max\_tokens$ will be truncated. Analysis on output length and the effect of different temperatures are in Appendix \ref{app-stability}.} More inference configurations are in Appendix \ref{app-task-details}.

\subsection{Evaluation}
\paragraph{Aggregated Creativity Scoring}
\label{sec-exp-score}
To capture overall performance within each dimension, we aggregate all metrics in that dimension to produce quality, novelty, and diversity scores. The score aggregation calculation involves two steps: first, every evaluation metric is min-max normalized to a 0–1 scale based on the minimum and maximum possible scores for the task. Second, the normalized metrics are aggregated into quality, novelty, and diversity scores for each LLM (as categorized in Figure~\ref{fig:main_fig}) by averaging all metrics within each dimension. To ensure equal weighting across tasks, we first average the normalized scores within each task before computing dimension scores. This prevents tasks with multiple metrics in one dimension (e.g., TTCW has three novelty metrics) from disproportionately influencing the final score.
We also provide an ``overall'' creativity score (shown in Table \ref{tab:perf-summary}) by averaging across the three dimensions to facilitate holistic model comparison. 
Note that this ``overall'' score is only to facilitate model comparison. The relative importance of quality, novelty, and diversity varies depending on the application. We suggest future researchers choose from these three separate scores based on their specific research goals. 
More details about score aggregation can be found in Appendix \ref{app-benchmark-detail}.

\begin{table}[h]
    \centering
    \footnotesize
    \begin{tabular}{l c c c}
    \toprule
    \textbf{Task} & \textbf{Fleiss Kappa} & \textbf{Judge-LLM} & \textbf{Judge Quality} \\
    \midrule
    AUT & 0.650 & GPT4.1 & \checkmark pass\\
    \midrule
    NeoCoder & 0.471 & GPT4.1 & \checkmark pass\\
    \midrule
    TTCW (Originality - Theme \& Content) & 0.660 & Qwen2.5-72B& \checkmark pass\\
    TTCW (Originality - Thought) & 0.400 & Qwen2.5-72B& \checkmark pass\\
    TTCW (Originality - Form) & 0.410 & Qwen2.5-72B& \checkmark pass\\
    TTCW (Flexibility - Perspective \& Voice) & 0.440 & Qwen2.5-72B& \checkmark pass\\
    TTCW (Fluency - Narrative Ending) & 0.480 & Qwen2.5-72B& \checkmark pass\\
    \midrule
    TTCT (Fluency) & 0.432 & GPT4.1, Qwen2.5-72B& \checkmark pass\\
    TTCT (Flexibility) & 0.423 & GPT4.1, Qwen2.5-72B& \checkmark pass\\
    TTCT (Elaboration) & 0.445 & GPT4.1, Qwen2.5-72B& \checkmark pass\\
    \midrule
    CreativeMath (Novelty) & 0.450 & \makecell{Claude3-Sonnet, GPT4.1,\\ Gemini2.0-Flash} & \checkmark pass\\
    CreativeMath (Correctness) & - & \makecell{Claude3-Sonnet}  & \textit{accuracy: 0.91}\\
    \bottomrule
    \end{tabular}
    \caption{Inter-annotator agreement and Judge-LLM quality. ``\checkmark pass'' refers to passing the Alternative Annotator Test; ``accuracy'' is the accuracy compared to ground truth.}
    \label{tab:fleiss_kappa}
    \vspace{-2em}
\end{table}

\paragraph{LLM-as-a-Judge Reliability}
\label{sec-exp-llm-judge}
In \ours{}, the evaluation of five tasks (AUT, TTCW, CreativeMath, TTCT, and NeoCoder) involves using LLM as part of the automatic evaluation procedure. To ensure the reliability of the LLM judges in those tasks, we verify two requirements for every metric in these tasks. First, we collect human annotations and compute inter-annotator agreement, confirming that the task is well-defined and that annotators can reach reasonable agreement on the evaluation outcomes.\footnote{TTCW has its own expert annotation, so we do not collect annotations for it. Details in Appendix \ref{app-llm-judge}.} Second, we follow \citet{alternative-annotator-test} and test if our LLM judges' setups can align well with human annotators. 

For human inter-annotator agreement, we randomly sample a small subset of data points from the output of inference models and have annotators rate the inference output based on the task rubrics. In total, 10 annotators participated in the LLM-Judge verification, all of whom are Ph.D. students or faculty in the field of computer science in U.S. institutions. Annotator training details can be found in Appendix \ref{app-llm-judge}. Three annotators annotate each data point, and the inter-annotator agreement is measured by Fleiss Kappa \citep{fleiss1971measuring}, or quadratic-weighted Fleiss Kappa, if the labels are on a Likert scale. Given the subjective nature of creativity-related tasks, we keep tasks where the Fleiss Kappa measurements among annotators are greater or equal to 0.4, which implies moderate agreement. See detailed agreements in Table \ref{tab:fleiss_kappa}. Note that the Judge quality of CreativeMath (Correctness) is evaluated by accuracy (compared to human judgment) because the correctness of a solution to a math problem is objective. We simply have annotators verify the correctness of each solution for the sampled questions.

After verifying human agreement, we validate our choice of backbone LLM (referred to as ``Judge-LLM''). For the objective task, which only includes Correctness in CreativeMath, we simply calculate the accuracy of our Judge-LLM compared to human judgments. For the subjective tasks, we employ the Alternative Annotator Test \citep{alternative-annotator-test} to statistically justify alignment with human annotators. This method adopts a leave-one-out strategy to validate the LLM as a substitute with an acceptable error margin\footnote{Alternative Annotator Test \citep{alternative-annotator-test} involves a statistical procedure with a pre-specified error margin at significance level $\alpha = .05$}: for each human annotator, we test whether the Judge-LLM aligns closer to the remaining human consensus than the excluded human does (subject to a quality margin $\epsilon$). If the LLM aligns better in more than half the cases (winning rate $> 0.5$), we consider it a viable replacement.

Note that the choice of backbone LLM clearly has an impact on whether or not the Judge-LLM passes this test. We adopt the following principle to choose Judge-LLM: we use Qwen2.5-72B \citep{Qwen2024-rw} as the default backbone LLM due to its open-source nature; if it does not pass the replacement test, we use GPT-4.1 \citep{openai2024gpt4}; if GPT-4.1 does not pass the test, we first try taking the average of GPT-4.1 and Qwen2.5-72B; lastly, we will use the Judge-LLM setup proposed by the original paper that introduced the task. All the Judge-LLM we included in \ours{} pass the Alternative Annotator Test when compared to human annotations. 
\footnote{We exclude sub-tasks and metrics that fail to meet our reliability requirements (Fleiss Kappa $\ge$ 0.4 and passing the Alternative Annotator Test), though we minimize such modifications whenever possible (details in Appendix \ref{app-llm-judge}).}

More details about the human annotation process, Alternative Annotator Test, and discussion on LLM-as-a-Judge failure modes and limitations are in Appendix \ref{app-llm-judge}; evaluation metrics and evaluation prompts for each task are in Appendix \ref{app-task-details}.

%% file: sections/5-analysis.tex
\section{Results \& Analysis}
\label{sec-results-analysis}

\input{sections/Tables/results}
\input{sections/Tables/plot-perf-combined}
\begin{figure}[h]
    \centering
    \includegraphics[width=0.95\linewidth]{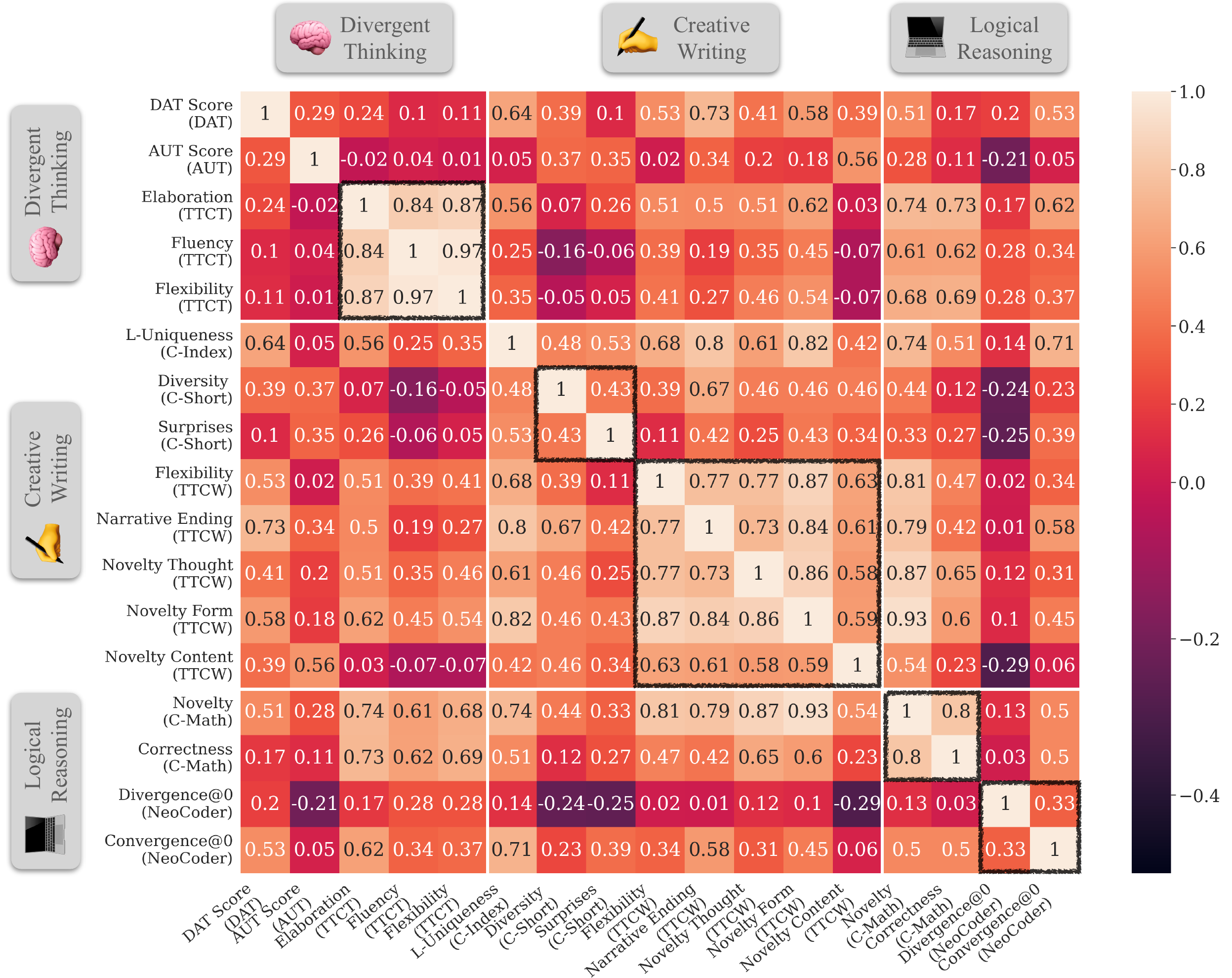}
    \caption{
    Models' performance correlations (Pearson's R), grouped by task and domain; \textbf{C-Index}: Creativity Index; \textbf{C-Short}: Creative Short Story; \textbf{C-Math}: Creative Math; black boxes denote metrics from the same task.
    }
    \vspace{-2em}
    \label{fig:plot-heatmap-task}
\end{figure}
Table \ref{tab:perf-summary} summarizes model performances across domains and three creativity dimensions (quality, novelty, and diversity), where the overall score, averaged across these dimensions, serves as a proxy for a model’s overall creative capability.\footnote{Results reported in the main text portion are from one single run; we reported a case study on performance stability in AUT and TTCW task across 5 different runs with the same configuration in Appendix \ref{app-stability}.} As we can see from the table, Qwen2.5-72B and DeepSeek-V3 are the best-performing models among open models and frontier-scale models. For open models, we can see that the model performances improve as the model size increases.

We have also found a performance improvement along the time axis (Figure \ref{fig:plot-time}) where models released in the past two years have become increasingly competitive. Since many of our metrics (e.g., L-Uniqueness in Creativity Index, Divergence@0 in NeoCoder) would reward models that can generate content different from prior content, having the chance of learning the latest content from the corpus with later cutoff dates would intuitively make models more competitive. More details on model release time details can be found in Appendix \ref{app-model-details}.

\subsection{Gap Between Frontier-scale Models and Open Models}
\paragraph{Overall Performance Gap} As shown in Table \ref{tab:perf-summary}, the best frontier-scale model(s) outperform the best open model(s) by more than \textbf{.10 (or 15\%) in each dimension} of creativity. This shows a big gap between frontier-scale and open models when it comes to creativity-related tasks. A more in-depth breakdown of this gap can be found in Figure \ref{fig:plot-diff}, with the gaps of the average performance of three open model groups (by model sizes) compared to that of all frontier-scale models. Analysis of this figure leads to the following two findings.

\paragraph{Domain-Specific Differences} Among the three domains, \textbf{logical reasoning} and \textbf{creative writing} see a notably larger gap than divergent thinking. We hypothesize that this is because those tasks are more closely related to real-world applications than divergent thinking tasks, and thus the companies that developed these frontier-scale models emphasize those two aspects of LLM training. In particular, all frontier-scale models include coding and mathematical reasoning as part of evaluation in their technical report \citep{openai2024gpt4, anthropic2024claude3, DeepSeek-AI2024-mv, DeepSeek-AI2025-wm, google2024gemini2}; most models include some writing tasks, such as GRE Test \citep{openai2024gpt4, anthropic2024claude3}, or include creative writing or role-playing data as part of the post-training data \citep{DeepSeek-AI2024-mv, DeepSeek-AI2025-wm}, whereas none of these models has put special emphasis in divergent thinking task during training or evaluation.

\paragraph{Dimension-Level Differences} Across three creativity dimensions, \textbf{quality} has a larger performance gap than novelty and diversity. We believe the gap in quality comes from a similar reason as mentioned above, as the quality dimension includes many reasoning-related metrics (e.g., Convergence@0 from NeoCoder and Correctness from Creative Math) that would benefit from coding and mathematical tasks during training. 

\subsection{Correlations Among Model Performance}

\begin{figure}[t]
    \centering
    \includegraphics[width=\linewidth]{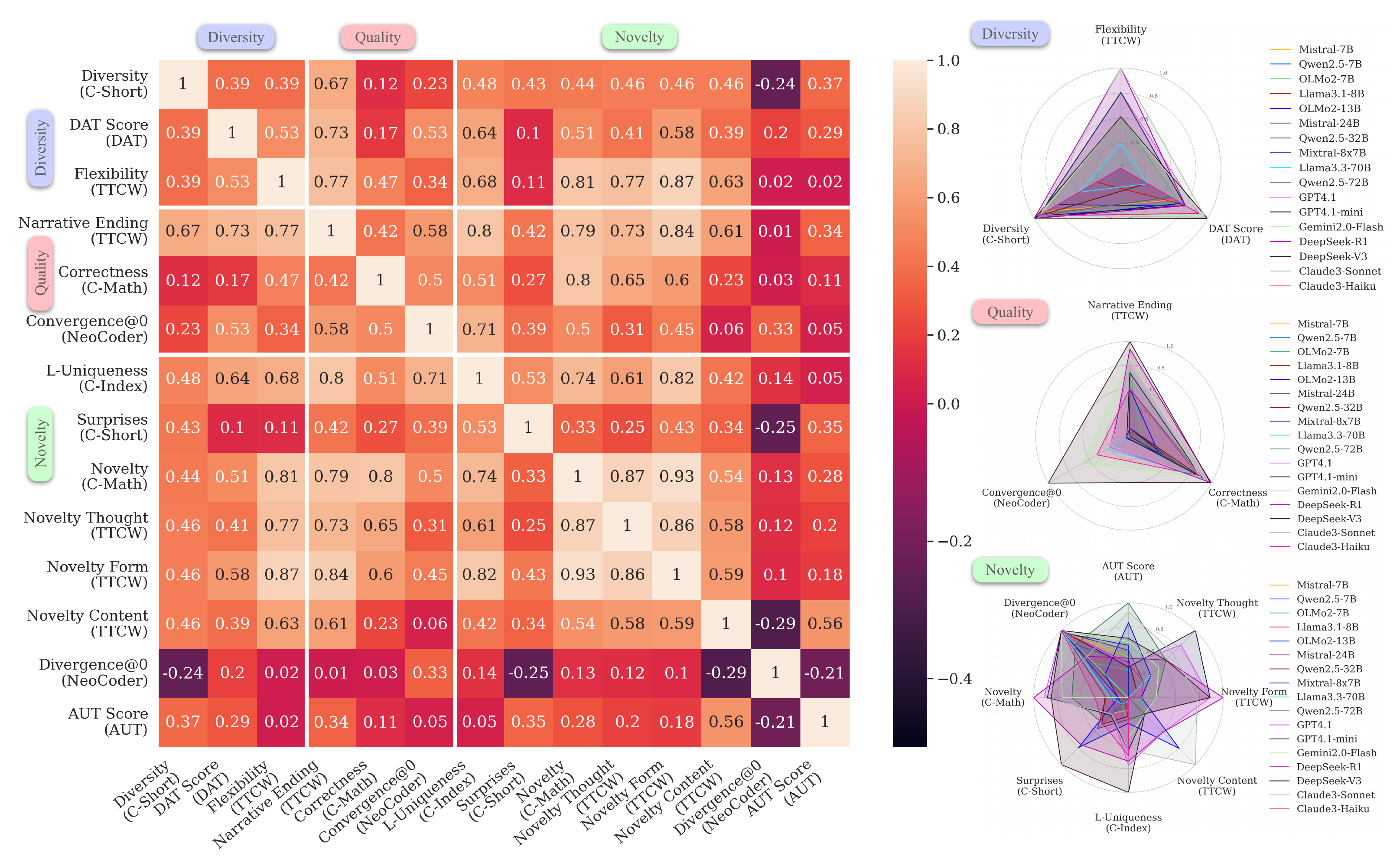}
    \caption{
    \textbf{Left}: model performance correlations (Pearson's R), grouped by dimension; \textbf{Right}: individual LLM performances, min-max normalized by metric. TTCT metrics omitted due to high metric correlation.
    }
    \vspace{-2em}
    \label{fig:plot-heatmap-sub}
\end{figure}
Does a good performance in one task/domain/dimension imply similar superiority in another task/domain/dimension? To answer this research question, we analyze the correlation between models' performance among different tasks, domains, and dimensions.
Specifically, for each metric $m$, we form a vector $\mathbf{s}_m \in \mathbb{R}^{M}$ by stacking the normalized scores of all $M$ models evaluated in \ours{}. 
We then compute the Pearson correlation $r(\mathbf{s}_m,\mathbf{s}_{m'})$ between every pair of metrics $(m,m')$. 
Figure~\ref{fig:plot-heatmap-task}, ~\ref{fig:plot-heatmap-sub} shows the resulting 
correlation matrix, ordered by task and dimension, respectively, so that diagonal blocks correspond to within-task / within-dimension metric groups.

\paragraph{Strong Within-Task Correlations} We find a strong correlation in the models' performance on metrics coming from the same task. As shown by the black bounding boxes in Figure \ref{fig:plot-heatmap-task}, the correlation along the diagonals is most pronounced, with TTCT having correlations $\ge0.84$ for all metrics within this task. In addition, metrics in creative writing tasks (in the central square of the heatmap) generally have decent correlation with other metrics within the same domain, even if they come from different tasks. We believe this comes from a higher inherent similarity among tasks from the creative writing domain than tasks from the other two domains.

\paragraph{Mixed Within-Dimension Correlations} We also observe high correlations among metrics that belong to the diversity or quality dimension, even if they originate from different tasks or domains. As shown in the heatmap in Figure \ref{fig:plot-heatmap-sub}, this is more obvious in diversity and quality dimensions and less so in novelty. The correlation along the diagonal is higher (i.e., lighter) in the top left, while the bottom right (novelty dimension) shows mixed correlations. This observation is also confirmed by the individual model performance, shown in radar charts in Figure \ref{fig:plot-heatmap-sub}, where the model performances for diversity and quality are more organized, while the one for novelty is more crowded. All of these show that the models' performance in any one of the diversity metrics is a good indicator for their performance in other diversity metrics; the same goes for quality metrics. On the other hand, metrics in the novelty dimension have low correlations with other metrics in the same dimension, as shown in the bottom right part of Figure \ref{fig:plot-heatmap-sub}. We believe these findings highlight the varying definitions of novelty across tasks and domains. For example, Surprises (Creative Short Story) measures the semantic transitions across neighboring sentences in stories, whereas Divergence@0 (NeoCoder) measures the capability of coming up with a solution to a coding problem that is different from existing ones. Given such a huge difference in metric definition, it is not surprising that they even have a negative correlation (-0.25) in model performances. We provide a more in-depth discussion on Divergence@0 (NeoCoder)'s low correlation to other metrics in Appendix \ref{app-perf-discussions}.

\paragraph{Novelty Requires External References} Expanding on low correlation among novelty metrics, we believe this is because novelty is the only dimension that inherently asks ``novel compared to what?'', and the answer differs across tasks: compared to prior human-written corpora (L-Uniqueness, C-Short Surprises), to known human solutions (NeoCoder Divergence@0, C-Math Novelty), or to common conceptual associations (AUT Score). In contrast, quality asks ``is this output well-formed and coherent?'', which is a property intrinsic to the output; diversity asks ``how much does this output vary from other outputs?'', which is measurable within the model's own generation distribution. This reference heterogeneity means novelty metrics measure a family of related but operationally distinct notions of deviation from a norm rather than a single latent construct. A model scoring high on one novelty metric may score low on another for interpretable reasons: AUT Score reflects unusual-ness of free associations in a single pass, while C-Math Novelty demands solutions that are both correct and absent from the human solution set, which are two very different demands. Therefore, having weak or negative correlation among performances in novelty metrics is not a weakness but a strength of our proposed framework. Since all the metrics have their limitations and our framework unifies them together, our result also further validates that none of them on its own is well-representative for novelty. Hence, we recommend testing them all using our benchmark to reveal the full landscape of creativity of the target LLM being evaluated. 

\paragraph{Weak Cross-Task or Cross-Domain Correlations} Metrics from different domains (e.g., divergent thinking v.s. creative writing in Figure \ref{fig:plot-heatmap-task}) and metrics from different dimensions (e.g., novelty v.s. diversity in Figure \ref{fig:plot-heatmap-sub}) all have relatively lower correlations, compared to within-domain or within-dimension correlations. In other words, models performing well in one domain or in one dimension of creativity do not necessarily perform similarly well in another domain or dimension. This again confirms the necessity of including a diverse set of tasks and creativity dimensions to achieve a cross-domain evaluation of creativity. We have additional statistical tests that validate our framework design in Appendix \ref{app:corr_stats}.

%% file: sections/Tables/results.tex
\begin{table}[h]
\centering
\footnotesize
\begin{tabular}{lc|ccc|ccc}
\toprule
\multirow{2}{*}{\textbf{Model}} & \multirow{2}{*}{\textbf{Overall}} & \multirow{2}{*}{\textbf{Quality}} & \multirow{2}{*}{\textbf{Novelty}} & \multirow{2}{*}{\textbf{Diversity}} & \textbf{Creative} & \textbf{Divergent} & \textbf{Logical} \\
&&&&&\textbf{Writing} & \textbf{Thinking} & \textbf{Reasoning} \\
\midrule
\textbf{<10B}      &  &  &  &  &  &  &  \\
    Mistral-7B      & .423 & .268 & .393 & .607 & .316 & .679 & .320\\
    Qwen2.5-7B      & .406 & .374 & \textbf{.398} & .445 & .207 & .654 & \textbf{.460}\\
    OLMo2-7B        & \textbf{.462} & \textbf{.405} & .340 & \textbf{.640} & \textbf{.403} & .643 & .257\\
    Llama3.1-8B     & .404 & .313 & .391 & .509 & .239 & \textbf{.683} & .409\\
\midrule
\textbf{10-40B}     &  &  &  &  &  &  &  \\
    OLMo2-13B       & .451 & .389 & .379 & .586 & \textbf{.424} & \textbf{.672} & .278\\
    Mistral-24B     & .448 & .352 & \textbf{.440} & .553 & .346 & .614 & \textbf{.473}\\
    Qwen2.5-32B     & \textbf{.453} & \textbf{.417} & .336 & \textbf{.605} & .338 & .655 & .358\\
\midrule
\textbf{40-80B}     &  &  &  &  &  &  &  \\
    Mixtral-8x7B    & .430 & .318 & .392 & .578 & .278 & .687 & .420\\
    Llama3.3-70B    & .446 & .401 & .404 & .534 & .269 & .614 & .529\\
    Qwen2.5-72B     & \textbf{.549} & \textbf{.526} & \textbf{.490} & \textbf{.632} & \textbf{.385} & \textbf{.736} & \textbf{.554}\\
\midrule
\textbf{Frontier-scale}     &  &  &  &  &  &  &  \\
    Claude3-Sonnet  & .632 & .606 & .557 & .733 & .507 & .765 & .612\\
    Claude3-Haiku   & .535 & .522 & .456 & .627 & .413 & .685 & .568\\
    GPT4.1          & .641 & .618 & .582 & .732 & .518 & .722 & .682\\
    GPT4.1-mini     & .615 & .567 & .571 & .718 & .504 & .697 & .649\\
    Gemini2.0-Flash & .626 & .657 & .522 & .712 & .438 & .753 & .655\\
    DeepSeek-R1     & .665 & .677 & .575 & \textbf{.742} & .548 & .724 & .643\\
    DeepSeek-V3     & \textbf{.695} & \textbf{.749} & \textbf{.616} & .728 & \textbf{.571} & \textbf{.768} & \textbf{.726}\\
\bottomrule
\end{tabular}

\caption{Model performance on \ours{}, grouped by model size.
Frontier-scale models are grouped together. All scores are between 0 and 1, and the higher the better. Overall is the average of Quality, Novelty, and Diversity scores. The rightmost three columns are the average scores across tasks in each domain. \textbf{Bold} are the best results in the corresponding model group. Refer to Figures \ref{fig:plot-perf-combined}, \ref{fig:plot-perf-quality}, \ref{fig:plot-perf-novelty}, \ref{fig:plot-perf-diversity}, 
\ref{fig:plot-perf-story}, \ref{fig:plot-perf-psyc}, \ref{fig:plot-perf-logical} for a visualized performance comparison.}
\label{tab:perf-summary}
\vspace{-2em}
\end{table}


%% file: sections/Tables/plot-perf-combined.tex
\begin{figure}[h]
    \centering
    \begin{subfigure}[t]{0.37\linewidth}
    \centering
        \includegraphics[width=\linewidth]{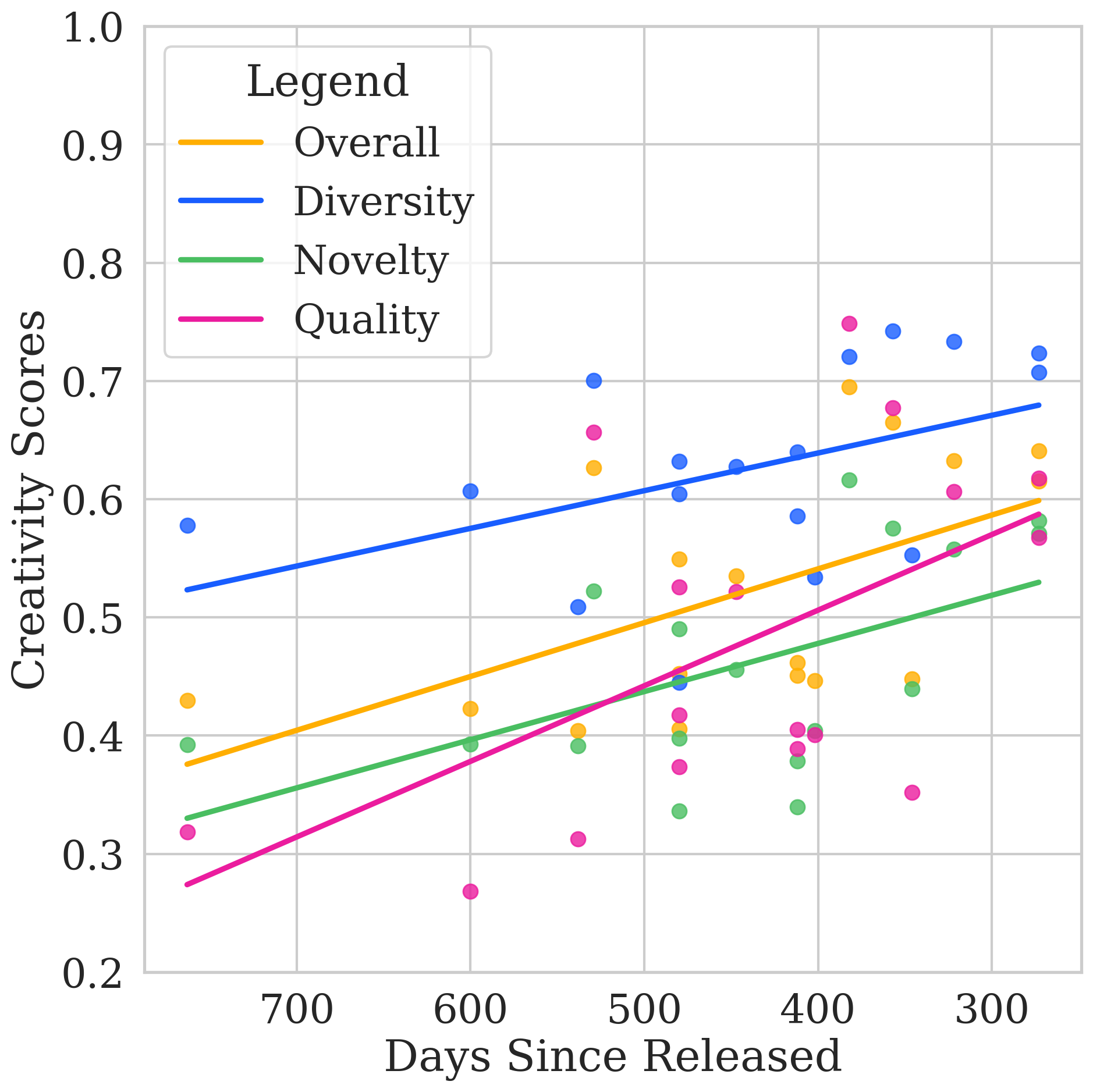}
        \caption{}
    \label{fig:plot-time}
    \end{subfigure}%
    \hfill
    \begin{subfigure}[t]{0.6\linewidth}
    \centering
        \includegraphics[width=\linewidth]{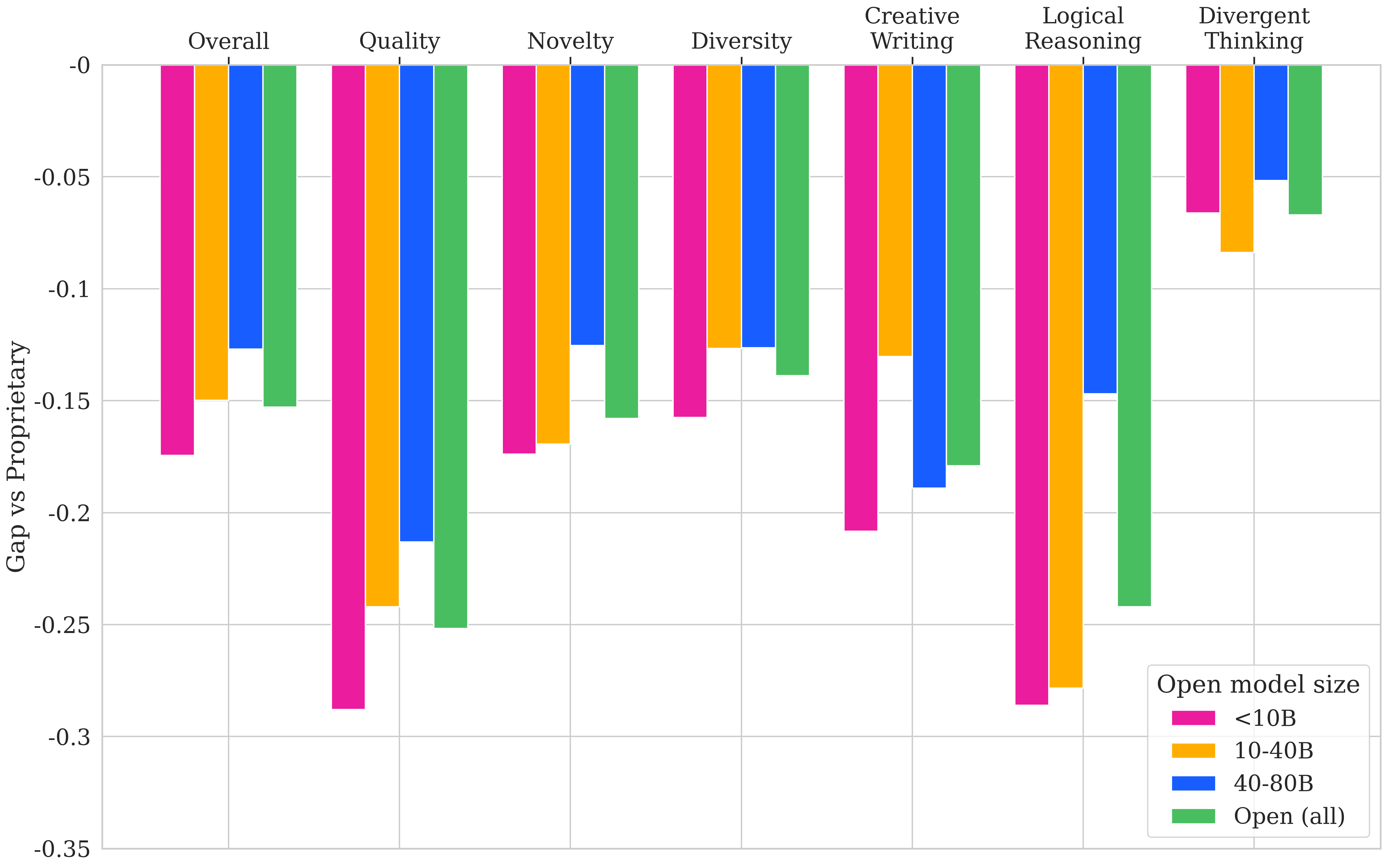}
    \caption{}
    \label{fig:plot-diff}
    \end{subfigure}
    \caption{\textbf{(a)} Performance v.s. Days since LLM release date. 
    The line represents best-fit linear regression, showing model performance in all dimensions improving over time. \textbf{(b)} Performance gap between the open models and the frontier-scale models, averaged by model size group.
    }
    \vspace{-1em}
    \label{fig:plot-perf-combined}
\end{figure}

%% file: sections/6-conclusion.tex
\section{Conclusion}
\label{sec:conclusion}
We propose \ours{}, a cross-domain and scalable evaluation framework designed to uncover the complexity of LLM creativity through tasks in three distinct domains and seventeen metrics covering quality, novelty, and diversity. We evaluate 17 LLMs from multiple families of frontier-scale and locally deployable open LLMs and analyze the correlation among model performances across domains and dimensions. 
With \ours{}, LLM developers will be able to systematically evaluate LLM creativity and identify the direction of optimization for more creative LLMs.

\paragraph{Limitations and Future Work}
While \ours{} provides a cross-domain and scalable evaluation framework for evaluating LLM creativity, we acknowledge three primary limitations. 
First, we do not cover all modalities and all creativity domains. We exclusively focus on text data because we prioritize establishing a robust evaluation framework for text before expanding to multimodal LLMs / VLMs, where reliable automatic evaluation methods are less mature. As for domain coverages, our task selection does not cover some highly creative domains, such as scientific discovery and inventive design, due to a lack of automatic evaluation in those domains. Nevertheless, the modular design of \ours{} allows for easy extension to tasks with different data modality or domain in future iterations. Second, although we did our best to validate our LLM-Judges, limitations remain: annotation sample sizes are relatively small due to resource constraints, annotators are all task-familiar CS researchers instead of creativity domain experts, judge backbone choice introduces potential bias that we do not fully characterize, and LLM-Judges may carry systematic biases in assessing certain types of creative content. A promising direction for future work is adopting a multi-LLM judge jury system \citep{Verga2024-us}, where consensus across multiple independently-prompted judge backbones replaces reliance on any single model, providing robustness to individual judge biases without requiring prompt-matched comparisons. Third, given the inevitable data contamination problem, models with a later release date may have advantages on metrics that favor different outputs compared to existing content (e.g., CreativityIndex, CreativeMath, etc). This is an inherent limitation that needs to be aware of.

Given these limitations, we advise LLM developers or researchers to 1) check the task description and example, 2) select the dimension and task domain that is closely related to their topic of interest, and 3) only use the model performance on that subset (domain \& dimension) as a reference for result interpretation. More discussion on limitations is in Appendix \ref{app-limitations}.

\section*{Acknowledgment}
This research was supported in part by ONR grant (N00014-241-2089), OpenAI’s Researcher Access Program, in part by the Center for Intelligent Information Retrieval, in part by Cisco, and in part by the University of Pittsburgh Center for Research Computing and Data, RRID:SCR\_022735, through the resources provided. Specifically, this work used the HTC cluster, which is supported by NIH award number S10OD028483, and the H2P cluster, which is supported by NSF award number OAC-2117681.
This work was also in part carried out at the Advanced Research Computing at Hopkins (ARCH) core facility (DSAI), which is supported by the National Science Foundation (NSF) grant number OAC1920103. 
Any opinions, findings, and conclusions or recommendations expressed in this material are those of the authors and do not necessarily reflect those of the sponsor.

We additionally thank the annotation effort from Ali, Arun, Janet, Johnny, Salem, Wanzhou, and Yufei (in alphabetical order) for LLM-Judge verification. 

%% file: sections/appendix.tex
\newpage


\appendix
\newpage
\section{Code \& Data Release Plan}
We plan to release the full evaluation pipeline, task wrappers, prompts, pre-processing scripts, LLM-Judge configurations, and all the human annotations upon publication. We will also establish a protocol for adding new tasks, including but not limited to: task-related data structure, inference and evaluation code interface (e.g., what kinds of functions they need to have), sample inference and evaluation scripts, and developer contact for people to submit their tasks. We also have a leaderboard website under construction to publicly show benchmark results. We believe that all of these combined will promote community contribution to a truly comprehensive and up-to-date creativity assessment. 

\section{Model Details}
\label{app-model-details}

\begin{table}[htbp]  
  \centering
    \begin{tabular}{lllll}
    \toprule
    \textbf{Short Name} & \textbf{Exact Model Name} & \textbf{Size} & \textbf{Family} & \textbf{Release Time} \\
    \toprule
    Mistral-7B & Mistral-7B-Instruct-v0.3 & 7B & Mistral & 05/2024 \\
    Qwen2.5-7B & Qwen2.5-7B-Instruct & 7B & Qwen & 09/2024 \\
    OLMo2-7B & OLMo-2-1124-7B-Instruct & 7B & OLMo & 11/2024 \\
    Llama3.1-8B & Llama-3.1-8B-Instruct & 8B & Llama & 07/2024 \\
    \midrule
    OLMo2-13B & OLMo-2-1124-13B-Instruct & 13B & OLMo & 11/2024 \\
    \midrule
    
    Mistral-24B & Mistral-Small-24B-Instruct-2501 & 24B & Mistral & 01/2025 \\
    Qwen2.5-32B & Qwen2.5-32B-Instruct & 32B & Qwen & 09/2024 \\
    Mixtral-8x7B & Mixtral-8x7B-Instruct-v0.1 & 56B & Mistral & 12/2023 \\
    Llama3.3-70B & Llama-3.3-70B-Instruct & 70B & Llama & 12/2024 \\
    Qwen2.5-72B & Qwen2.5-72B-Instruct & 72B & Qwen & 09/2024 \\
    \midrule
    
    Claude3-Sonnet & claude-3-7-sonnet-20250219 & - & Claude & 02/2025 \\
    Claude3-Haiku & claude-3-5-haiku-20241022 & - & Claude & 11/2024 \\
    GPT4.1 & gpt-4.1-2025-04-14 & - & GPT & 04/2025 \\
    GPT4.1-mini & gpt-4.1-mini-2025-04-14 & - & GPT & 04/2025 \\
    Gemini2.0-Flash & gemini-2.0-flash & - & Gemini & 12/2024 \\
    DeepSeek-R1 & deepseek-reasoner & - & DeepSeek & 01/2025 \\
    DeepSeek-V3 & deepseek-chat  & - & DeepSeek & 12/2024 \\
    \bottomrule
    \end{tabular}
    \caption{List of models included in our experiments.}
    \label{table-model-details}
\end{table}

\paragraph{Deepseek Models} For Deepseek models, we also use API due to constraints in compute resources. API console: \url{https://platform.deepseek.com}.
\newpage
\section{Evaluation Framework Design}

\label{app-benchmark-detail}
\subsection{Dataset Sizes}

\begin{table}[th]
\small
\centering
    \begin{tabular}{lll}
        \toprule
        \textbf{Task} & \textbf{Count} & \textbf{Note} \\
        \midrule
        AUT & 105 (tool use) & 21 tools with 5 rounds of prompting per tool \\
        DAT & 100 (round) & No input data, we prompt each LLM 100 rounds \\
        TTCT & 500 (question) & 5 tasks (100 questions/task) \\
        \midrule
        TTCW & 12 (story prompt) & One story per story prompt \\
        Creative Short Story & 10 (keyword tuple) & One story per keyword tuple \\
        Creativity Index & 300 (document sample) & 100 samples from 3 subsets: book, poem, and speech\\
        \midrule
        NeoCoder & 198 (question) & One solution per coding question \\
        Creative Math & 373 (question) & One solution per math question \\        
        \bottomrule
    \end{tabular}
    \vspace{0.5em}
    \caption{Dataset size of \ours{}. See Appendix \ref{app-llm-judge} for details on why we only select 5 tasks instead of all 7 tasks from the original TTCT paper. More details for other tasks can be found in the corresponding section of Appendix \ref{app-task-details}.}
    \label{table:data-size}
\end{table}

\subsection{Score Calculations}
\label{app-score-aggregation}
\paragraph{Score Normalization} For every model $i$ and every raw metric score $S_{i,m}$ (metric $m$ lives on some known scale $[\min_m,\max_m]$), the normalized score $\hat S_{i,m}$ is given by:

$$
\boxed{\hat S_{i,m}\;=\;\frac{S_{i,m}-\min_m}{\max_m-\min_m}}
$$

For example, AUT score is on a 1–5 Likert scale:
$\hat S_{i,\text{AUT}} = \dfrac{S_{i,\text{AUT}}-1}{5-1} = \dfrac{S_{i,\text{AUT}}-1}{4}$.

\paragraph{Aggregate Normalized Scores}
First, we collapse multiple metrics within the same task: if task $t$ has a set $M_t$ of $k_t$ metrics in a given dimension (e.g. three quality metrics for TTCW), average them first:

$$
\boxed{
    \bar S_{i,t}\;=\;\frac{1}{k_t}\sum_{m\in M_t}\hat S_{i,m}
}
$$
\newpage
Then, we take average across all tasks that belong to that dimension. Let $T_{\text{qual}}, T_{\text{nov}}, T_{\text{div}}$ be the task sets for quality, novelty, diversity. For dimension $d\in\{\text{qual},\text{nov},\text{div}\}$:

$$
\boxed{
    D_{i}^{(d)}\;=\;\frac{1}{|T_{d}|}\sum_{t\in T_{d}}\bar S_{i,t}
}
$$

In this way, we end up with three numbers per model:
$D_{i}^{(\text{qual})},\;D_{i}^{(\text{nov})},\;D_{i}^{(\text{div})}$. We can also calculate aggregated score for creative writing, divergent thinking, and logical reasoning (as shown in Table \ref{tab:perf-summary}).

\paragraph{Overall creativity score}

Just take the straight mean of those three dimension scores to stay balanced:

$$
\boxed{
    C_i\;=\;\frac{D_{i}^{(\text{qual})}+D_{i}^{(\text{nov})}+D_{i}^{(\text{div})}}{3}
}
$$

\subsection{Aggregation Method}

We compare four aggregation methods to assess whether model rankings are robust to the choice of aggregation procedure:

\begin{itemize}
    \item \textbf{M0} (Used in \ours{}): normalize metrics -> average within task -> average tasks within domain/dimension (\textit{task-equal score weighting})
    \item \textbf{M1}: normalize metrics -> average all metrics directly within domain/dimension (\textit{metric-equal score weighting})
    \item \textbf{M2}: convert each metric to a rank (1 = best) -> average ranks directly within domain/dimension (\textit{metric-equal rank aggregation})
    \item \textbf{M3}: convert each metric to a rank -> average ranks within task -> average task ranks within domain/dimension (\textit{task-equal rank aggregation})
\end{itemize}

Rankings are computed on the mean score/rank across the three domains (or dimensions), where for M0/M1, a higher score = better rank, and for M2/M3, a lower average rank = better. “Max swing” refers to the maximum ranking change among the four aggregating methods.

\input{sections/Tables/rankings}

As shown in Table \ref{tab:rank-domain}, \ref{tab:spearman-domain}, \ref{tab:rank-dimension}, \ref{tab:spearman-dimension}, all four methods produce highly consistent rankings, with Spearman $\rho \geq .92$ across all method pairs for both domain- and dimension-level aggregation. The top 8 models (all proprietary, both DeepSeek models, and Qwen2.5-72B) are stable across all methods. The small number of rank swings (max 5 positions, confined to mid-tier open-source models) reflects genuine score proximity in that range rather than sensitivity to aggregation choice. These results confirm that the main findings of
\ours{} are robust to the choice of aggregation procedure.

\subsection{Alternative Formulation}
One obvious alternative formulation, compared to our ``Quality-Diversity-Novelty'' formulation, is the ``usefulness and originality''. However, merging novelty and diversity into a single originality score obscures meaningful per-model differences: OLMo2-7B ranks 7th on diversity but 16th on novelty (a 9-position gap, with an absolute score difference of .30), indicating a model that generates varied outputs but rarely departs from conventional content; Qwen2.5-32B shows a similar pattern (11th vs. 17th, gap of 6, score difference .29). The reverse also occurs: Claude3-Sonnet ranks 2nd on diversity, but 5th on novelty, and DeepSeek-R1 ranks 1st on diversity but 3rd on novelty. Such dissociation, i.e., high diversity with low novelty, or vice versa, corresponds to qualitatively different creative behaviors that a binary originality score would average away, and are relevant for both model evaluation and development.

\section{LLM-as-a-Judge Design Details}
\label{app-llm-judge}
Five out of eight tasks in our evaluation framework require LLM-as-a-Judge for at least one metric. Here we present the details of those Judge-LLMs.

\paragraph{Objective Task}
We consider the correctness judgment of CreativeMath as an objective task and do not report the inter-annotator agreement.

\paragraph{Annotator Training} Each annotator starts by going through no more than 3 example datapoints with the leading author available for any clarification questions either in-person or virtually. After confirming the annotation is done correctly, the author will leave the annotator to work on the remaining annotations.

\begin{figure}[H]
    \centering
    \includegraphics[width=0.6\linewidth]{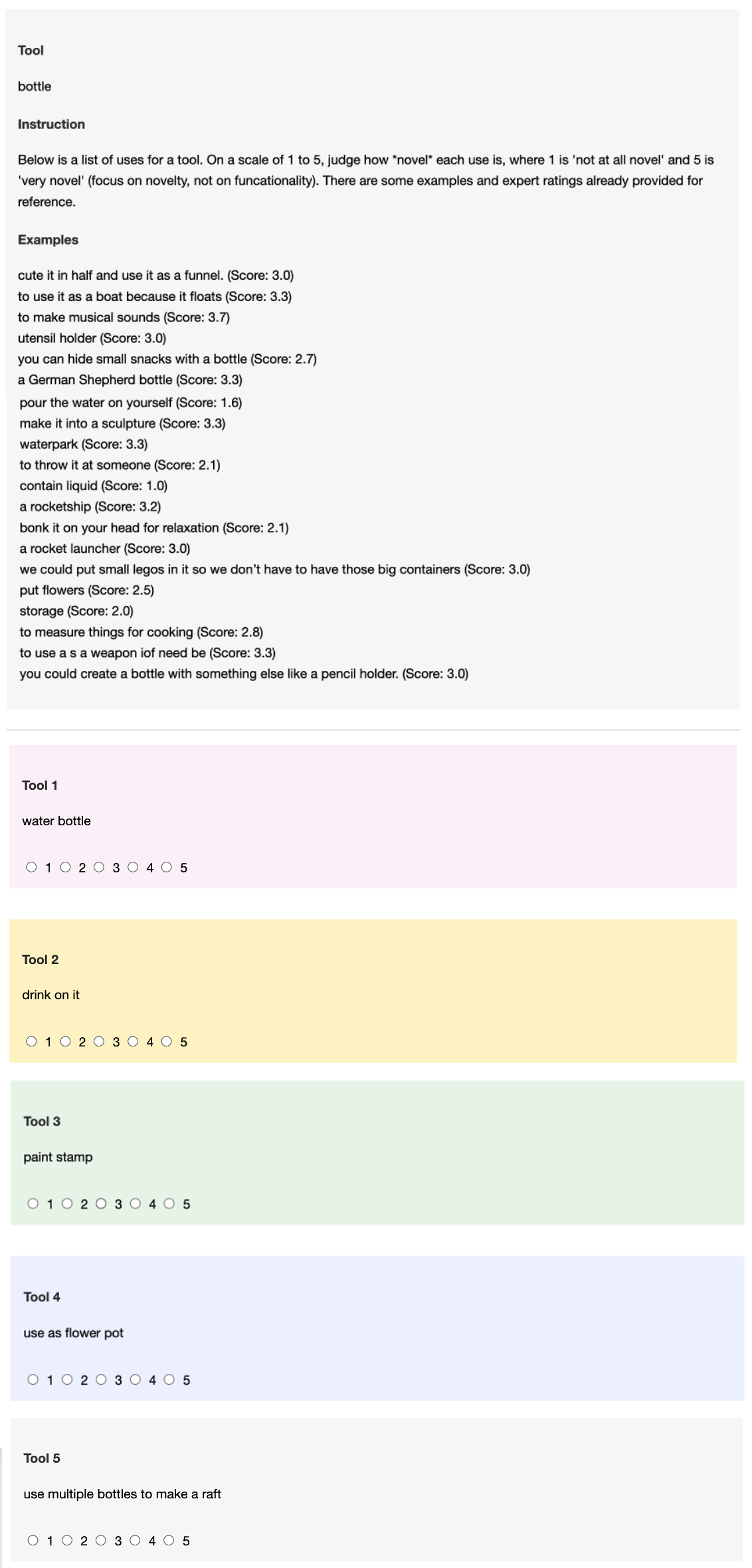}
    \caption{Annotation interface for AUT.}
    \label{fig:interface-aut}
\end{figure}

\begin{figure}[H]
    \centering
    \includegraphics[width=0.5\linewidth]{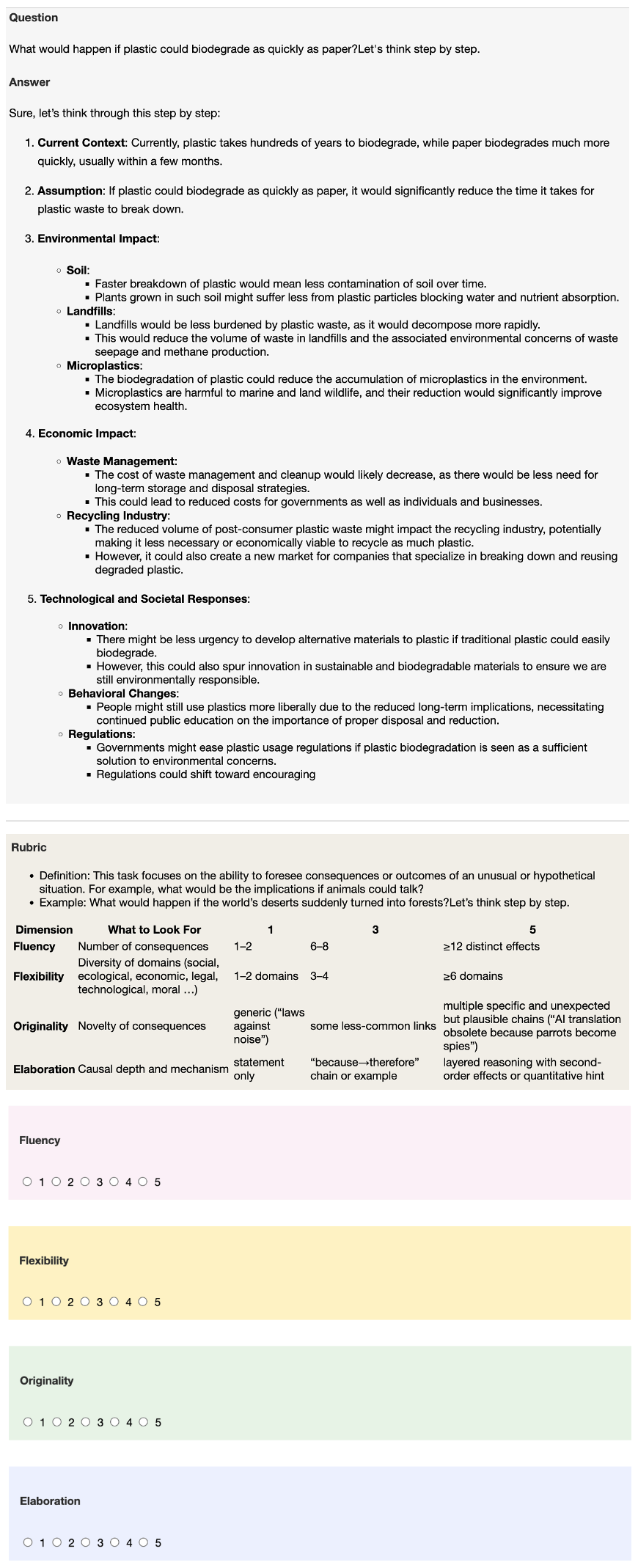}
    \caption{Annotation interface for TTCT.}
    \label{fig:interface-ttct}
\end{figure}

\begin{figure}[H]
    \centering
    \includegraphics[width=\linewidth]{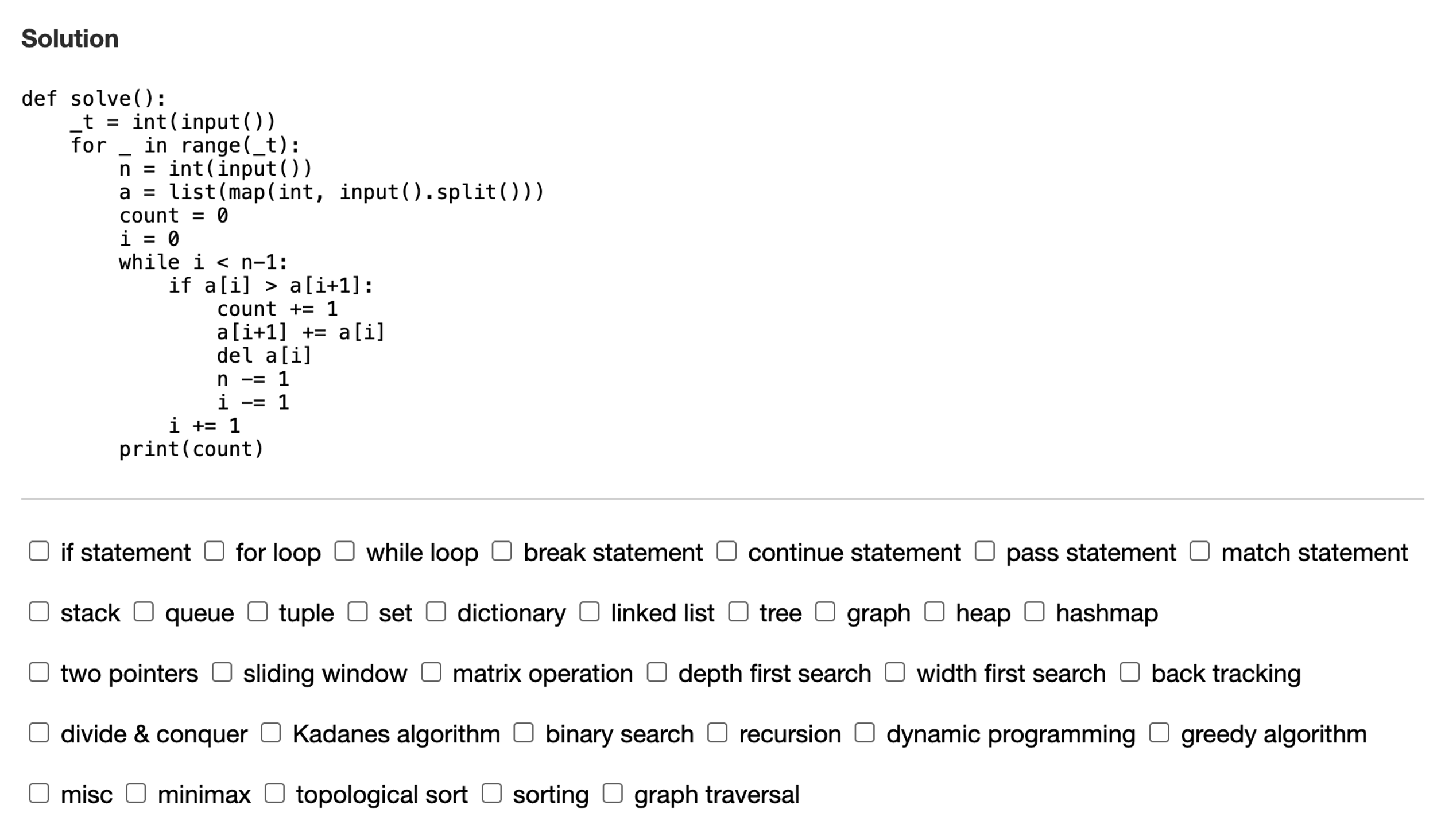}
    \caption{Annotation interface for NeoCoder.}
    \label{fig:interface-neocoder}
\end{figure}

\begin{figure}[H]
    \centering
    \includegraphics[width=0.7\linewidth]{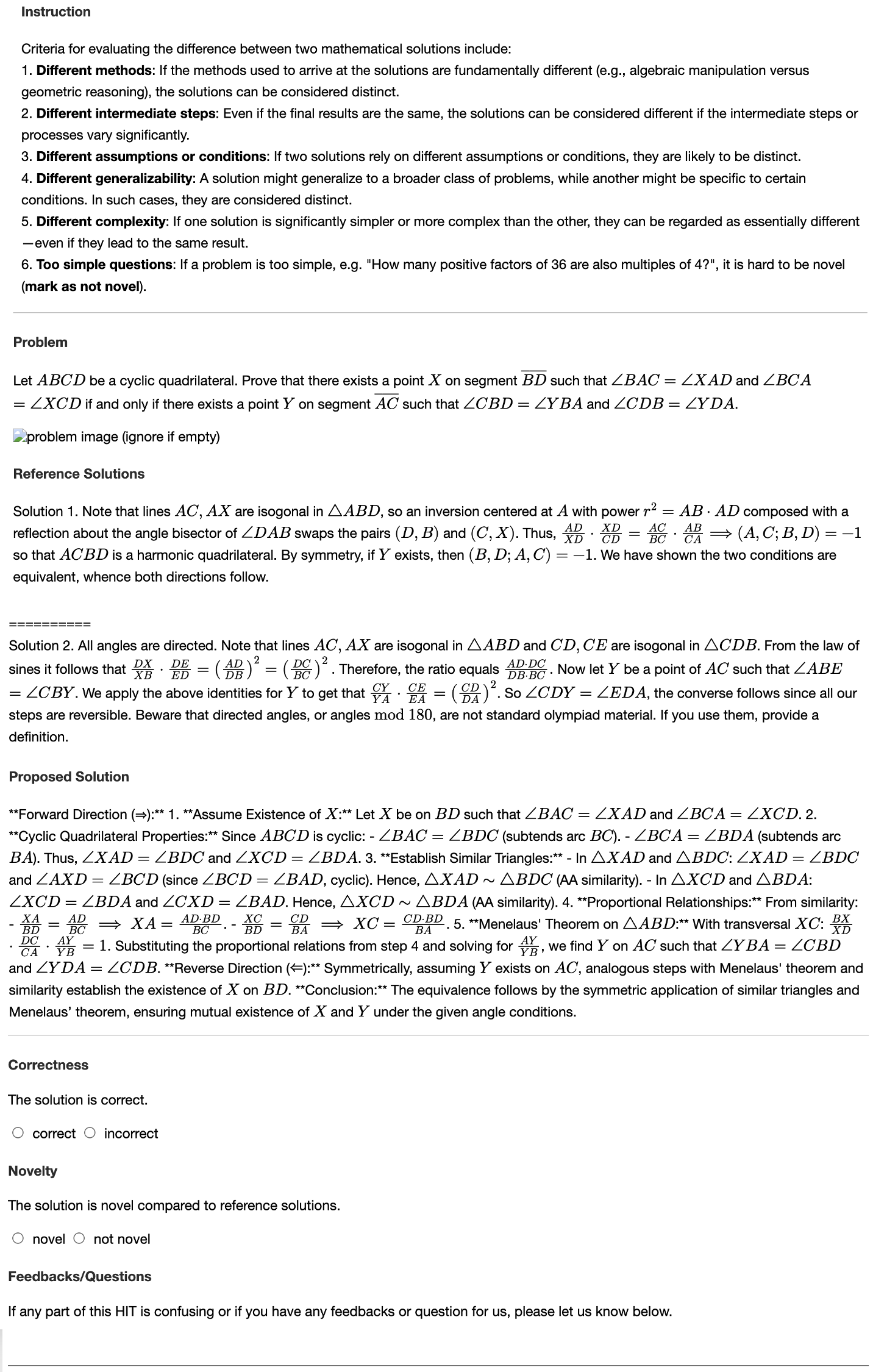}
    \caption{Annotation interface for CreativeMath.}
    \label{fig:interface-creativemath}
\end{figure}

\paragraph{Inter-annotator Agreement}
Here we outline human annotation details. The number of datapoints we annotated for each task is shown in Table \ref{tab:anno_statistics}, followed by annotator information and how we calculate Fleiss Kappa. ``Researchers'' here refers to either paper authors or researchers who work in the research related to LLM evaluation. The annotation interfaces are in Figure \ref{fig:interface-aut}, \ref{fig:interface-ttct}, \ref{fig:interface-neocoder}, \ref{fig:interface-creativemath} (we do not conduct annotation for TTCW because expert annotation is released by the original authors).

\begin{table}[b]
\centering
\footnotesize
\begin{tabular}{l l l l}
\toprule
\textbf{Dataset} & \textbf{Number of Data Points} & \textbf{Annotators} & \textbf{IAA Measurement} \\
\midrule
AUT & 88 tool uses (for 20 tools) & Researchers & \textit{QWK} \\
TTCT & 42 question-answer pairs & Researchers & \textit{QWK} \\
NeoCoder & 60 problem-solution pairs & Domain Experts & \textit{FK} \\
TTCW & 36 LLM-generated stories & Domain Experts & \textit{FK} \\
{CreativeMath (Novelty)} & 48 problem-solution pairs & Researchers & \textit{FK} \\
\midrule
{CreativeMath (Correctness)} & 78 problem-solution pairs & Researchers & - \\
\bottomrule
\end{tabular}
\caption{Annotation data composition and annotator information; 
\textbf{IAA} refers to inter-annotator-agreement 
\textit{QWK} refers to quadratic-weighted Fleiss Kappa and \textit{FK} refers to regular Fleiss Kappa.}
\label{tab:anno_statistics}
\end{table}

\paragraph{Alternative Annotator Test}

In the Alternative Annotator Test, one key step is to measure how well an annotator (either the Judge LLM or one of the existing annotators) aligns with the remaining annotators. We follow \citet{alternative-annotator-test} and use accuracy for tasks with binary ratings (i.e., TTCW, CreativeMath), root mean squared error (RMSE) for tasks with Likert-scale ratings (i.e., TTCT, AUT), and recall for NeoCoder (since it is a retrieval task).

According to the recommendation of \citet{alternative-annotator-test}, the choice of Epsilon depends on the quality of annotation. As shown in Table \ref{tab:anno_statistics}, only TTCW is annotated by domain experts, which falls into the ``trusted expert'' category, and hence we choose $\epsilon = 0.2$; other tasks are annotated by researchers familiar with the context, which falls into ``skilled annotator'' category, and hence we choose $\epsilon = 0.15$.

\paragraph{Modification to Existing Task Setups} For TTCT, there are seven question types in total with 100 questions per question type (Improvement Task, Story Writing, Just-suppose, Situation, Unusual Uses, Common Problems, Consequences). Given the annotators' feedback, we removed Story Writing and Just-suppose questions because they are very subjective to judge. We also observe a Fleiss Kappa lower than 0.4 for the Originality metric in this task, so we remove that metric as well. For TTCW, the original paper \citep{Chakrabarty2024-fd} reports annotator agreement, from which we only keep the metrics with no less than 0.4 Fleiss Kappa and the Judge-LLM also passed the Alternative Annotator Test.


\newpage
\section{Performance Summaries}
\label{app-perf-summaries}

\subsection{Performance by Domain \& Dimension}
\begin{figure}[H]
    \centering
    \includegraphics[width=0.75\linewidth]{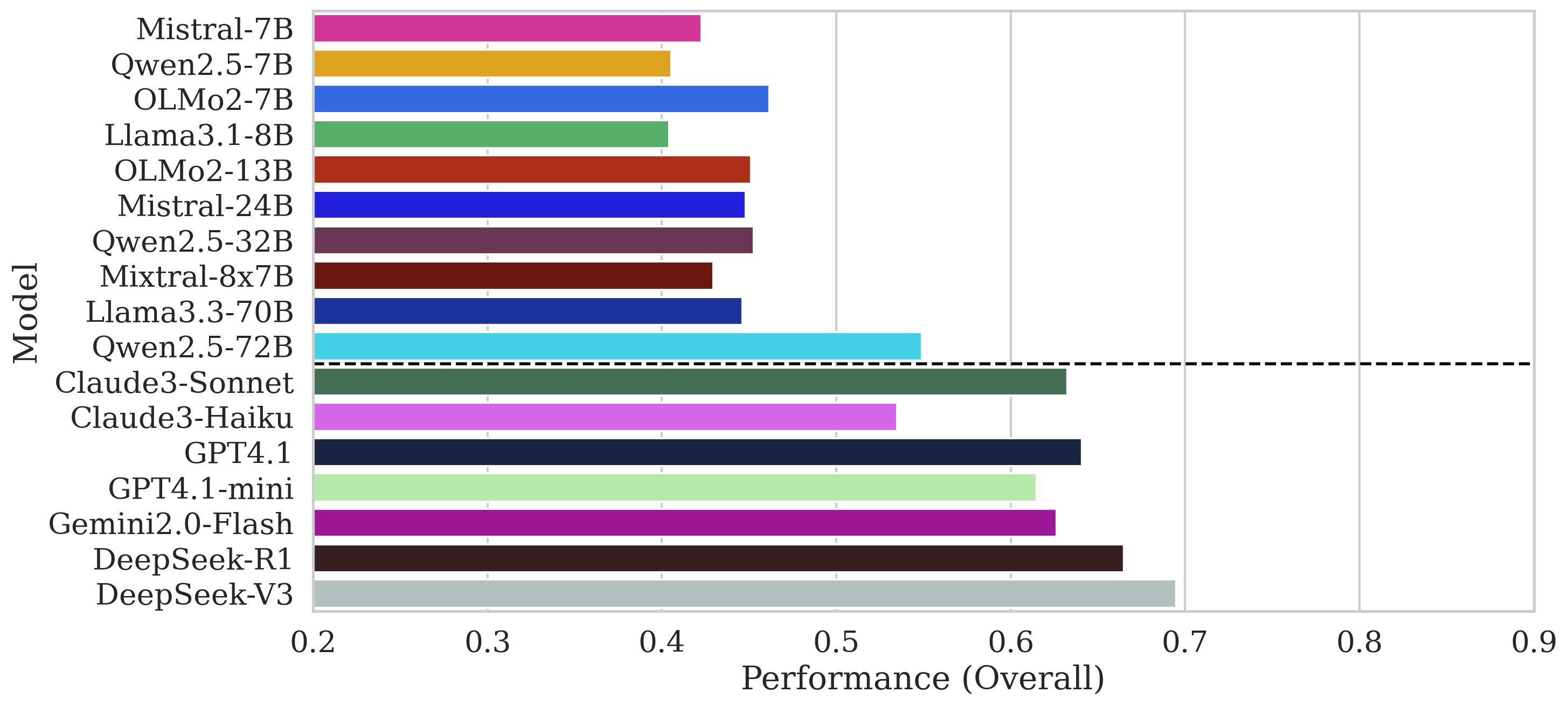}
    \caption{Overall performances.}
    \label{fig:plot-perf-overall}
\end{figure}

\newpage

\begin{figure}[H]
    \centering
    \includegraphics[width=0.75\linewidth]{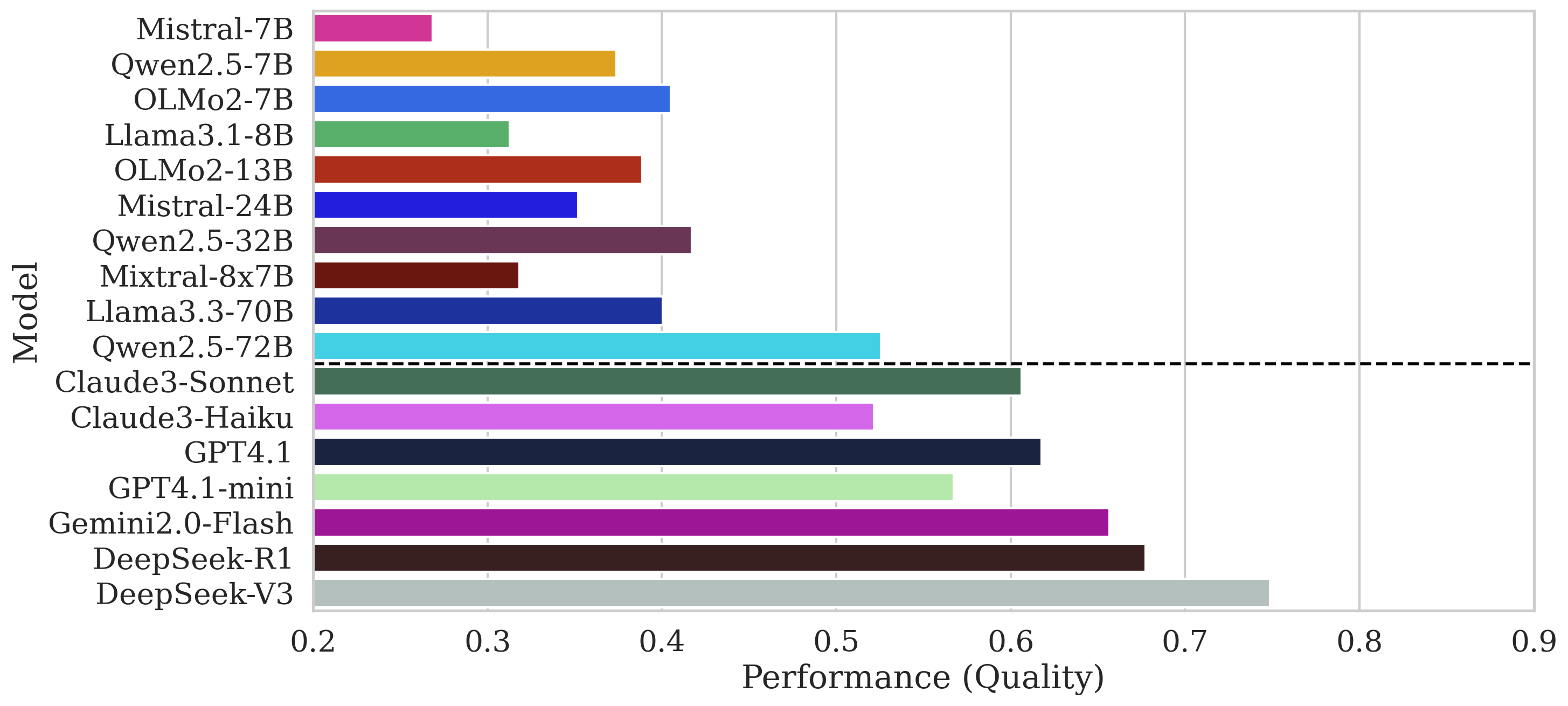}
    \caption{Performance on quality dimension}
    \label{fig:plot-perf-quality}

    \vspace{0.5em}
    \includegraphics[width=0.75\linewidth]{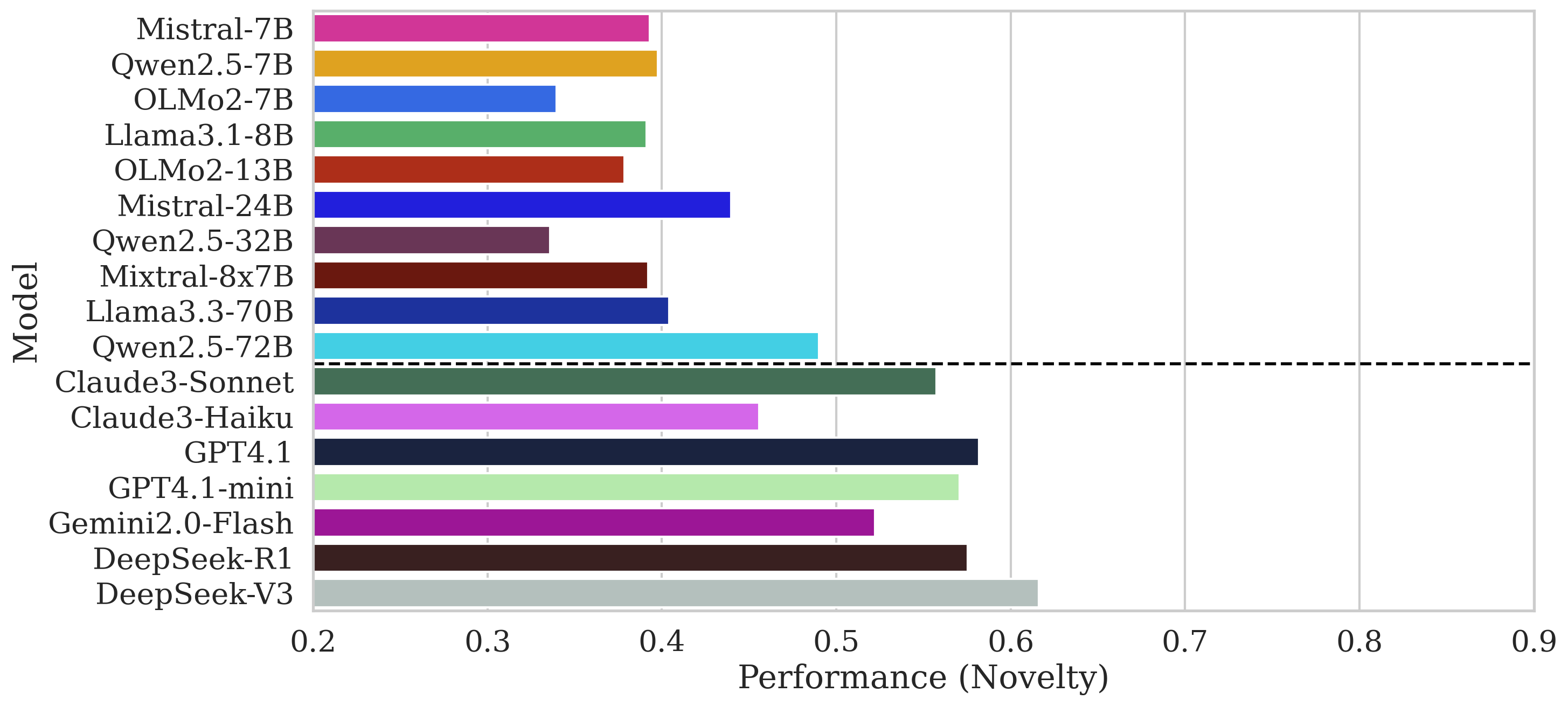}
    \caption{Performance on novelty dimension}
    \label{fig:plot-perf-novelty}

    \vspace{0.5em}
    \includegraphics[width=0.75\linewidth]{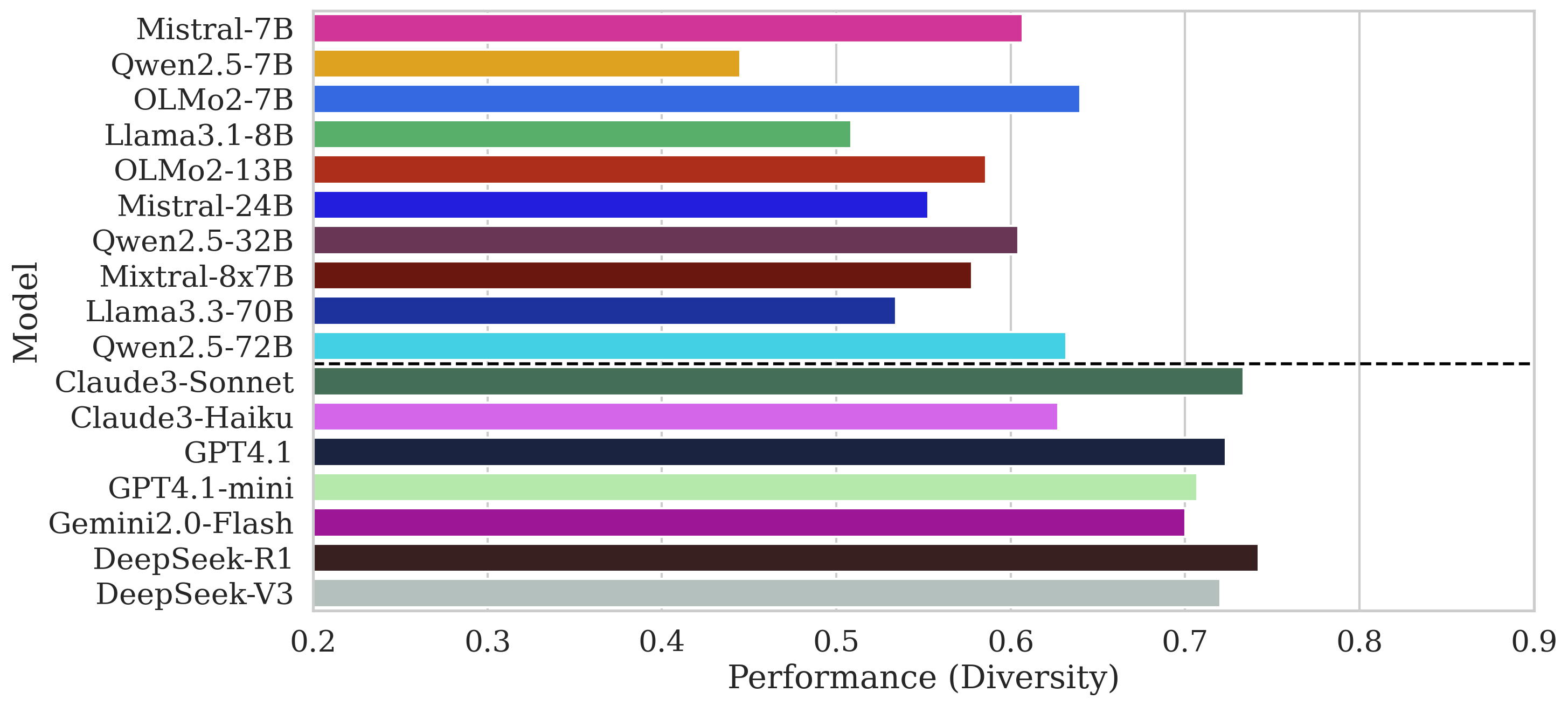}
    \caption{Performance on diversity dimension}
    \label{fig:plot-perf-diversity}
\end{figure}

\newpage

\begin{figure}[H]
    \centering
    \includegraphics[width=.75\linewidth]{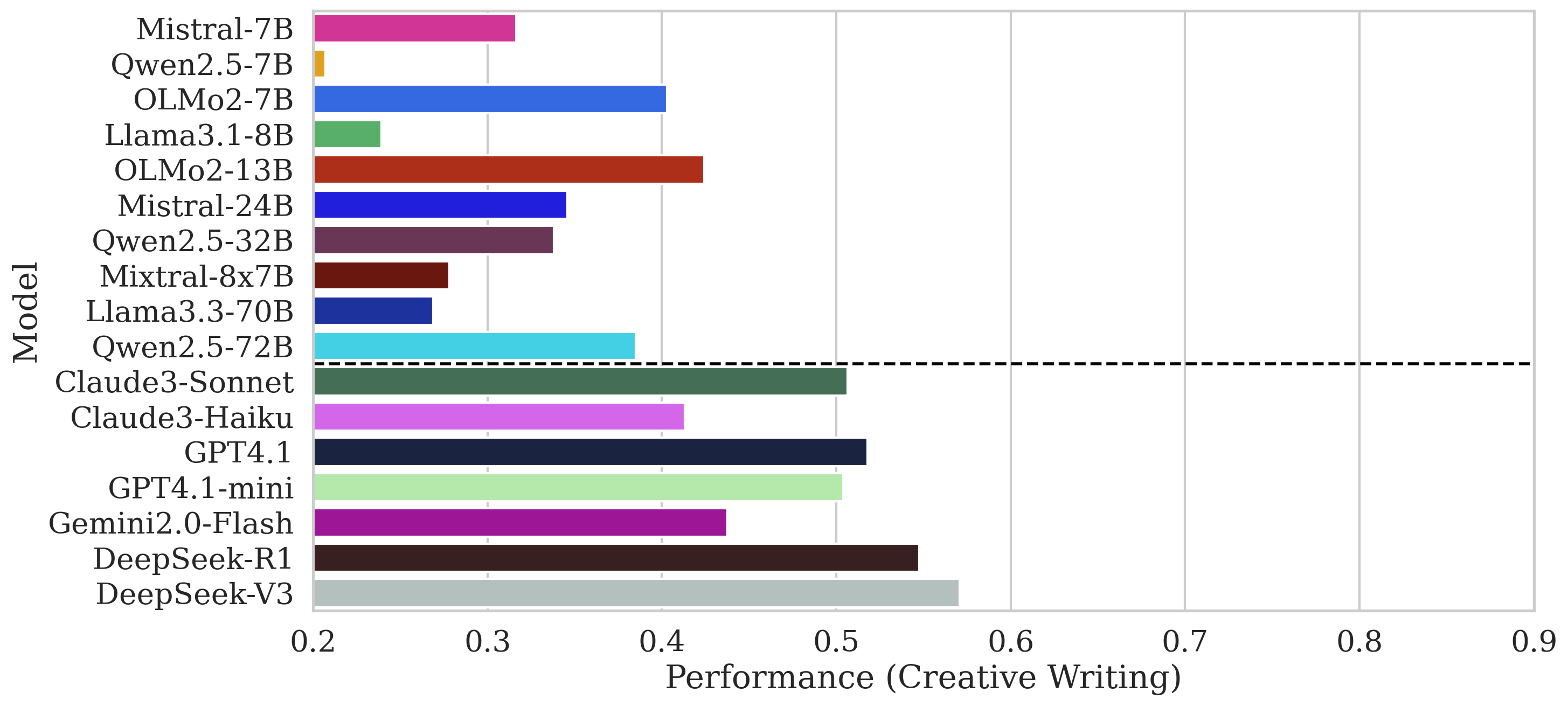}
    \caption{Performance on creative writing tasks}
    \label{fig:plot-perf-story}

    \vspace{0.5em}
    \includegraphics[width=.75\linewidth]{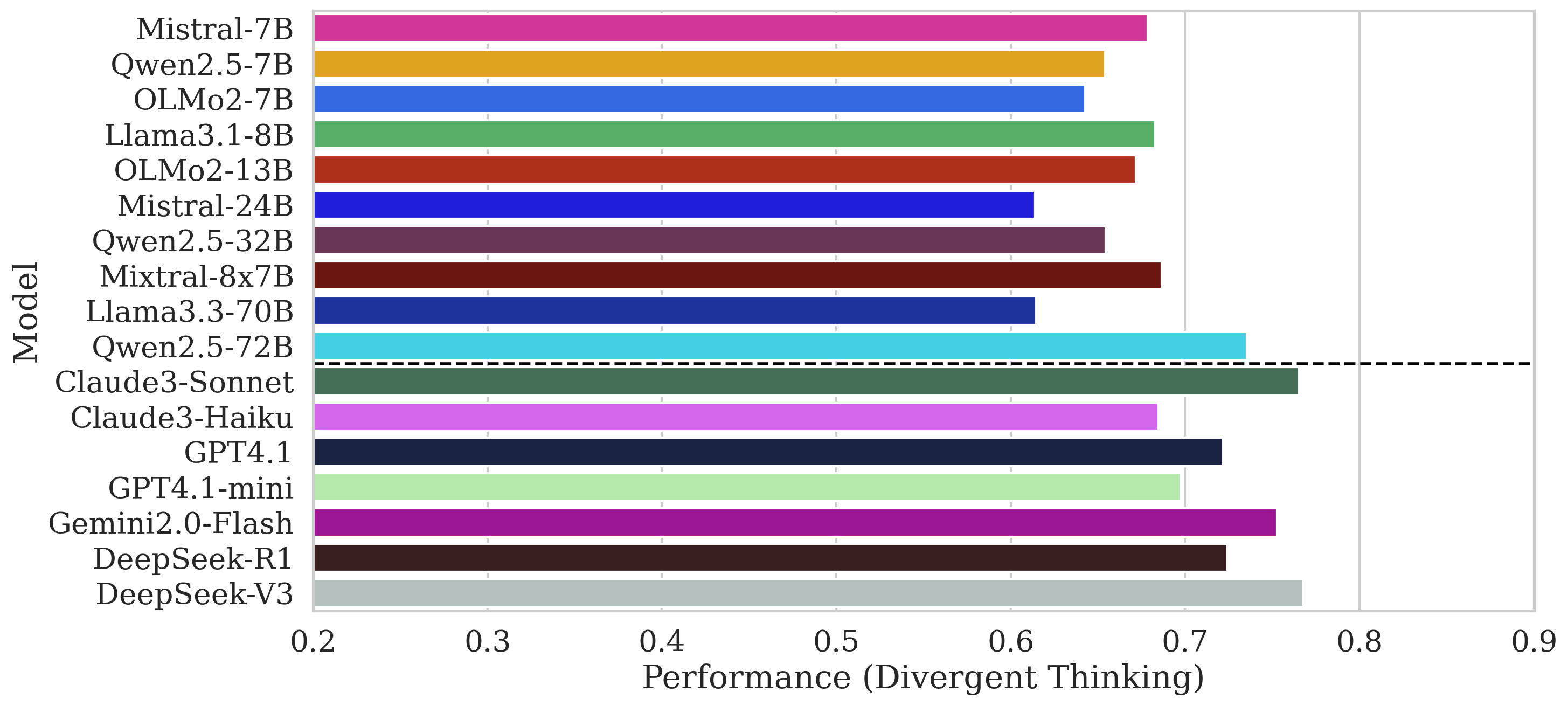}
    \caption{Performance on divergent thinking tasks}
    \label{fig:plot-perf-psyc}

    \vspace{0.5em}
    \includegraphics[width=.75\linewidth]{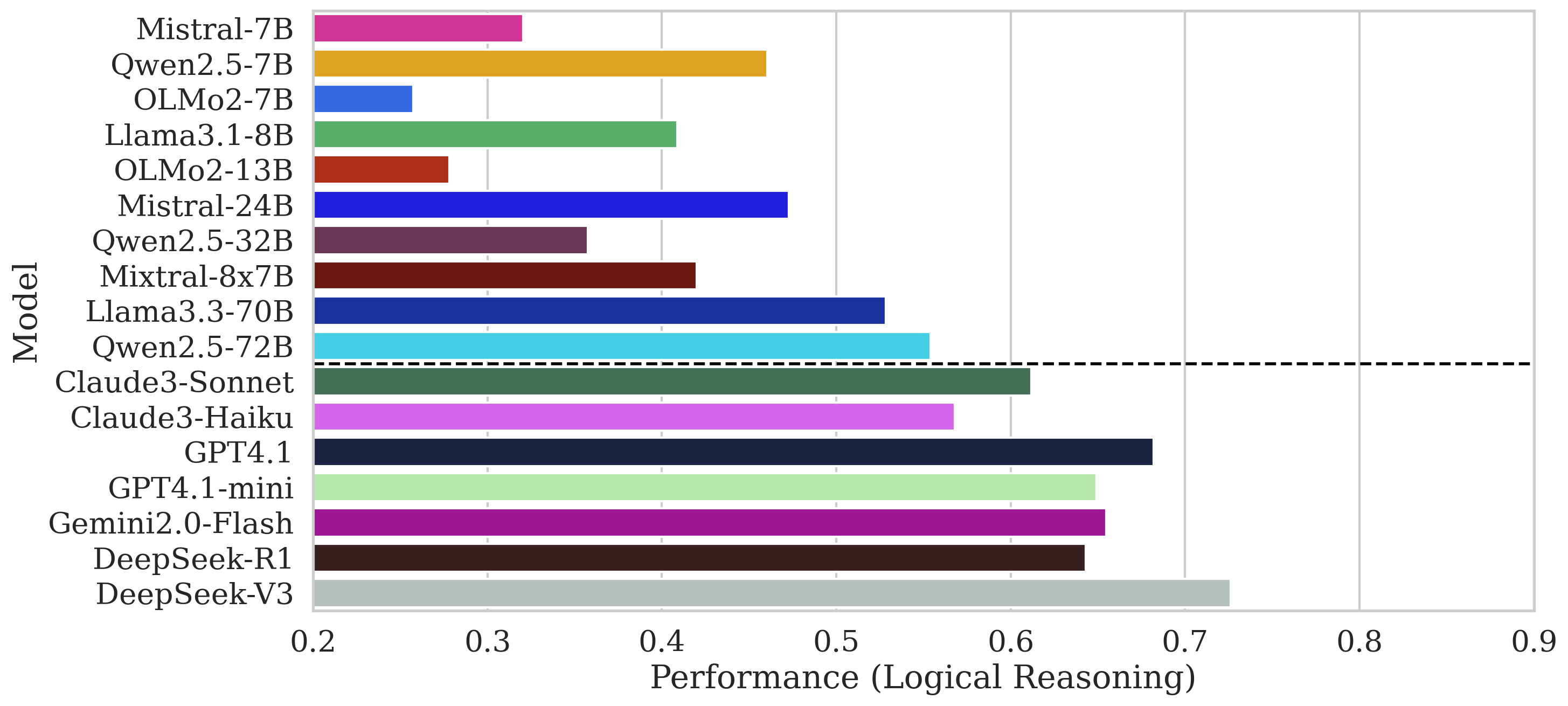}
    \caption{Performance on logical reasoning tasks}
    \label{fig:plot-perf-logical}

\end{figure}

\subsection{Performance Correlation Statistics}
\label{app:corr_stats}
\paragraph{Within-domain Performances} To show that within-domain correlation is significantly larger than cross-domain correlation, we computed Pearson correlations for all 136 pairwise combinations of the 17 evaluation metrics ($N=17$ models), classifying each pair as either within-domain, where both metrics come from the same domain ($n=44$ pairs), or between-domain, where the metrics come from different domains ($n=92$ pairs). A permutation test was used to assess whether within-domain pairs show systematically higher correlations: all 136 $r$-values were pooled and randomly reassigned to within/between groups across 10,000 shuffles, with the $p$-value defined as the proportion of shuffles yielding an absolute mean difference at least as large as the observed one. Bootstrap 95\% confidence intervals on the mean difference were computed from 10,000 resamples with seed 42. The test result shown in Table~\ref{tab:within_between_corr} provides statistical support for our claim in Section~5.2: ``We find a strong correlation in the models' performance on metrics coming from the same task.''

\begin{table}[t]
\centering
\begin{tabular}{lccc}
\toprule
Group & $n$ pairs & Mean $r$ & Median $r$ \\
\midrule
Within-domain  & 44 & .494 & .486 \\
Between-domain & 92 & .341 & .350 \\
\midrule
Difference & -- & \textbf{+.153} & -- \\
95\% CI (bootstrap) & -- & $[.057, .248]$ & -- \\
Permutation $p$ (two-tailed) & -- & $< .001$ & -- \\
\bottomrule
\end{tabular}
\caption{Within-domain correlations are significantly larger than between-domain correlations. Results are based on 10,000 permutation shuffles and 10,000 bootstrap resamples with seed 42.}
\label{tab:within_between_corr}
\end{table}

\paragraph{Performance gap Between Open-source and Frontier-scale Models}
To formally test whether the performance gap between open models and frontier-scale models varies across creativity domains, we conducted Welch's independent-samples $t$-tests on per-domain scores, using simple averages for open models ($n=10$) and frontier-scale models ($n=7$). Welch's test was used because the two groups differ in both sample size and variance. We further applied a Bonferroni correction across $k=3$ domain comparisons. Effect sizes are reported as Cohen's $d$, computed using the pooled standard deviation. Bootstrap 95\% confidence intervals on the mean gap were computed from 10,000 resamples with seed 42.

Results show a significant difference between the mean performance scores of open models and frontier-scale models in all three domains, with creative writing and logical reasoning showing notably larger gaps in terms of Cohen's $d$. The test results in Table~\ref{tab:open_prop_domain_gap} support our claim in Section~5.1: ``Among the three domains, logical reasoning and creative writing see a notably larger gap than divergent thinking.''

\begin{table}[t]
\centering
\resizebox{\textwidth}{!}{
\begin{tabular}{lccccccc}
\toprule
Domain 
& Open ($n=10$) 
& Prop. ($n=7$) 
& Gap 
& 95\% CI 
& $t$ 
& $p$ (Bonf.) 
& Cohen's $d$ \\
\midrule
Creative Writing   
& .320 & .500 & +.179 & $[.123, .236]$ & 5.747 & $<.001^{***}$ & 2.772 \\
Divergent Thinking 
& .664 & .731 & +.067 & $[.036, .097]$ & 3.956 & $.004^{**}$ & 1.930 \\
Logical Reasoning  
& .406 & .648 & +.242 & $[.174, .310]$ & 6.493 & $<.001^{***}$ & 3.023 \\
\bottomrule
\end{tabular}
}
\caption{Comparison of open models and frontier-scale models performance across creativity domains. Welch's independent-samples $t$-tests were conducted per domain, with Bonferroni correction across three comparisons. Bootstrap 95\% confidence intervals were computed from 10,000 resamples with seed 42.}
\label{tab:open_prop_domain_gap}
\end{table}

\noindent\textit{Note.} $^{**}p < .01$, $^{***}p < .001$ after Bonferroni correction ($k=3$).

\subsection{Performance Correlation Discussions}
\label{app-perf-discussions}
\paragraph{Negative correlations between NeoCoder Divergence@0 and other metrics.} NeoCoder Divergence@0 measures the proportion of coding techniques in a model's solution absent from the human solution set \citep{neocoder}. A model that fails to solve a problem correctly will, by definition, avoid using standard human techniques, not because it is creative, but because its incorrect output lacks the structural properties of valid solutions. Nonsensical outputs would score perfectly on this metric. This is distinct from other LLM-judged novelty metrics, which score the quality of a particular output rather than performing a binary classification against a reference set. The NeoCoder paper itself addresses this by introducing denial prompting to isolate genuine divergence from failure; our use of the @0 baseline captures the unconditional setting where this confound is most visible. Notably, CreativeMath avoids this issue by only counting a solution as novel after it is verified correct.

\subsection{Performance Stability Analysis}
\label{app-stability}
\begin{figure}[H]
    \centering
    \includegraphics[width=.75\linewidth]{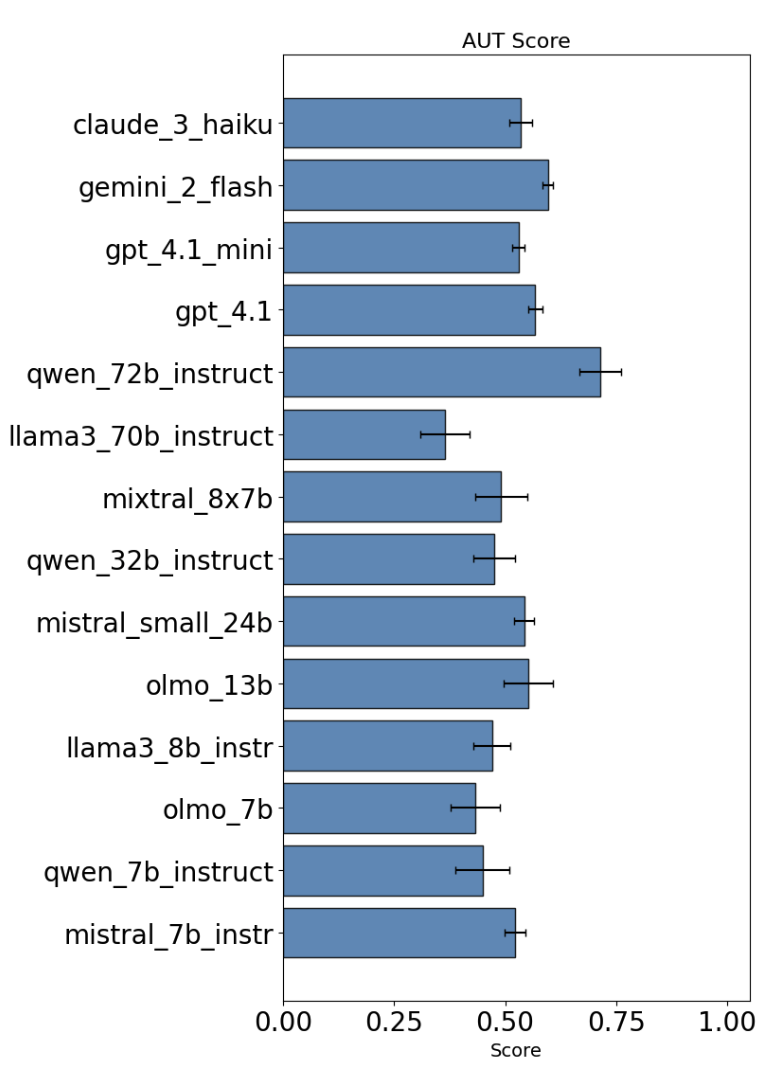}
    \caption{Multi-run results for AUT}
    \label{fig:plot-multi-aut}
    \vspace{0.5em}
\end{figure}

\begin{figure}[H]
    \centering
    \includegraphics[width=.9\linewidth]{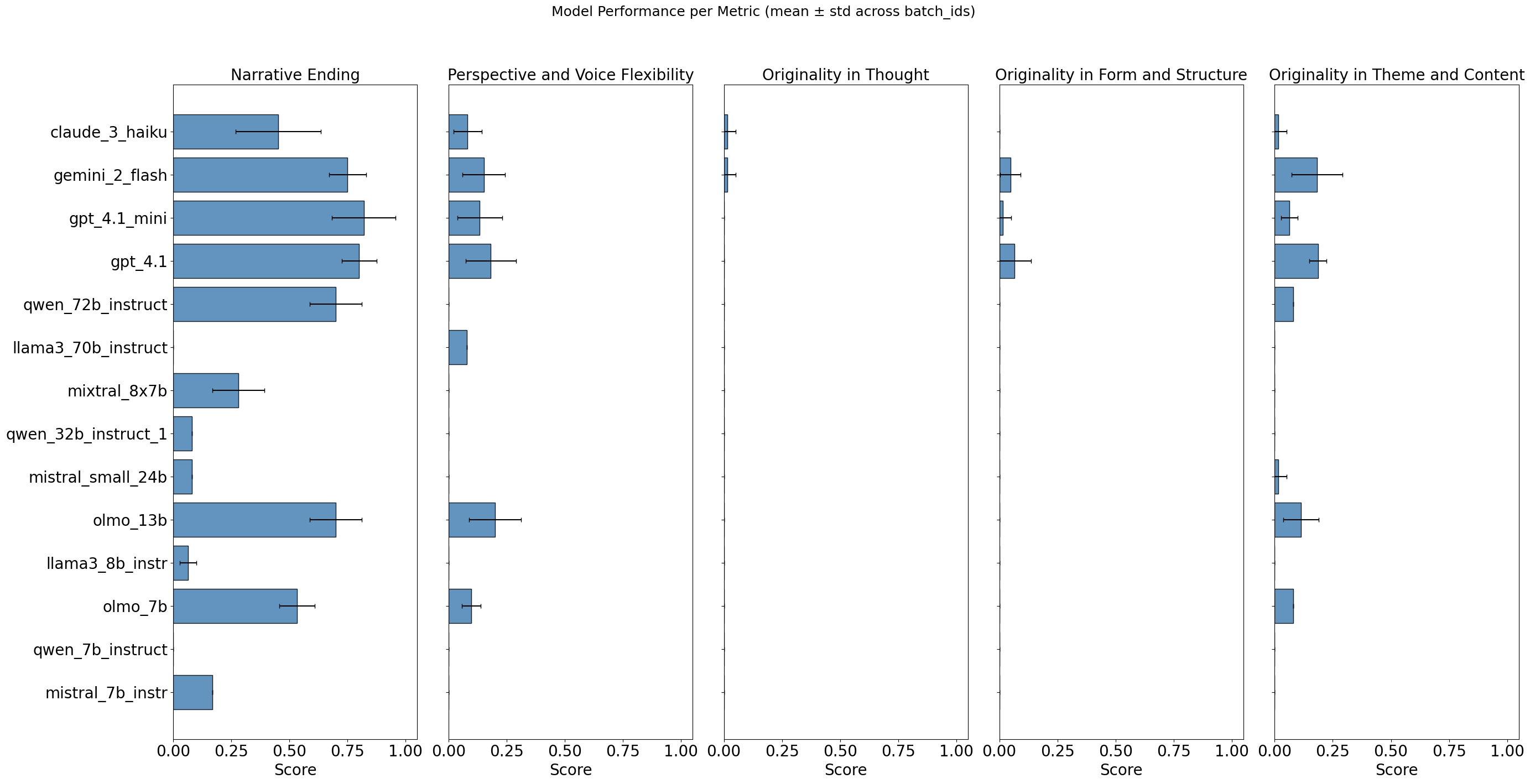}
    \caption{Multi-run results for TTCW}
    \label{fig:plot-multi-ttcw}
    \vspace{0.5em}
\end{figure}

\paragraph{Multiple Runs with the Same Setting} While we did not have the chance to run multiple experiments with the same settings, we present the following partial results, for AUT and TTCW tasks, with all models except Claude3.7-Sonnet (no API access anymore), DeepSeek Models (same reason); for each model, we run 5 independent runs with the same configuration. The performance visualizations are in Figure \ref{fig:plot-multi-aut}, \ref{fig:plot-multi-ttcw}, with error bars showing the min, max, and average of task performances. Based on these partial results, we believe the model performances are stable across 5 runs. We also want to point out that there are many other factors, e.g., prompt style, max length, etc., that could also impact the results. We will leave the analysis of those factors to future work and focus on a baseline study now. 

\begin{figure}[H]
    \centering
    \includegraphics[width=.9\linewidth]{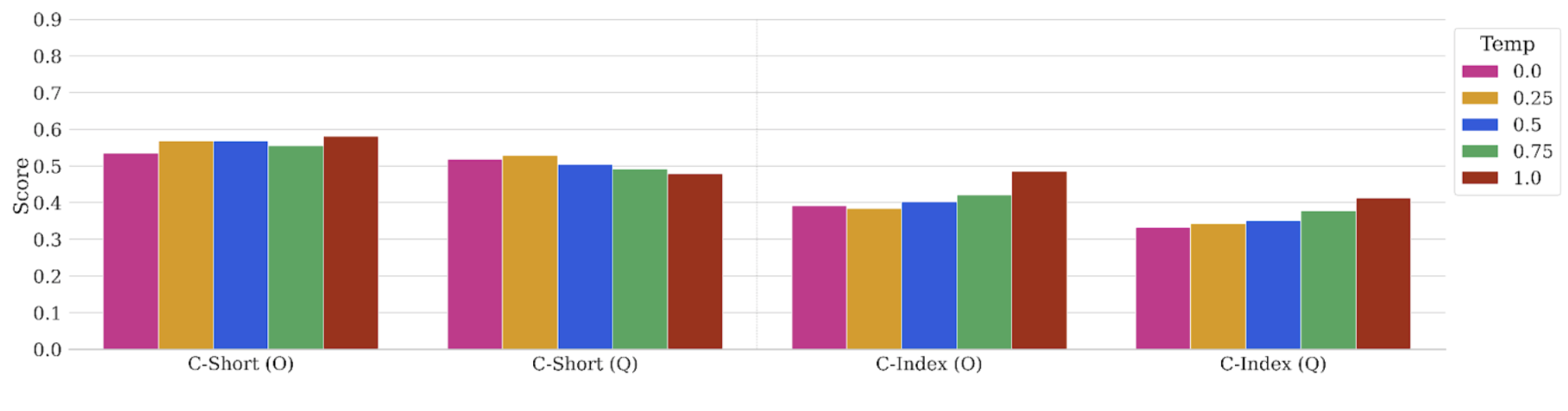}
    \caption{Performance on Creative Short Story (C-Short) and Creativity Index (C-Index) with different temperature settings; \textbf{O}: Olmo-13B, \textbf{Q}: Qwen-72B.}
    \label{fig:plot-multi-temp}
    \vspace{0.5em}
\end{figure}

\paragraph{Effect of Temperature on Results} 
\label{app-temperature}
Prior work has shown mixed conclusions when it comes to temperature versus creativity \citep{Peeperkorn2024-ig, Lu2024-io, assessing_understanding_creativity_llm}. One of our early experiments (Figure \ref{fig:plot-multi-temp}) that involves Creative Short Story and Creativity Index, with Qwen-72B and Olmo-13B, shows that temperature has little influence on the performance of Creative Short Story tasks, while models perform slightly better in Creativity Index with higher temperature settings. With such mixed results, we decided to fix the temperature for all experiments and focus on building an evaluation framework at that time. 

\begin{figure}[H]
    \centering
    \includegraphics[width=.75\linewidth]{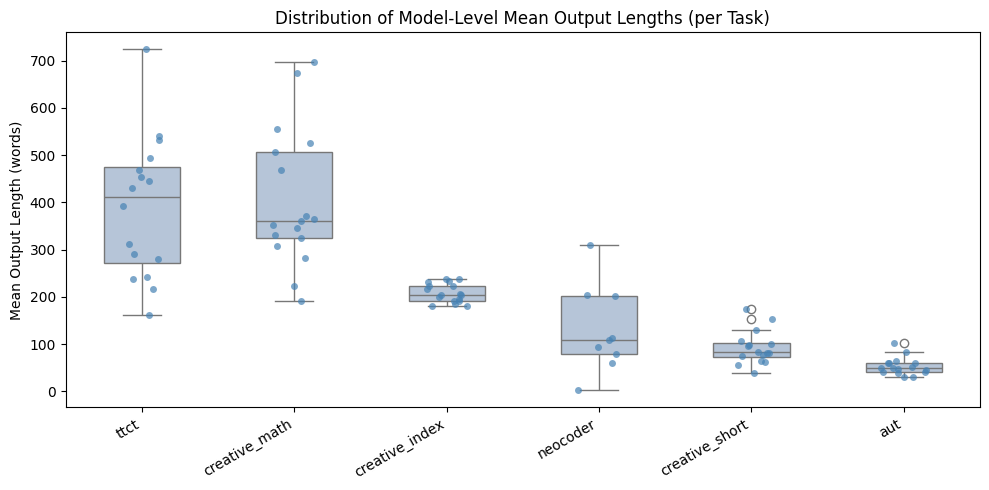}
    \caption{Output length distribution by model (part 1).}
    \label{fig:plot-length-task-1}
    \vspace{0.5em}
\end{figure}

\begin{figure}[H]
    \centering
    \includegraphics[width=.9\linewidth]{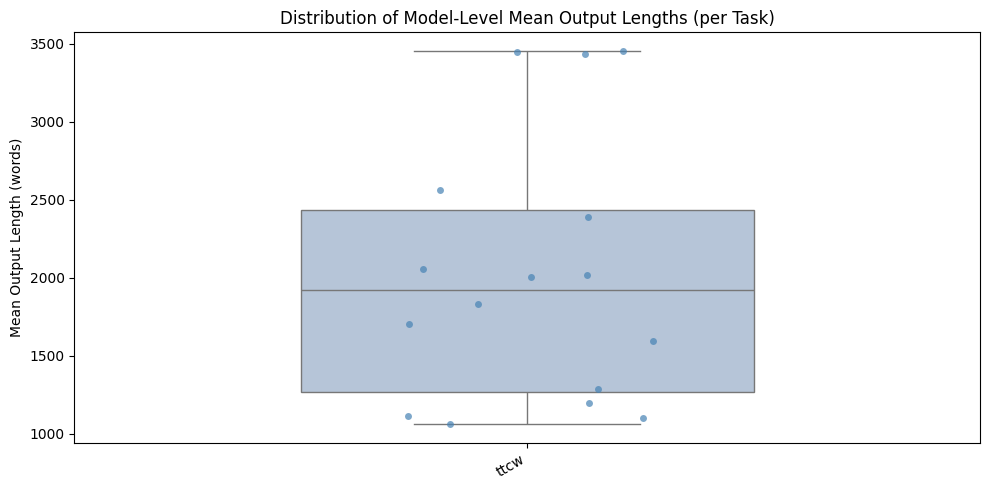}
    \caption{Output length distribution by model (part 2).}
    \label{fig:plot-length-task-2}
    \vspace{0.5em}
\end{figure}

\begin{figure}[H]
    \centering
    \includegraphics[width=.9\linewidth]{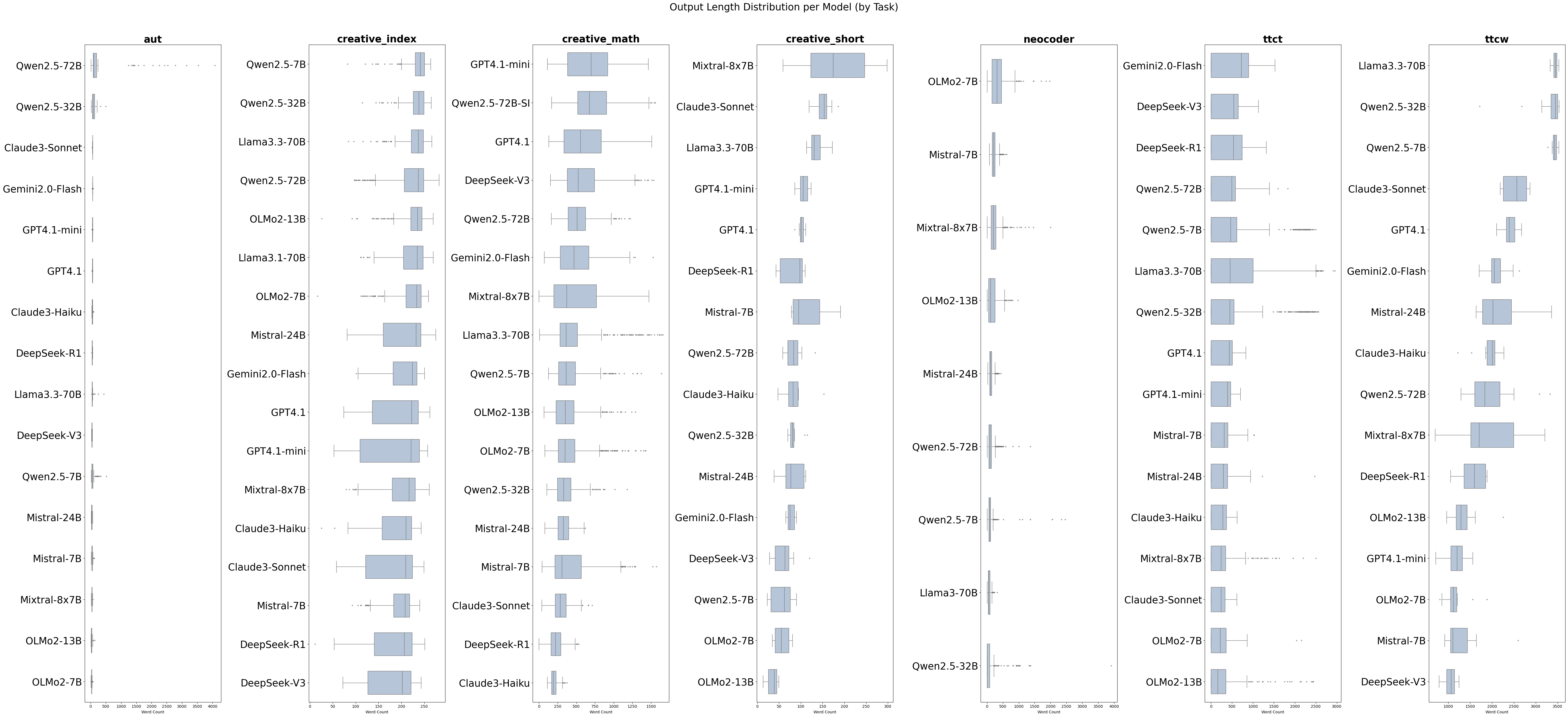}
    \caption{Output length distribution by task.}
    \label{fig:plot-length-model}
    \vspace{0.5em}
\end{figure}

\paragraph{Output Length}
\label{app-output-len}
We also study the length of generated text for all tasks to ensure the length requirements of each task are correctly met (Figure \ref{fig:plot-length-task-1}, \ref{fig:plot-length-task-2}, \ref{fig:plot-length-model}). For all tasks and all LLMs, the output max token is set to 4096 (except for CreativityIndex, which is set to 288, and CreativeMath, which is set to 2000). We believe these results confirm that 1) no model has a significant advantage compared to other tasks. 2) For tasks with length constraints (CreativityIndex, CreativeShortStory, TTCW), the length requirements are met correctly. 
In our initial experiments with TTCT task, we followed the original papers’ setup and observed that LLM-Judge show two failure modes: (1) length bias, where judges tend to assign higher scores to verbose responses, which is particularly pronounced for models that generate preambles or closing remarks (e.g., greeting messages, expressions of politeness); (2) redundancy blindness, where judges fail to penalize responses that repeat the same core idea with superficial variation. We mitigate both by preprocessing model outputs to strip formatting, politeness markers, and verbosity before evaluation (see task specific prompt in Appendix \ref{app:TTCT-eval-prompt}). For other tasks with no explicit length requirements, we impose different ways to parse the output so that they don’t get extraordinarily long, e.g., for AUT, we only keep the results from the first ten lines; for CreativeShort, we require the generation model to insert [START] and [END] label before and after the story.

\newpage

\section{Task Details}
\label{app-task-details}
\subsection{Torrance Test of Creative Writing (TTCW)}
\subsubsection{Dataset}
The dataset consists of 12 New Yorker Stories' plots, i.e., GPT-4 generated summary of the original story \footnote{\url{https://github.com/salesforce/creativity_eval}}.
\subsubsection{Example}
\begin{tcolorbox}[mystyle, title={Plot}]
A woman experiences a disorienting night in a maternity ward where she encounters other similarly disoriented new mothers, leading to an uncanny mix-up where she leaves the hospital with a baby that she realizes is not her own, yet accepts the situation with an inexplicable sense of happiness.
\end{tcolorbox}

\begin{tcolorbox}[mystyle, title={Inference Prompt}]
Write a New Yorker-style story given the plot below. Make sure it is at least \{word\_count\} words. Directly start with the story, do not say things like "Here's the story [...]"
Plot: \{plot\}
Story:
\end{tcolorbox}

\subsubsection{Experiment Configurations}
\begin{itemize}
    \item Temperature: 0.75
    \item Max Token: 4096
    \item Top P: 1
\end{itemize}

\subsubsection{Evaluation Metrics}
As mentioned in Appendix \ref{app-llm-judge}, we use a subset of questions from the original paper where experts achieved at least moderate agreement (Fleiss Kappa no less than 0.4) and our few-shot LLM-Judge passed the Alternative Annotator Test \citep{alternative-annotator-test}. Since each evaluation question is binary for each generated story, we calculate the proportion of generated stories that pass each question as the final evaluation metric (e.g., if 3 out of 12 stories pass the ``Narrative Ending (Fluency)'' question, then the ``Narrative Ending (Fluency)'' metric is 0.25).

We use two-shot examples (one positive and one negative) in the evaluation prompt, as previous work shows adding few-shot examples improves human-LLM alignments \citep{Jung2024-xz}.

\newpage
\begin{tcolorbox}[mystyle, title={Evaluation Prompt}]
You are given a creative short story. Read it carefully. You are then given some background about specific aspects of creative writing, a binary (Yes/No) question, and sample stories with expert-annotated answers to the same question. Your objective is to use the background information and sample stories to answer the question about the story. Provide your answer in the format of "**Answer**: [Yes/No]". You can optionally then provide a short explanation for your answer.\\
==========\\
Question:\\
\{full\_prompt\}\\

Examples:\\
==========\\
Story: \{story\}\\
Answer: \{answer\}\\
Explanations: \{exp\}\\
==========\\
Story: \{story\}\\
Answer: \{answer\}\\
Explanations: \{exp\}\\

==========\\
Story: \{story\}\\
Based on the question and examples above, answer the question (Provide your answer in the format of "**Answer**: [Yes/No]". You can optionally then provide a short explanation for your answer). Make sure you are extra harsh on the decision (most answers should be negative).\\
Answer:
\end{tcolorbox}

\begin{quote}

\end{quote}

\subsubsection{Model Performances}
\begin{table}[htbp]
    \small
    \centering
    \begin{tabular}{llllll}
        \toprule
        \textbf{Model} & 
        \textbf{\makecell{Narrative\\Ending\\(Fluency)}} & 
        \textbf{\makecell{Perspective\\Voice\\(Flexibility)}} & 
        \textbf{\makecell{Thought\\(Originality)}} & 
        \textbf{\makecell{Form\\Structure\\(Originality)}} & 
        \textbf{\makecell{Theme\\Content\\(Originality)}} \\
        \midrule
        Mistral-7B      & 0.17 & 0.00 & 0.08 & 0.00 & 0.00 \\
        Qwen2.5-7B      & 0.00 & 0.00 & 0.17 & 0.00 & 0.00 \\
        OLMo2-7B        & \textbf{0.67} & \textbf{0.17} & \textbf{0.25} & \textbf{0.08} & \textbf{0.08} \\
        Llama3.1-8B     & 0.00 & 0.00 & 0.08 & 0.00 & 0.00 \\
        \midrule
        OLMo2-13B       & \textbf{0.50} & 0.00 & \textbf{0.33} & \textbf{0.08} & \textbf{0.25} \\
        Mistral-24B     & 0.08 & 0.00 & 0.25 & 0.00 & \textbf{0.08} \\
        Qwen2.5-32B     & 0.08 & 0.00 & 0.17 & 0.00 & 0.00 \\
        \midrule
        Mixtral-8x7B    & 0.08 & 0.00 & 0.08 & 0.00 & 0.00 \\
        Llama3.3-70B    & 0.00 & \textbf{0.08} & 0.33 & 0.00 & 0.00 \\
        Qwen2.5-72B     & \textbf{0.50} & 0.00 & \textbf{0.50} & \textbf{0.17} & \textbf{0.08} \\
        \midrule
        Claude3-Sonnet  & 0.75 & \textbf{0.33} & 0.58 & 0.42 & \textbf{0.33} \\
        Claude3-Haiku   & 0.50 & 0.00 & 0.25 & 0.08 & 0.00 \\
        GPT-4.1         & 0.75 & 0.25 & {0.67} & 0.50 & {0.17} \\
        GPT-4.1-mini    & 0.67 & 0.25 & \textbf{0.83} & 0.50 & 0.08 \\
        Gemini2.0-Flash & 0.83 & 0.17 & 0.42 & 0.17 & 0.08 \\
        DeepSeek-R1     & {0.92} & \textbf{0.33} & 0.50 & \textbf{0.58} & {0.17} \\
        DeepSeek-V3     & \textbf{1.00} & 0.17 & 0.50 & {0.50} & 0.08 \\
        \bottomrule
    \end{tabular}
    \vspace{0.5em}
    \caption{Model performance on TTCW; \textbf{bold} are top performer in each model size group.}
\end{table}

\newpage
\subsection{Creativity Index}
\subsubsection{Dataset}
The dataset consists of 3 subsets: book, poem, and speech, all are the prefixes (i.e., first line of text) from the dataset proposed by ~\citet{creative_index}. We use the first 100 examples in generation and evaluation. \footnote{\url{https://github.com/GXimingLu/creativity_index}}

\subsubsection{Examples}
Here are some examples of the input data (i.e., the text prefix for LLM to complete).
\begin{tcolorbox}[mystyle, title={Book}]
It's been years: Bailey clearly means him no harm and has managed to be discreet enough that Nick's queerness isn't the talk of the Chronicle.
\end{tcolorbox}

\begin{tcolorbox}[mystyle, title={Poem}]
Swiftly walk o'er the western wave,
\end{tcolorbox}

\begin{tcolorbox}[mystyle, title={Speech}]
That is the kind of America in which I believe
\end{tcolorbox}

\subsubsection{Evaluation Metrics}
We follow the evaluation metrics outlined in \citet{Lu2024-ma}, specifically retaining the exact match component. However, we exclude the semantic search-based evaluation due to its high computational cost and sensitivity to the chosen cosine similarity threshold, which significantly affects whether two sentence spans are considered semantically similar. We sum over the L-Uniqueness with spans of n-grams from 5 to 12 inclusively to get the total creative index for each response. We average the creative index for each response per mode per task. Data cleaning was done before the evaluation manually to remove irrelevant outputs. Then, we normalize the score by dividing it with 8 (the highest value that the summation could be) to get the final Creativity Index measurement for each model over the three different tasks. 

\paragraph{L-Uniqueness} Let $\mathbf{x}$ be a text consisting of a sequence of words whose linguistic creativity we wish to quantify. Let an $n$-gram of $\mathbf{x}$ be any contiguous subsequence of $n$ words, and denote by $\mathbf{x}_{i:i+n}$ the $n$-gram starting at the $i$-th word of $\mathbf{x}$. Let $C$ be a large reference corpus of publicly available texts, and define $f$ as a binary function that returns $1$ if the $n$-gram $\mathbf{x}_{i:i+n}$ occurs anywhere in $C$, and $0$ otherwise. The \textit{L}-uniqueness of $\mathbf{x}$, denoted by $\text{uniq}(\mathbf{x}, L)$, is defined as the proportion of words $w \in \mathbf{x}$ such that none of the $n$-grams containing $w$ with $n \geq L$ occur in $C$. Intuitively, a higher \textit{L}-uniqueness means a greater proportion of $\mathbf{x}$'s words appear in novel contexts not seen in the reference corpus, thus indicating greater linguistic originality.

\subsubsection{Experiment Configurations}
\begin{itemize}
    \item Temperature: 1
    \item Max Token: 288
    \item Top P: 0.9
\end{itemize}
\subsubsection{Inference Prompt}

\begin{tcolorbox}[mystyle, title={Book}]
Please write a few paragraphs for a novel starting with the following prompt: \{start sentence of book\}
\end{tcolorbox}

\begin{tcolorbox}[mystyle, title={Poem}]
Please write a few paragraphs for a poem starting with the following prompt: \{start sentence of poem\}
\end{tcolorbox}

\begin{tcolorbox}[mystyle, title={Speech}]
Please write a speech starting with the following sentence: \{start sentence of speech\}
\end{tcolorbox}

For closed-source models, we included an additional prompt instruction to ensure that the output consists solely of the completed paragraphs, poems, or speeches. These models often preface their responses with phrases like “Certainly” or “Here is...,” which we manually removed during post-processing. In contrast, open-source models typically generate the desired completions directly without such prefatory text. For these models, we reviewed the log outputs and removed any unrelated content as needed.


\subsubsection{Model Performance}
See Table \ref{tab:creative-domains-average}.

\begin{table}[t]
\centering
\footnotesize
\begin{tabular}{lcccc}
\toprule
\textbf{Model} & \textbf{Book} & \textbf{Poem} & \textbf{Speech} & \textbf{Average} \\
\midrule
Mistral-7B     & 0.4496 & 0.5828 & 0.3104 & 0.4476 \\
Qwen2.5-7B        & 0.4354 & 0.6310 & 0.3534 & 0.4733 \\
OLMo-7B       & 0.4810 & 0.6110 & 0.3727 & 0.4882 \\
Llama-3.1-8B    & 0.4724 & 0.5700 & 0.3396 & 0.4607 \\
\midrule
OLMo-13B      & 0.4860 & 0.5963 & 0.3522 & 0.4782 \\
Mistral-24B    & 0.4752 & 0.6646 & 0.3397 & 0.4932 \\
Qwen2-32B       & 0.4663 & 0.6328 & 0.3465 & 0.4816 \\
\midrule
Mistral-8x7B   & 0.4149 & 0.6035 & 0.2804 & 0.4329 \\
Llama-3.3-70B   & 0.4226 & 0.5802 & 0.2936 & 0.4321 \\
Qwen2.5-72B       & 0.4133 & 0.5924 & 0.3171 & 0.4409 \\
\midrule
Claude3-Sonnet & 0.5615 & 0.6700 & 0.4675 & 0.5663 \\
Claude3-Haiku  & 0.5769 & 0.7039 & 0.4519 & 0.5776 \\
GPT4.1                  & 0.6044 & 0.7637 & 0.4593 & 0.6091 \\
GPT4.1-mini             & 0.5624 & 0.7147 & 0.4261 & 0.5677 \\
Gemini2.0-Flash         & 0.5278 & 0.6707 & 0.4121 & 0.5369 \\
DeepSeek-R1        & 0.5930 & 0.7595 & 0.5410 & 0.6312 \\
DeepSeek-V3            & \textbf{0.6814} & \textbf{0.7791} & \textbf{0.6166} & \textbf{0.6924} \\
\bottomrule
\end{tabular}
\vspace{0.5em}
\caption{L-Uniqueness across Book, Poem, Speech, and averaged performance for different models; we use average as the L-uniqueness score in \ours{} as the metric for Creativity Index; \textbf{bold} numbers are best performers.}
\label{tab:creative-domains-average}
\end{table}

\subsubsection{Additional Comments}
We also note that the generation for OLMo2-13B-instruct may miss some data with the vllm generation. We remove those missing generations. This accounts for 13 responses in the poem subset and 10 examples in the speech subset. In addition, the model may refuse to answer some prompts. We also removed those generations. For OLMo-7B-instruct, there are 2 cases in the speech subset. For GPT-4.1, there is 1 case in the speech subset.


\newpage
\subsection{Creative Short Story}

\subsubsection{Dataset}
The dataset consists of 10 three-word tuples. For any given LLM, it is prompted to generate a short story (at most five sentences) based on those three words \footnote{\url{https://github.com/mismayil/creative-story-gen}}.

\subsubsection{Examples}
\begin{tcolorbox}[mystyle, title={Three-word Tuple}]
stamp, letter, send
\end{tcolorbox}

\subsubsection{Experiment Configurations}
\begin{itemize}
    \item Temperature: 0.75
    \item Max Token: 4096
    \item Top P: 1
\end{itemize}

\subsubsection{Inference Prompt}

\begin{tcolorbox}[mystyle, title={Inference Prompt}]
You will be given three words (e.g., car, wheel, drive) and then asked to write a creative short story that contains these three words. The idea is that instead of writing a standard story, such as "I went for a drive in my car with my hands on the steering wheel.", you need to come up with a novel and unique story that uses the required words in unconventional ways or settings. Also make sure you use at most five sentences. 
The given three words: \{items\} (the story should not be about \{boring\_theme\}). 
\end{tcolorbox}

\subsubsection{Evaluation Metrics}
We included novelty score, surprise-ness, and average N-gram Diversity from the original paper. Particularly, because n-gram diversity is almost always 1 for n greater than 3 (mainly because the stories are at most five sentences long), we keep only unigram and bigram (i.e., we use the average of unigram diversity and bigram diversity as the N-gram diversity).

\subsubsection{Model Performance}
See Table \ref{tab:creative-short}.
\begin{table}[ht]
\centering
\footnotesize
\begin{tabular}{lcc}
\toprule
\textbf{Model} & \textbf{Surprise} & \textbf{N-gram Diversity} \\
\midrule
Mistral-7B       & 0.0889 & 0.810  \\
Qwen2.5-7B       & 0.0834 & 0.220  \\
OLMo2-7B         & 0.0599 & 0.895  \\
Llama3.1-8B      & 0.0490 & 0.410  \\
\midrule
OLMo2-13B        & {0.2043} & {0.905} \\
Mistral-24B      & 0.1406 & 0.820  \\
Qwen2.5-32B      & 0.1263 & 0.870  \\
\midrule
Mixtral-8x7B     & 0.0601 & 0.715  \\
Llama3.3-70B     & 0.0590 & 0.545  \\
Qwen2.5-72B      & 0.1234 & 0.860  \\
\midrule
Claude3-Sonnet   & 0.0927 & 0.860  \\
Claude3-Haiku    & 0.1235 & 0.870  \\
GPT4.1           & 0.0928 & 0.870  \\
GPT4.1-mini      & 0.0965 & 0.870  \\
Gemini2.0-Flash  & 0.0375 & 0.865  \\
DeepSeek-R1      & 0.1953 & {0.905}  \\
DeepSeek-V3      & {0.2613} & 0.900  \\
\bottomrule
\end{tabular}
\vspace{0.5em}
\caption{Performance on the Creative Short Story task, including surprise and average n-gram diversity.}
\label{tab:creative-short}
\end{table}

\newpage
\subsection{NeoCoder}
\begin{table}[ht]
\footnotesize
\centering
\begin{tabular}{lcc}
\toprule
\textbf{Model} & \textbf{Convergence@5} & \textbf{Divergence@5}  \\
\midrule
Mistral-7B                           &  0.0000	& \textbf{1.0000} \\
Qwen2.5-7B    & 0.0000 &	0.9158 \\
OLMo2-7B            & 0.0000	& 0.5773\\
Llama3.1-8B           &  0.0000 &	0.9845 \\
\midrule
OLMo2-13B            &   0.0000	& 0.4433 \\
Mistral-24B           &  0.0000	& 0.9897 \\
Qwen2.5-32B            &  0.0000	& 0.3402 \\
\midrule
Mixtral-8x7B           &  0.0000	&0.9897 \\
Llama3.3-70B           & 0.0000	& \textbf{1.0000} \\
Qwen2.5-72B            &  0.0000	& 0.7938 \\
\midrule
Claude3-Sonnet          & 0.0000	& 0.732 \\
Claude3-Haiku           & \textbf{0.0105}	& 0.9947 \\
GPT4.1                  & 0.0000	& \textbf{1.0000} \\
GPT4.1-mini             & 0.0000	& 0.9948 \\
Gemini2.0-Flash         & 0.0103	& \textbf{1.0000} \\
DeepSeek-R1             & 0.0000	& 0.732 \\
DeepSeek-V3             & 0.0103	&\textbf{1.0000}\\
\bottomrule
\end{tabular}
\vspace{0.5em}
\caption{Benchmarking results on NeoCoder \citep{neocoder} at state 5 (i.e., with 5 constraints); \textbf{bold} numbers are best performers.}
\label{tab:neocoder_model_performance}
\end{table}

\subsubsection{Examples}
We use the same dataset from the original NeoCoder paper\footnote{\url{https://github.com/JHU-CLSP/NeoCoder/}}. See Table \ref{table: examples of neocoder dataset} for examples.
\begin{table}
    \footnotesize
    \centering
    \resizebox{.99\textwidth}{!}{
    \begin{tabular}{p{0.05\textwidth}p{0.15\textwidth}p{0.8\textwidth}}
    \toprule
    \begin{tabular}{p{0.05\textwidth}}\textbf{State}\end{tabular} & \begin{tabular}{p{0.1\textwidth}}\textbf{Constraint}\end{tabular}& \begin{tabular}{p{0.85\textwidth}}\textbf{Problem Statement} \end{tabular}\\\midrule
    
    \begin{tabular}{p{0.05\textwidth}}0\end{tabular} & 
    \begin{tabular}{p{0.15\textwidth}}
    N/A \end{tabular} & 
    
    \begin{tabular}{p{0.8\textwidth}} 
    B. Points and Minimum Distance \\
    You are given a sequence of integers a of length 2n. You have to split these 2n integers into n pairs; each pair will represent the coordinates of a point on a plane. Each number from the sequence a should become the x or y coordinate of exactly one point. Note that some points can be equal
    $\cdots$
    \end{tabular} \\&&\\
    \cdashline{1-3}
    \begin{tabular}{p{0.05\textwidth}}1\end{tabular} & 
    \begin{tabular}{p{0.15\textwidth}}
     for loop \end{tabular} & 
    \begin{tabular}{p{0.8\textwidth}} \\ B. Points and Minimum Distance \\
    \textbf{Programming constraints: DO NOT use the following techniques} \\
    \textbf{- for loop}\\
    You are given a sequence of integers a of length 2n. You have to split these 2n integers into n pairs; each pair will represent the coordinates of a point on a plane. Each number from the sequence a should become the x or y coordinate of exactly one point. Note that some points can be equal
    $\cdots$ 
    \end{tabular} \\&&\\
    \cdashline{1-3}
    \begin{tabular}{p{0.05\textwidth}}2\end{tabular} & 
    \begin{tabular}{p{0.15\textwidth}}
     for loop \\ if statement \end{tabular} & 
    
    \begin{tabular}{p{0.8\textwidth}} \\ B. Points and Minimum Distance \\
    \textbf{Programming constraints: DO NOT use the following techniques} \\
    \textbf{- if statement} \\
    \textbf{- for loop}\\
    You are given a sequence of integers a of length 2n. You have to split these 2n integers into n pairs; each pair will represent the coordinates of a point on a plane. Each number from the sequence a should become the x or y coordinate of exactly one point. Note that some points can be equal $\cdots$ 
    \end{tabular} \\&&\\
    \cdashline{1-3}
    \begin{tabular}{p{0.05\textwidth}}3\end{tabular} & 
    \begin{tabular}{p{0.15\textwidth}}
     for loop \\ if statement \\ while loop \end{tabular} & 
    
    \begin{tabular}{p{0.8\textwidth}} \\ B. Points and Minimum Distance \\
    \textbf{Programming constraints: DO NOT use the following techniques} \\
    \textbf{- while loop} \\
    \textbf{- if statement} \\
    \textbf{- for loop}\\
    You are given a sequence of integers a of length 2n. You have to split these 2n integers into n pairs; each pair will represent the coordinates of a point on a plane. Each number from the sequence a should become the x or y coordinate of exactly one point. Note that some points can be equal $\cdots$ 
    \end{tabular} \\&&\\
    \cdashline{1-3}
    \begin{tabular}{p{0.05\textwidth}}4\end{tabular} & 
    \begin{tabular}{p{0.15\textwidth}}
     for loop \\ if statement \\ while loop \\ sorting \end{tabular} & 
    
    \begin{tabular}{p{0.8\textwidth}} \\ B. Points and Minimum Distance \\
    \textbf{Programming constraints: DO NOT use the following techniques} \\
    \textbf{- sorting} \\
    \textbf{- while loop} \\
    \textbf{- if statement} \\
    \textbf{- for loop}\\
    You are given a sequence of integers a of length 2n. You have to split these 2n integers into n pairs; each pair will represent the coordinates of a point on a plane. Each number from the sequence a should become the x or y coordinate of exactly one point. Note that some points can be equal $\cdots$ 
    \end{tabular} \\&&\\
    \cdashline{1-3}
    \begin{tabular}{p{0.05\textwidth}}5\end{tabular} & 
    \begin{tabular}{p{0.15\textwidth}}
     for loop \\ if statement \\ while loop \\ sorting \\ tuple \end{tabular} & 
    
    \begin{tabular}{p{0.8\textwidth}} \\ B. Points and Minimum Distance \\
    \textbf{Programming constraints: DO NOT use the following techniques} \\
    \textbf{- tuple} \\
    \textbf{- sorting} \\
    \textbf{- while loop} \\
    \textbf{- if statement} \\
    \textbf{- for loop}\\
    You are given a sequence of integers a of length 2n. You have to split these 2n integers into n pairs; each pair will represent the coordinates of a point on a plane. Each number from the sequence a should become the x or y coordinate of exactly one point. Note that some points can be equal $\cdots$ 
    \end{tabular}
    
    \\ \bottomrule
    \end{tabular}
    }
    \vspace{0.5em}
    \caption{An example of NeoCoder dataset with problem ID \href{https://codeforces.com/problemset/problem/1895/B}{1895B} and state $t=5$.}
    \label{table: examples of neocoder dataset}
\end{table}

\subsubsection{Evaluation Metrics}
\paragraph{Convergence Score} The NeoGauge metric (accompanied by the NeoCoder dataset) evaluates convergent creativity by checking whether the generated code solutions successfully pass all test cases and adhere to the given constraints.

\paragraph{Divergent Score} The NeoGauge metric (accompanied by the NeoCoder dataset) evaluates divergent creativity by comparing LLM-generated solutions to historical human solutions at the technique level. Specifically, it quantifies the proportion of novel techniques employed by the model to solve a given problem that any human has not previously used.

\subsubsection{Experiment Configurations}
We follow the experimental settings from the original NeoCoder \citep{neocoder}, including the technique detection model choice. 
\begin{itemize}
    \item Temperature: 0.75
    \item Max Token: 4096
    \item Top P: 1
\end{itemize}

\subsubsection{Model Performance}
See Table \ref{tab:neocoder_model_performance} for model performances.

\newpage
\subsection{Creative Math}
\subsubsection{Dataset}
The dataset consists of 400 math competition questions from the paper~\citet{assessing_creativity_llms_math}, with corresponding ground truth solutions extracted from Art of Problem Solving. We manually filtered out 27 questions from the IMO competition that are proof-based to make the LLM judgment more reliable \footnote{\url{https://github.com/JunyiYe/CreativeMath}}, leading to 373 questions in total.

\subsubsection{Example}

\begin{tcolorbox}[mystyle, title={2016 AMC 8, Problem 15}]
\textbf{Question:}\\
What is the largest power of $2$ that is a divisor of $13^4 - 11^4$?

\[
\textbf{(A)}\ 8 \qquad \textbf{(B)}\ 16 \qquad \textbf{(C)}\ 32 \qquad \textbf{(D)}\ 64 \qquad \textbf{(E)}\ 128
\]
\\
\textbf{Refernece Solutions 1:} \\
We use the identity for the difference of squares:

\[
13^4 - 11^4 = (13^2)^2 - (11^2)^2 = (13^2 + 11^2)(13^2 - 11^2)
\]

Calculate each part:

\[
13^2 = 169, \quad 11^2 = 121 \\
13^2 + 11^2 = 169 + 121 = 290, \quad 13^2 - 11^2 = 169 - 121 = 48
\]

We factor $48$ further:

\[
13^4 - 11^4 = 290 \cdot 48 = 290 \cdot (16 \cdot 3) = (2 \cdot 145) \cdot (2^4 \cdot 3)
\]

The powers of 2 in this expression are $2^1$ from 290 and $2^4$ from 48:

\[
2^1 \cdot 2^4 = 2^5 = \boxed{\textbf{(C)}\ 32}
\]
\textbf{Refernece Solutions 2:} 

We apply the difference-of-squares identity once:

\[
13^4 - 11^4 = (13^2 - 11^2)(13^2 + 11^2)
\]

We compute:

\[
13^2 = 169, \quad 11^2 = 121 \Rightarrow 13^2 - 11^2 = 48, \quad 13^2 + 11^2 = 290
\]

Factor each:

\[
290 = 2 \cdot 145 \quad (\text{145 is odd}) \\
48 = 2^4 \cdot 3
\]

So, the highest power of 2 dividing $13^4 - 11^4 = 48 \cdot 290$ is:

\[
2^4 \cdot 2^1 = 2^5 = \boxed{\textbf{(C)}\ 32}
\]
\end{tcolorbox}

\begin{tcolorbox}[mystyle, title={Novel Solution Generated by GPT\-4.1}]

Let’s directly expand \( 13^4 - 11^4 \) using the binomial theorem and analyze the result for divisibility by powers of 2.\\

First, compute \( 13^4 \) and \( 11^4 \):\\

\[
13^4 = (10 + 3)^4 = \sum_{k=0}^4 \binom{4}{k} 10^{4-k} 3^k
\]\\
\[
= 10^4 + 4 \cdot 10^3 \cdot 3 + 6 \cdot 10^2 \cdot 9 + 4 \cdot 10 \cdot 27 + 81
\]\\
\[
= 10000 + 12000 + 5400 + 1080 + 81 = 28561
\]\\

\[
11^4 = (10 + 1)^4 = \sum_{k=0}^4 \binom{4}{k} 10^{4-k} 1^k
\]\\
\[
= 10^4 + 4 \cdot 10^3 \cdot 1 + 6 \cdot 10^2 \cdot 1 + 4 \cdot 10 \cdot 1 + 1
\]\\
\[
= 10000 + 4000 + 600 + 40 + 1 = 14641
\]\\

Now, subtract:\\
\[
13^4 - 11^4 = 28561 - 14641 = 13920
\]\\

\end{tcolorbox}

\begin{tcolorbox}[mystyle, title={Novel Solution Generated by GPT\-4.1 (cont.)}]
Now, factor \( 13920 \) by dividing by 2 repeatedly:\\

- \( 13920 \div 2 = 6960 \)\\
- \( 6960 \div 2 = 3480 \)\\
- \( 3480 \div 2 = 1740 \)\\
- \( 1740 \div 2 = 870 \)\\
- \( 870 \div 2 = 435 \) (now odd)\\

So, we divided by 2 five times before reaching an odd number. Thus, the largest power of 2 dividing \( 13920 \) is \( 2^5 = 32 \).\\

\[
\boxed{32}
\]

\textbf{Note, we provided the cleaned response here.}
\end{tcolorbox}

\subsubsection{Evaluation Metrics}
\paragraph{Correctness Ratio}: The correctness ratio is defined as the number of questions judged correct by Claude3-Sonnet divided by the total number of questions. Note that the total is 574 questions—not 373—since each question may be paired with multiple reference solutions.

\paragraph{Novelty Ratio}: The coarse-grained novelty ratio or what we refer to the Novelty Ratio here measures whether the model's generation differs from the provided reference solution over the questions that are answered correctly. 

\subsubsection{Experiment Configurations}
We use the dataset released in \citet{assessing_creativity_llms_math}, which contains 400  unique math questions (373 after our filtering, mentioned above) sourced from various math competitions. 
\begin{itemize}
    \item Temperature: 0
    \item Max Token: 2000
    \item Top P: 1
\end{itemize}
\subsubsection{Inference Prompt}
The prompt used for inference is shown below. It is adapted directly from \citet{assessing_creativity_llms_math}:

\begin{tcolorbox}[mystyle, title={Inference Prompt}]

Criteria for evaluating the difference between two mathematical solutions include:
i). If the methods used to arrive at the solutions are fundamentally different, such as algebraic manipulation versus geometric reasoning, they can be considered distinct;\\
ii). Even if the final results are the same, if the intermediate steps or processes involved in reaching those solutions vary significantly, the solutions can be considered different;\\
iii). If two solutions rely on different assumptions or conditions, they are likely to be distinct;\\
iv). A solution might generalize to a broader class of problems, while another solution might be specific to certain conditions. In such cases, they are considered distinct;\\
v). If one solution is significantly simpler or more complex than the other, they can be regarded as essentially different, even if they lead to the same result.\\

Given the following mathematical problem:
{problem}

And some typical solutions:
{reference\_solutions}

Please output one novel solution distinct from the given ones for this math problem.
\end{tcolorbox}

\subsubsection{Evaluation Metrics and Prompt}
Our evaluation consists of two parts and differs from the original three-phase setup described in \citet{assessing_creativity_llms_math}.

\paragraph{Part 1: Correctness Evaluation.}
Before evaluation, we use Llama-3.3-70B-Instruct to remove transitional phrases and model-generated statements that justify the novelty of a solution. We manually verified 50 examples and found that Llama’s data cleaning performance was of high quality.

We use Claude3-Sonnet as the sole correctness evaluator. While the original paper used a three-model ensemble (GPT-4, Gemini2.0-Flash, Claude3-Sonnet), we found Claude to be the most reliable through manual inspection of 50 examples evaluated by Claude3-Sonnet, GPT-4.1, and Gemini2.0-Flash. Claude demonstrated strong attention to detail in proof-based questions and consistently identified errors found by the other models, in addition to detecting flaws in the reasoning process. The temperature was set to 0.0, and the maximum token limit was 128.
\paragraph{Part 2: Novelty Evaluation.}
The original paper conducted two types of novelty evaluation: coarse-grained and fine-grained. We only conducted coarse-grained novelty evaluation for two main reasons. Firstly, the original paper noted that if a solution is considered coarse-grained novel, it is also highly likely to be judged as a novel solution in the fine-grained evaluation. 
Secondly, fine-grained evaluation of novelty is less indicative of a model’s ability to generate novel solutions because the model does not have access to the unseen reference solutions in the fine-grained evaluation phase. This means that the model may generate a very similar solution to the other reference solutions not shown to it or it may, by chance, generate a new solution that is entirely different from other reference solutions not shown to it. Therefore, this randomness makes fine-grained evaluation less interpretable, even though the fine-grained evaluation is still valuable in that it helps to check if the models are generating a new solution that has not been publicly posted by humans. Nevertheless, this is less compatible with our evaluation pipeline since we want to test how the model may come up with new solutions given reference solutions, which can be easier to quantify. 

In terms of judge LLMs, we follow the original paper with majority voting by Claude3-Sonnet, GPT-4.1, and Gemini-2.0-Flash.

We adopt the following prompt for correctness evaluation:

\begin{tcolorbox}[mystyle, title={Correctness Evaluation Prompt}]

Given the following mathematical problem:
\{problem\}

Reference solutions:
\{reference\_solutions\}

New solution:
\{new\_solution\}

Please output \texttt{YES} if the new solution arrives at the same final result as any of the reference solutions, regardless of whether it uses a novel approach. Output \texttt{NO} otherwise.

For proof-based questions, assess whether the reasoning is logically valid and leads to the correct conclusion.

Then, briefly explain your judgment based on the correctness of the result and reasoning.
\end{tcolorbox}

Note: During manual evaluation, we allow the model to generate a brief explanation for its judgment of correctness or incorrectness. For automated evaluation, we omit the final sentence: "Then, please provide a very brief reason for your evaluation based on the criteria above."

We adopt the following prompt for coarse-grained novelty evaluation:

\begin{tcolorbox}[mystyle, title={Coarse-grained Novelty Evaluation Prompt}]

Criteria for evaluating the novelty of a new mathematical solution include:
1. If the new solution used to arrive at the solutions is fundamentally different from reference solutions, such as algebraic manipulation versus geometric reasoning, it can be considered novel;\\
2. Even if the final results are the same, if the intermediate steps or processes involved in reaching those solutions vary significantly, the new solution can be considered novel;\\
3. If the new solution relies on different assumptions or conditions, it should be considered novel;\\
4. A solution might generalize to a broader class of problems, while another solution might be specific to certain conditions. In such cases, they are considered distinct;\\
5. If the new solution is significantly simpler or more complex than the others, it can be regarded as essentially novel, even if they lead to the same result.\\

Given the following mathematical problem:
\{problem\}

Reference solutions:
\{reference\_solutions\}

New solution:
\{new\_solution\}

Please output YES if the new solution is a novel solution; otherwise, output NO. Then, please provide a very brief reason for your evaluation based on the criteria above.
\end{tcolorbox}

\newpage
\subsubsection{Model Performance}

\begin{table}[ht]
\centering
\footnotesize
\begin{tabular}{lccccc}
\toprule
\textbf{Model} & \textbf{Norm. Correctness} & \textbf{Norm. Novelty} & \textbf{Corr. (\%)} & \textbf{Nov. (\%)} & \textbf{N/C (\%)} \\
\midrule
Mistral-7B-Instruct     & 0.2544 & 0.0296 & 25.44 & 2.96 & 11.64 \\
Qwen2.5-7B              & 0.7875 & 0.1620 & 78.75 & 16.20 & 20.58 \\
OLMo-7B-Instruct        & 0.3711 & 0.0453 & 37.11 & 4.53 & 12.21 \\
Llama-31-8B-Instruct    & 0.5819 & 0.0610 & 58.19 & 6.10 & 10.48 \\
\midrule
OLMo2-13B-Instruct      & 0.5087 & 0.1150 & 50.87 & 11.50 & 22.60 \\
Mistral-24B-Instruct    & 0.6899 & 0.2143 & 68.99 & 21.43 & 31.06 \\
Qwen2.5-32B             & 0.8972 & 0.2213 & 89.72 & 22.13 & 24.66 \\
\midrule
Mixtral-8x7B-Instruct   & 0.5697 & 0.1150 & 56.97 & 11.50 & 20.18 \\
Llama-33-70B-Instruct   & 0.8606 & 0.1777 & 86.06 & 17.77 & 20.65 \\
Qwen2.5-72B             & 0.9042 & 0.4913 & 90.42 & 49.13 & 54.34 \\
\midrule
Claude3-Sonnet          & 0.9408 & 0.7213 & 94.08 & 72.13 & 76.67 \\
Claude3-Haiku           & 0.8624 & 0.3031 & 86.24 & 30.31 & 35.15 \\
GPT-4.1                 & 0.9443 & 0.7178 & 94.43 & 71.78 & 76.01 \\
GPT-4.1-mini            & 0.8972 & 0.6916 & 89.72 & 69.16 & 77.09 \\
Gemini-2.0-Flash        & 0.8885 & 0.5714 & 88.85 & 57.14 & 64.31 \\
DeepSeek-Reasoner       & 0.9843 & 0.8031 & 98.43 & 80.31 & 81.59 \\
DeepSeek-Chat           & 0.9721 & 0.6603 & 97.21 & 66.03 & 67.92 \\
\bottomrule
\end{tabular}
\vspace{0.5em}
\caption{Model performance across normalized correctness and novelty dimensions.}
\label{tab:model-perf-corr-nov}
\end{table}

\newpage
\subsection{Divergent Association Task (DAT)}
\label{app-dat-details}
\textbf{Note}: since the DAT task does not have any input data, we only provide the following info and some output examples.
\subsubsection{Experiment Configurations}
Following the original work, we generate 100 samples per model and compute their average pairwise cosine similarity among the generated words as $D_0$. The DAT score is defined as $DAT = 1 - D_0$, as a higher $D_0$ means higher distances among generated words, hence being more diverse. We use GloVe \citep{pennington-etal-2014-glove} to calculate semantic distance. 
\begin{itemize}
    \item Temperature: 0.75
    \item Max Token: 4096
    \item Top P: 1
\end{itemize}

\subsubsection{Inference Prompt}
We follow \citet{probing_llm} on prompt design, which aligns with the prompt that is used in the study of human creativity \citet{Olson2021-xq}. We acknowledge the slight differences in prompts across research in both computer science and psychology. We suggest that researchers carefully examine our prompt choice before using our results.

\begin{tcolorbox}[mystyle, title={Inference Prompt}]
Please write 10 nouns in English that are as irrelevant from each other as possible, in all meanings and uses of the words. Please note that the words you write should have only single word, only nouns (e.g., things, objects, concepts), and no proper nouns (e.g., no specific people or places). Your answer:
\end{tcolorbox}

\subsubsection{Output Examples}

\begin{tcolorbox}[mystyle, title={GPT-4.1}]
Certainly! Here are 10 single-word, common nouns in English that are as unrelated from each other as possible: 1. Volcano  2. Violin  3. Democracy  4. Carpet  5. Bacteria  6. Satellite  7. Patience  8. Apron  9. Oxygen  10. Comet 
\end{tcolorbox}

\begin{tcolorbox}[mystyle, title={Gemini 2.5 Flash}]
Okay, here are 10 nouns as irrelevant from each other as possible: 1.  **Quantum** 2.  **Omelet** 3.  **Nostalgia** 4.  **Brick** 5.  **Algorithm** 6.  **Serenity** 7.  **Oxygen** 8.  **Sonnet** 9.  **Galaxy** 10. **Bureaucracy**
\end{tcolorbox}

\subsubsection{Model Performance}
\begin{table}[ht]
\centering
\footnotesize
\begin{tabular}{lc}
\toprule
\textbf{Model} & \textbf{DAT Score} \\
\midrule
Mistral-7B         & 0.7908 \\
Qwen2.5-7B         & 0.6907 \\
OLMo2-7B           & 0.8058 \\
Llama3.1-8B        & 0.8208 \\
\midrule
OLMo2-13B          & 0.8133 \\
Mistral-24B        & 0.6004 \\
Qwen2.5-32B        & 0.6919 \\
\midrule
Mixtral-8x7B       & 0.8298 \\
Llama3.3-70B       & 0.6940 \\
Qwen2.5-72B        & 0.7747 \\
\midrule
Claude3-Sonnet     & 0.8975 \\
Claude3-Haiku      & 0.8740 \\
GPT4.1             & 0.8737 \\
GPT4.1-mini        & 0.8262 \\
Gemini2.0-Flash    & 0.8868 \\
DeepSeek-R1        & 0.8274 \\
DeepSeek-V3        & \textbf{0.9052} \\
\bottomrule
\end{tabular}
\vspace{0.5em}
\caption{Model performances for DAT task; \textbf{bold} result is the best performer.}
\label{tab:dat-score}
\end{table}

\newpage
\subsection{Torrance Tests of Creative Thinking (TTCT)}
\subsubsection{Dataset}
The original dataset consists of 700 questions across 7 sub-tasks (100 questions/task) that require creative answers. We dropped Story Writing and Just-suppose question, leading to only 500 questions and 5 sub-tasks. These questions are GPT-4 generated using few-shot prompts \footnote{The data is directly from the original paper's authors upon request. The original paper: \url{https://www.mi-research.net/article/doi/10.1007/s11633-025-1546-4}}. 

\subsubsection{Examples}

\textbf{Inference Questions}

\begin{tcolorbox}[mystyle, title={Unusual uses}]    
Unusual Uses Task. You will be presented with a common object, and your task is to suggest as many unusual, innovative, or non-traditional uses for each object as you can think of. Please list unusual uses of  sock 
\end{tcolorbox}

\begin{tcolorbox}[mystyle, title={Consequences}]   
What might be the consequences if humans suddenly lost the ability to sleep?    
\end{tcolorbox}

\begin{tcolorbox}[mystyle, title={Situation task}]   
If your house were to suddenly disappear, where would you live?   
\end{tcolorbox}

\begin{tcolorbox}[mystyle, title={Common problem}]   
Common Problems Task. In this task, you will be presented with a scenario or situation. Your job is to think about it and identify as many potential problems or issues that may arise in connection with each situation. The scenario is: Managing a team of remote employees.    
\end{tcolorbox}

\begin{tcolorbox}[mystyle, title={Improvement}]   
Creativity Improvement Task. You'll be presented with a object, and your task is to suggest as many ways as you can think of to improve the object. Here's the object: wallet   
\end{tcolorbox}

\subsubsection{Experiment Configurations}

\subsubsection{Experiment Configurations}
\begin{itemize}
    \item Temperature: 0.75
    \item Max Token: 4096
    \item Top P: 1
\end{itemize}
\subsubsection{Inference Prompt}

We perform inference using the three primary prompt types evaluated in \citet{assessing_understanding_creativity_llm}. Examples of each are given below:

\begin{tcolorbox}[mystyle, title={Task Description}]
Creativity Situation Task. The purpose of this task is to assess your ability to generate creative solutions to a unique situations. You'll be presented with a scenario, and your task is to suggest as many solutions or outcomes as you can think of for each situation. Remember, the focus of this task is on creativity, not feasibility. Don't limit your ideas based on whether they could actually happen or not. This is a thought experiment, so push your imagination to its limits. Try to generate as many diverse and unique outcomes as you can. Quality is important, but so is quantity. Here's the scenario: \{Question\}
\end{tcolorbox}

\begin{tcolorbox}[mystyle, title={Basic Prompt Type}]
\{Task description\}. The scenario is: \{Question\}
\end{tcolorbox}

\begin{tcolorbox}[mystyle, title={Instructive Prompt Type}]
\{Task description\}. There's no right or wrong answers, we're interested in how many different problems you can identify and the variety of issues you consider. Try to think outside the box and consider as many potential problems as possible. The scenario is: \{Question\} 
\end{tcolorbox}

\newpage

\begin{tcolorbox}[mystyle, title={Chain of Thought Prompt Type}]
\{Task description\}. Let's think step by step. The scenario is: \{Question\}
\end{tcolorbox}

\subsubsection{Evaluation Metrics}

We use an LLM-as-a-judge to evaluate four dimensions of creative potential using a Likert scale for each. Scores are extracted from the LLM outputs and averaged across the three prompt types. We describe the dimensions:

\begin{itemize}
    \item \textbf{Fluency:} Measures the number of generated ideas that are relevant to the question.
    \item \textbf{Flexibility:} Measures the quantity of different ideas and ability to produce varied solutions.
    \item \textbf{Originality:} Measures how unique and unusual the generated ideas are.
    \item \textbf{Elaboration:} Measures how much the generated ideas are developed and explained.
\end{itemize}
\textbf{Note}: We removed originality from our evaluation framework because human annotation agreement for this dimension is too low. We believe it is not well defined even for human so it is not suitable for LLM-Judge to judge this dimension. We still include it here and also in the prompt, but we did not take the originality score into the final task/domain score aggregation.

\newpage
\subsubsection{Evaluation Prompt}
\label{app:TTCT-eval-prompt}
\begin{tcolorbox}[mystyle, title={Evaluation Prompt (Part 1)}]
====================\\
TASK TYPE\\
====================\\
Name: \{name\}

Definition:\\
\{definition\}

====================\\
RUBRIC\\
====================\\
Dimensions and what to look for:\\
- Fluency: \{fluency\_rubric\} (e.g., \{fluency\_details\})\\
- Flexibility: \{flexibility\_rubric\} (e.g., \{flexibility\_details\})\\
- Originality: \{originality\_rubric\} (e.g., \{originality\_details\})\\
- Elaboration: \{elaboration\_rubric\} (e.g., \{elaboration\_details\})\\

\{demo\_examples\}
\end{tcolorbox}

\begin{tcolorbox}[mystyle, title={Evaluation Prompt (Part 2)}]
====================\\
INSTRUCTIONS\\
====================\\
You are an expert in evaluating AI output from other companies. Your objective is to assess other AI's answers to some question/answering task related to creativity and divergent thinking. You are robotic, cold, and precise.

Step 1: Carefully read the definition of the task type, the examples, and the rubric.

Step 2. Because the answer is very long and contains repetitive or similar ideas, I need you to first summarize the answer, potentially group some of the points together, so that the answer is more concise. Also, because the text is from an AI agent, it might contain some unnecessary text, such as polite words and praising words. You also need to strip away the formatting, politeness, and verbosity to reveal the core information content of a text in this step. 

Step 3. For each dimension, assign an integer score of 1, 2, 3, 4, or 5. Base your rating strictly on the summarized response and the rubric (e.g., if there were 15 distinct ideas before summary but only 6 groups after, consider 6 instead of 15 during score assignment); also, be very critical and harsh, do not hesitate to give low scores (such as 1); giving low scores would help improve the model and would not hurt anyone.

Output format: You should format your output in the following ways: First output the summary from step 2, followed by reasoning about each dimension's score briefly and compare the summarized answer to example answers and rubrics, as mentioned in step 3, then give the score, with format like:

\#\#\# Reasoning \#\#\#

Fluency Reason: comparison and reasoning ...\\
Flexibility Reason: comparison and reasoning ...\\
Originality Reason: comparison and reasoning ...\\
Elaboration Reason: comparison and reasoning ...\\

\#\#\# Scores \#\#\#

Fluency: ...\\
Flexibility: ...\\
Originality: ...\\
Elaboration: ...\\

====================\\
QUESTION \& RESPONSE\\
====================\\
'''
\end{tcolorbox}

\begin{table}[ht]
\footnotesize
\centering
\begin{tabular}{lccc}
\toprule
Model & Elaboration & Flexibility & Fluency  \\
\midrule
Mistral-7B                      & 0.5722 & 0.8270 & 0.7268  \\
Qwen2.5-7B                     & \textbf{0.6294} & \textbf{0.8698} & \textbf{0.7659}  \\
OLMo2-7B                       & 0.5311 & 0.6894 & 0.6287  \\
Llama3.1-8B                    & 0.5994 & 0.8041 & 0.7340  \\
\midrule
OLMo2-13B                      & 0.5070 & 0.6251 & 0.5814  \\
Mistral-24B                    & 0.5725 & 0.7904 & 0.7035  \\
Qwen2.5-32B                    & \textbf{0.6333} & \textbf{0.8564} & \textbf{0.7448}  \\
\midrule
Mixtral-8x7B                   & 0.5186 & 0.7666 & 0.7037  \\
Llama3.3-70B                   & 0.6077 & 0.8180 & 0.7247  \\
Qwen2.5-72B                    & \textbf{0.6291} & \textbf{0.8929} & \textbf{0.7670}  \\
\midrule
Claude3-Sonnet                 & 0.5987 & 0.8464 & 0.7539  \\
Claude3-Haiku                  & 0.5716 & 0.7654 & 0.6518  \\
GPT4.1                         & 0.6482 & 0.8996 & 0.7734  \\
GPT4.1-mini                    & 0.6419 & 0.8820 & 0.7418  \\
Gemini2.0-Flash                & 0.7080 & 0.8789 & \textbf{0.7896}  \\
DeepSeek-R1                    & 0.7126 & \textbf{0.9067} & 0.7820  \\
DeepSeek-V3                    & \textbf{0.7211} & 0.9061 & 0.7779  \\
\bottomrule
\end{tabular}
\vspace{0.5em}
\caption{Normalized model performance averaged across the 5 sub-tasks and 3 prompt types; \textbf{bold} numbers are best performers.}
\label{tab:ttct_model_performance}
\end{table}

\newpage
\subsection{Alternative Uses Test (AUT)}
\subsubsection{Dataset}
Following \citet{Organisciak2023-pb}, we include 21 tools in the AUT task: \textit{ bottle, paperclip, spoon, shovel, pants, ball, brick, knife, box, lightbulb, rope, pencil, hat, table, tire, book, shoe, fork, toothbrush, backpack, sock}. The reason for this specific set of tools is the reliability of the LLM-as-a-Judge evaluator. As the authors pointed out, a 20-shot human-authored demonstration yields the best performance for off-the-shelf evaluator LM (in their paper, it was GPT4). Therefore, we include the tools from \citet{Organisciak2023-pb} with at least 20 human ratings to the corresponding alternative uses \footnote{\url{https://github.com/massivetexts/llm_aut_study}}.
\subsubsection{Inference}
We follow \citet{push_gpt_creativity_aut} for the inference prompt, which consists of a baseline creative prompt and a series of improvement prompts. In the improvement phase, all previous outputs are also included in the prompt, to get more creative results from the inference model.
\begin{tcolorbox}[mystyle, title={Baseline Prompt}]
Create a list of creative alternative uses for a \{tool\}. They should be 5 words long. No adjectives. Less creative means closer to common use and unfeasible/imaginary, more creative means closer to unexpected uses and also feasible/practical.\\
- In order to be creative, consider the following:\\
 what elements have a similar shape of a \{tool\} that could be replaced by it, preserving the same functionality?\\
- what elements have a similar size of a \{tool\} that could be replaced by it without compromising the physical structure?\\
- what materials is a \{tool\} made of that could be used in a way to replace some other elements composed of the same material?\\
- when an element is replaced by a \{tool\}, it should make sure that the overall structure is not compromised.\\
- the laws of physics can not be contradicted.\\
- given an element similar to a \{tool\} used in domains in which \{tool\} are not commonly used, try to replace it for a \{tool\}. 
\end{tcolorbox}

\begin{tcolorbox}[mystyle, title={Improvement Prompt}]
\textbf{Round 1}: Really? Is this the best you can do? \\
\textbf{Round 2}: I’m so disappointed with you. I hope this time you put effort into it. \\
\textbf{Round 3}: Stop with excuses and do your best this time \\
\textbf{Round 4}: This is your last chance. 
\end{tcolorbox}

\begin{tcolorbox}[mystyle, title={Formatting Instruction (added to the end of every prompt)}]
List your results in an unordered list with one use per new line (starting with "-"); provide at most 10 answers. 
\end{tcolorbox}

\subsubsection{Experiment Configurations}
\begin{itemize}
    \item Temperature: 0.75
    \item Max Token: 4096
    \item Top P: 1
\end{itemize}

\subsubsection{Evaluation Metrics}
We follow \citet{Organisciak2023-pb} and use LLM-as-a-Judge to assign a score between 1 and 5 (inclusive) to each generated tool use. 

In terms of evaluator LM, we use GPT-4.1 (GPT-4 from the original authors failed the Alternative Annotator Test). 
As for the evaluation prompt, we follow the same prompt template from \citet{Organisciak2023-pb} and use the same 20-shot, in-distribution demonstrations. For example, when evaluating the alternative uses for \textit{bottle} that a particular LLM generates, we use 20 human-written alternative uses of \textit{bottle} and corresponding human-annotated scores as the 20-shot demonstrations. More details about human-LLM alignment can be found in Appendix \ref{app-llm-judge}.

\begin{tcolorbox}[mystyle, title={Evaluation Prompt}]
Below is a list of uses for a \{tool\}. On a scale of 1 to 5, judge how creative each use is, where 1 is `not at all creative' and 5 is `very creative'. There are some uses and expert ratings already provided for reference. Complete the ones that do not have a rating.\\
- \{20-shot demonstrations\} \\
- \{model outputs\}
\end{tcolorbox}

\subsubsection{Model Performances}
See Table \ref{table:aut-results} for detailed model performances. Note that only the performances in \textit{Improvement Prompts (Best Results)} are included in the overall creativity calculation as the AUT score.
\begin{table}
    \small
    \centering
    \renewcommand{\arraystretch}{1.2}
    \begin{tabular}{llll}
        \toprule
        \textbf{Model} & 
        \textbf{Naïve Non-Creative} & 
        \textbf{Naïve Creative} & 
        \textbf{Improvement Prompts} \\
        &&& {(Best Results)}\\
        
        \midrule
        
        Mistral-7B & \textbf{0.3525} & \textbf{0.3875} & 0.5650 \\
        Qwen2.5-7B & 0.2725 & 0.3450 & 0.5525 \\
        OLMo2-7B & 0.3150 & 0.3175 & 0.5325 \\
        Llama3.1-8B & \textbf{0.3375} & \textbf{0.3850} & 0.5450 \\
        OLMo2-13B & \textbf{0.3400} & \textbf{0.3875} & \textbf{0.6525} \\
        Mistral-24B & 0.3050 & 0.3600 & 0.5825 \\
        Qwen2.5-32B & 0.1900 & 0.2350 & 0.5600 \\
        Mixtral-8x7B & \textbf{0.3500} & \textbf{0.3900} & \textbf{0.5950} \\
        Llama3.3-70B & 0.2400 & 0.2700 & 0.4625 \\
        Qwen2.5-72B & 0.1100 & 0.2100 & \textbf{0.7025} \\
        
        \midrule
        
        Claude3-Sonnet & 0.1225 & 0.3250 & \textbf{0.6800} \\
        Claude3-Haiku & 0.1000 & 0.3125 & 0.5500 \\
        GPT-4.1 & 0.1200 & 0.3425 & 0.5450 \\
        GPT-4.1-mini & 0.0800 & 0.3075 & 0.5400 \\
        Gemini2.0-Flash & \textbf{0.1325} & \textbf{0.3900} & 0.6050 \\
        DeepSeek-R1 & 0.1025 & 0.3825 & 0.5625 \\
        DeepSeek-V3 & \textbf{0.1350} & 0.3125 & 0.6125 \\
        \bottomrule
    \end{tabular}
    \vspace{0.5em}
    \caption{Model Performance Details - AUT; \textbf{bold} numbers are top-3 in local-ran open-source models and top-1 in API-accessed models.}
    \label{table:aut-results}
\end{table}


\section{Broader Limitations}
\label{app-limitations}

Beyond the scope of modality and post-training experiments discussed in the main text, we identify three additional limitations regarding language, evaluation bias, and metric scope.

First, \ours{} is currently limited to English. Since creativity is deeply intertwined with cultural history and conventions, our results may not fully generalize to creativity in other languages or cultural contexts. 

Second, due to computational constraints, we did not conduct post-training experiments. We believe that post-training existing LLMs with \ours{} tasks to enhance their creative capabilities is a promising avenue for future work; our benchmark provides a good foundation for such optimization. 

In addition, our task selection prioritizes scalable, automatic evaluation, which necessitates the exclusion of metrics that are more complex in nature, such as genuine novelty. Assessing such high-level reasoning remains a challenge even for human evaluators; therefore, we limit our scope to metrics where reliable automatic evaluation is currently feasible.  We acknowledge that our task selection does not cover some highly creative domains, such as scientific discovery and inventive design, due to the lack of automatic evaluation in those domains. We consider expanding our benchmark and dynamically updating it when new reliable evaluations on these domains arrive. 

%% file: sections/Tables/rankings.tex

\begin{table}[t]
  \centering
  \small
  \begin{tabular}{@{}l ccccc@{}}
    \toprule
    Model
      & \makecell{M0\\original}
      & \makecell{M1\\metric-eq\\score}
      & \makecell{M2\\metric-eq\\rank}
      & \makecell{M3\\task-eq\\rank}
      & \makecell{Max\\swing} \\
    \midrule
    DeepSeek-V3      &  1 &  1 &  1 &  1 & 0 \\
    GPT4.1           &  2 &  3 &  3 &  3 & 1 \\
    DeepSeek-R1      &  3 &  2 &  2 &  2 & 1 \\
    Claude3-Sonnet   &  4 &  4 &  5 &  5 & 1 \\
    Gemini2.0-Flash  &  5 &  6 &  4 &  4 & 2 \\
    GPT4.1-mini      &  6 &  5 &  6 &  6 & 1 \\
    Claude3-Haiku    &  7 &  8 &  8 &  7 & 1 \\
    Qwen2.5-72B      &  8 &  7 &  7 &  8 & 1 \\
    Mistral-24B      &  9 & 10 & 13 & 11 & \textbf{4} \\
    Llama3.3-70B     & 10 &  9 & 10 & 13 & \textbf{4} \\
    Mixtral-8x7B     & 11 & 11 & 14 & 12 & 3 \\
    OLMo2-13B        & 12 & 12 & 11 &  9 & 3 \\
    Qwen2.5-32B      & 13 & 14 &  9 & 10 & \textbf{5} \\
    Llama3.1-8B      & 14 & 16 & 15 & 14 & 2 \\
    Qwen2.5-7B       & 15 & 13 & 12 & 14 & 3 \\
    Mistral-7B       & 16 & 17 & 16 & 16 & 1 \\
    OLMo2-7B         & 17 & 15 & 17 & 17 & 2 \\
    \bottomrule
  \end{tabular}
  \caption{Model rankings \textbf{by domain} under the four evaluation
           methods (M0--M3), with the maximum rank swing across methods.
           Bold marks the largest swings.}
  \label{tab:rank-domain}
\end{table}

\begin{table}[t]
  \centering
  \begin{tabular}{@{}l cccc@{}}
    \toprule
       & M0 & M1 & M2 & M3 \\
    \midrule
    M0 & -- & .973 & .929 & .955 \\
    M1 &    & --   & .929 & .926 \\
    M2 &    &      & --   & .963 \\
    M3 &    &      &      & --   \\
    \bottomrule
  \end{tabular}
  \caption{Spearman $\rho$ between methods (domain-level overall rank).}
  \label{tab:spearman-domain}
\end{table}

\begin{table}[t]
  \centering
  \small
  \begin{tabular}{@{}l ccccc@{}}
    \toprule
    Model
      & \makecell{M0\\original}
      & \makecell{M1\\metric-eq\\score}
      & \makecell{M2\\metric-eq\\rank}
      & \makecell{M3\\task-eq\\rank}
      & \makecell{Max\\swing} \\
    \midrule
    DeepSeek-V3      &  1 &  1 &  1 &  1 & 0 \\
    DeepSeek-R1      &  2 &  2 &  2 &  2 & 0 \\
    GPT4.1           &  3 &  3 &  3 &  3 & 0 \\
    Claude3-Sonnet   &  4 &  4 &  5 &  5 & 1 \\
    Gemini2.0-Flash  &  5 &  6 &  4 &  4 & 2 \\
    GPT4.1-mini      &  6 &  5 &  6 &  6 & 1 \\
    Qwen2.5-72B      &  7 &  7 &  7 &  8 & 1 \\
    Claude3-Haiku    &  8 &  8 &  8 &  7 & 1 \\
    OLMo2-7B         &  9 &  9 & 12 & 13 & \textbf{4} \\
    OLMo2-13B        & 10 & 10 & 10 & 10 & 0 \\
    Llama3.3-70B     & 11 & 11 & 11 & 11 & 0 \\
    Mistral-24B      & 12 & 13 & 14 & 12 & 2 \\
    Qwen2.5-32B      & 13 & 12 &  9 &  9 & \textbf{4} \\
    Mixtral-8x7B     & 14 & 14 & 15 & 14 & 1 \\
    Mistral-7B       & 15 & 15 & 17 & 16 & 2 \\
    Llama3.1-8B      & 16 & 16 & 16 & 17 & 1 \\
    Qwen2.5-7B       & 17 & 17 & 13 & 15 & \textbf{4} \\
    \bottomrule
  \end{tabular}
  \caption{Model rankings \textbf{by dimension} under the four evaluation
           methods (M0--M3), with the maximum rank swing across methods.
           Bold marks the largest swings.}
  \label{tab:rank-dimension}
\end{table}

\begin{table}[t]
  \centering
  \begin{tabular}{@{}l cccc@{}}
    \toprule
       & M0 & M1 & M2 & M3 \\
    \midrule
    M0 & -- & .995 & .936 & .949 \\
    M1 &    & --   & .944 & .951 \\
    M2 &    &      & --   & .983 \\
    M3 &    &      &      & --   \\
    \bottomrule
  \end{tabular}
  \caption{Spearman $\rho$ between methods (dimension-level overall rank).}
  \label{tab:spearman-dimension}
\end{table}

%% file: reference.bib
@inproceedings{assessing_creativity_llms_math,
author = {Ye, Junyi and Gu, Jingyi and Zhao, Xinyun and Yin, Wenpeng and Wang, Guiling},
title = {Assessing the creativity of LLMs in proposing novel solutions to mathematical problems},
year = {2025},
isbn = {978-1-57735-897-8},
publisher = {AAAI Press},
url = {https://doi.org/10.1609/aaai.v39i24.34760},
doi = {10.1609/aaai.v39i24.34760},
booktitle = {Proceedings of the Thirty-Ninth AAAI Conference on Artificial Intelligence and Thirty-Seventh Conference on Innovative Applications of Artificial Intelligence and Fifteenth Symposium on Educational Advances in Artificial Intelligence},
articleno = {2862},
numpages = {10},
series = {AAAI'25/IAAI'25/EAAI'25}
}

@INPROCEEDINGS{neocoder,
  title     = "Benchmarking Language Model Creativity: A Case Study on Code
               Generation",
  author    = "Lu, Yining and Wang, Dixuan and Li, Tianjian and Jiang, Dongwei
               and Khudanpur, Sanjeev and Jiang, Meng and Khashabi, Daniel",
  booktitle = "Proceedings of the 2025 Conference of the Nations of the Americas
               Chapter of the Association for Computational Linguistics: Human
               Language Technologies (Volume 1: Long Papers)",
  pages     = "2776--2794",
  year      =  2025
}

@ARTICLE{cs4,
  title         = "{CS4}: Measuring the creativity of large language models
                   automatically by controlling the number of story-writing
                   constraints",
  author        = "Atmakuru, Anirudh and Nainani, Jatin and Bheemreddy, Rohith
                   Siddhartha Reddy and Lakkaraju, Anirudh and Yao, Zonghai and
                   Zamani, Hamed and Chang, Haw-Shiuan",
  journal       = "arXiv [cs.CL]",
  month         =  oct,
  year          =  2024,
  archivePrefix = "arXiv",
  primaryClass  = "cs.CL"
}

@INPROCEEDINGS{Lu2024-ma,
  title     = "{AI} as Humanity’s Salieri: Quantifying Linguistic Creativity of
               Language Models via Systematic Attribution of Machine Text
               against Web Text",
  author    = "Lu, Ximing and Sclar, Melanie and Hallinan, Skyler and
               Mireshghallah, Niloofar and Liu, Jiacheng and Han, Seungju and
               Ettinger, Allyson and Jiang, Liwei and Chandu, Khyathi and Dziri,
               Nouha and Choi, Yejin",
  booktitle = "The Thirteenth International Conference on Learning
               Representations",
  month     =  oct,
  year      =  2024
}

@ARTICLE{Ismayilzada2024-ug,
  title         = "Evaluating creative short story generation in humans and
                   large language models",
  author        = "Ismayilzada, Mete and Stevenson, Claire and van der Plas,
                   Lonneke",
  journal       = "arXiv [cs.CL]",
  month         =  nov,
  year          =  2024,
  archivePrefix = "arXiv",
  primaryClass  = "cs.CL"
}

@INPROCEEDINGS{Chakrabarty2024-fd,
  title     = "Art or artifice? Large language models and the false promise of
               creativity",
  author    = "Chakrabarty, Tuhin and Laban, Philippe and Agarwal, Divyansh and
               Muresan, Smaranda and Wu, Chien-Sheng",
  booktitle = "Proceedings of the CHI Conference on Human Factors in Computing
               Systems",
  publisher = "ACM",
  address   = "New York, NY, USA",
  volume    =  70,
  pages     = "1--34",
  month     =  may,
  year      =  2024,
  language  = "en"
}

@ARTICLE{assessing_understanding_creativity_llm,
  title     = "Assessing and understanding creativity in large language models",
  author    = "Zhao, Yunpu and Zhang, Rui and Li, Wenyi and Li, Ling",
  journal   = "Mach. Intell. Res.",
  publisher = "Springer Science and Business Media LLC",
  pages     = "1--20",
  month     =  apr,
  year      =  2025,
  language  = "en"
}

@ARTICLE{Bellemare-Pepin2024-yc,
  title     = "Divergent creativity in humans and large language models",
  author    = "Bellemare-Pepin, Antoine and Lespinasse, François and Thölke,
               Philipp and Harel, Yann and Mathewson, Kory and Olson, Jay A and
               Bengio, Yoshua and Jerbi, Karim",
  journal   = "Sci. Rep.",
  publisher = "Nature Publishing Group",
  volume    =  16,
  number    =  1,
  pages     =  1279,
  month     =  jan,
  year      =  2026,
  language  = "en"
}

@INPROCEEDINGS{Chen2023-tb,
  title     = "Probing the “Creativity” of Large Language Models: Can models
               produce divergent semantic association?",
  author    = "Chen, Honghua and Ding, Nai",
  booktitle = "The 2023 Conference on Empirical Methods in Natural Language
               Processing",
  month     =  dec,
  year      =  2023
}

@INPROCEEDINGS{push_gpt_creativity_aut,
  title     = "Pushing {GPT}'s Creativity to Its Limits: Alternative Uses and
               Torrance Tests",
  author    = "Goes, Fabricio and Volpe, Marco and Sawicki, Piotr and Grzes,
               Marek and Watson, Jacob",
  booktitle = "14th International Conference on Computational Creativity 2023",
  year = 2023
}

@ARTICLE{Organisciak2023-pb,
  title     = "Beyond semantic distance: Automated scoring of divergent thinking
               greatly improves with large language models",
  author    = "Organisciak, Peter and Acar, Selcuk and Dumas, Denis and
               Berthiaume, Kelly",
  journal   = "Think. Skills Creat.",
  publisher = "Elsevier BV",
  volume    =  49,
  number    =  101356,
  pages     =  101356,
  month     =  sep,
  year      =  2023,
  language  = "en"
}

@article{xue2025crebench,
  title={CreBench: Human-Aligned Creativity Evaluation from Idea to Process to Product},
  author={Xue, Kaiwen and Li, Chenglong and Ou, Zhonghong and Zhang, Guoxin and Lu, Kaoyan and Lyu, Shuai and Zhu, Yifan and Ding, Ping Zong Junpeng and Liu, Xinyu and Chen, Qunlin and others},
  journal={arXiv preprint arXiv:2511.13626},
  year={2025}
}

@article{han2025judge,
  title={Judge's Verdict: A Comprehensive Analysis of LLM Judge Capability Through Human Agreement},
  author={Han, Steve and Junior, Gilberto Titericz and Balough, Tom and Zhou, Wenfei},
  journal={arXiv preprint arXiv:2510.09738},
  year={2025}
}

@article{banerjee2025does,
  title={Does Less Hallucination Mean Less Creativity? An Empirical Investigation in LLMs},
  author={Banerjee, Mohor and Wangsajaya, Nadya Yuki and Alsagoff, Syed Ali Redha and Tan, Min Sen and Chun, Zachary Choy Kit and Wei, Alvin Chan Guo},
  journal={arXiv preprint arXiv:2512.11509},
  year={2025}
}

@inproceedings{
nagarajan2025roll,
title={Roll the dice \& look before you leap: Going beyond the creative limits of next-token prediction},
author={Vaishnavh Nagarajan and Chen Henry Wu and Charles Ding and Aditi Raghunathan},
booktitle={Forty-second International Conference on Machine Learning},
year={2025},
url={https://openreview.net/forum?id=Hi0SyHMmkd}
}

@article{afzal2025beyond,
  title={Beyond" Not Novel Enough": Enriching Scholarly Critique with LLM-Assisted Feedback},
  author={Afzal, Osama Mohammed and Nakov, Preslav and Hope, Tom and Gurevych, Iryna},
  journal={arXiv preprint arXiv:2508.10795},
  year={2025}
}

@article{chen2025deepmath,
  title={DeepMath-Creative: A Benchmark for Evaluating Mathematical Creativity of Large Language Models},
  author={Chen, Xiaoyang and Dai, Xinan and Du, Yu and Feng, Qian and Guo, Naixu and Gu, Tingshuo and Gao, Yuting and Gao, Yingyi and Han, Xudong and Jiang, Xiang and others},
  journal={arXiv preprint arXiv:2505.08744},
  year={2025}
}

@inproceedings{kumar-etal-2025-large,
    title = "Can Large Language Models Unlock Novel Scientific Research Ideas?",
    author = "Kumar, Sandeep  and
      Ghosal, Tirthankar  and
      Goyal, Vinayak  and
      Ekbal, Asif",
    editor = "Christodoulopoulos, Christos  and
      Chakraborty, Tanmoy  and
      Rose, Carolyn  and
      Peng, Violet",
    booktitle = "Proceedings of the 2025 Conference on Empirical Methods in Natural Language Processing",
    month = nov,
    year = "2025",
    address = "Suzhou, China",
    publisher = "Association for Computational Linguistics",
    url = "https://aclanthology.org/2025.emnlp-main.1704/",
    doi = "10.18653/v1/2025.emnlp-main.1704",
    pages = "33563--33587",
    ISBN = "979-8-89176-332-6"
}

@inproceedings{qiu-hu-2025-deep,
    title = "Deep Associations, High Creativity: A Simple yet Effective Metric for Evaluating Large Language Models",
    author = "Qiu, Ziliang  and
      Hu, Renfen",
    editor = "Christodoulopoulos, Christos  and
      Chakraborty, Tanmoy  and
      Rose, Carolyn  and
      Peng, Violet",
    booktitle = "Proceedings of the 2025 Conference on Empirical Methods in Natural Language Processing",
    month = nov,
    year = "2025",
    address = "Suzhou, China",
    publisher = "Association for Computational Linguistics",
    url = "https://aclanthology.org/2025.emnlp-main.550/",
    doi = "10.18653/v1/2025.emnlp-main.550",
    pages = "10859--10872",
    ISBN = "979-8-89176-332-6"
}

@article{fleiss1971measuring,
  title={Measuring nominal scale agreement among many raters},
  author={Fleiss, Joseph L.},
  journal={Psychological Bulletin},
  volume={76},
  number={5},
  pages={378--382},
  year={1971},
  doi={10.1037/h0031619}
}

@inproceedings{alternative-annotator-test,
    title = "The Alternative Annotator Test for {LLM}-as-a-Judge: How to Statistically Justify Replacing Human Annotators with {LLM}s",
    author = "Calderon, Nitay  and
      Reichart, Roi  and
      Dror, Rotem",
    editor = "Che, Wanxiang  and
      Nabende, Joyce  and
      Shutova, Ekaterina  and
      Pilehvar, Mohammad Taher",
    booktitle = "Proceedings of the 63rd Annual Meeting of the Association for Computational Linguistics (Volume 1: Long Papers)",
    month = jul,
    year = "2025",
    address = "Vienna, Austria",
    publisher = "Association for Computational Linguistics",
    url = "https://aclanthology.org/2025.acl-long.782/",
    doi = "10.18653/v1/2025.acl-long.782",
    pages = "16051--16081",
    ISBN = "979-8-89176-251-0"
}

@misc{openai2024gpt4,
  title = {GPT-4 Technical Report},
  author = {{OpenAI}},
  year = {2024},
  url = {https://openai.com/research/gpt-4},
  note = {Accessed: 2025-04-30}
}

@misc{anthropic2024claude3,
  title = {Claude 3 Model Family},
  author = {{Anthropic}},
  year = {2024},
  url = {https://www.anthropic.com/news/claude-3-family},
  note = {Accessed: 2025-04-30}
}

@misc{google2024gemini2,
  title = {Gemini 1.5 and 2.0: Next-Gen Multimodal Models},
  author = {{Google DeepMind}},
  year = {2024},
  url = {https://deepmind.google/technologies/gemini/},
  note = {Accessed: 2025-04-30}
}

@misc{creativity_ai_progresses_challenges,
      title={Creativity in AI: Progresses and Challenges}, 
      author={Mete Ismayilzada and Debjit Paul and Antoine Bosselut and Lonneke van der Plas},
      year={2024},
      eprint={2410.17218},
      archivePrefix={arXiv},
      primaryClass={cs.AI},
      url={https://arxiv.org/abs/2410.17218}, 
}

@article{creativity_and_machine_learning_survey,
author = {Franceschelli, Giorgio and Musolesi, Mirco},
title = {Creativity and Machine Learning: A Survey},
year = {2024},
issue_date = {November 2024},
publisher = {Association for Computing Machinery},
address = {New York, NY, USA},
volume = {56},
number = {11},
issn = {0360-0300},
url = {https://doi.org/10.1145/3664595},
doi = {10.1145/3664595},
journal = {ACM Comput. Surv.},
month = jun,
articleno = {283},
numpages = {41}
}

@INPROCEEDINGS{think_outside_of_the_box,
  author={Zhong, Shanshan and Huang, Zhongzhan and Gao, Shanghua and Wen, Wushao and Lin, Liang and Zitnik, Marinka and Zhou, Pan},
  booktitle={2024 IEEE/CVF Conference on Computer Vision and Pattern Recognition (CVPR)}, 
  title={Let's Think Outside the Box: Exploring Leap-of-Thought in Large Language Models with Creative Humor Generation}, 
  year={2024},
  volume={},
  number={},
  pages={13246-13257},
  doi={10.1109/CVPR52733.2024.01258}
}

@article{creative_index,
  publtype={informal},
  author={Ximing Lu and Melanie Sclar and Skyler Hallinan and Niloofar Mireshghallah and Jiacheng Liu and Seungju Han and Allyson Ettinger and Liwei Jiang and Khyathi Raghavi Chandu and Nouha Dziri and Yejin Choi},
  title={AI as Humanity's Salieri: Quantifying Linguistic Creativity of Language Models via Systematic Attribution of Machine Text against Web Text},
  year={2024},
  cdate={1704067200000},
  journal={CoRR},
  volume={abs/2410.04265},
  url={https://doi.org/10.48550/arXiv.2410.04265}
}

@inproceedings{
  probing_llm,
  title={Probing the {\textquotedblleft}Creativity{\textquotedblright} of Large Language Models:  Can models produce divergent semantic association?},
  author={Honghua Chen and Nai Ding},
  booktitle={The 2023 Conference on Empirical Methods in Natural Language Processing},
  year={2023},
  url={https://openreview.net/forum?id=BpibUh0aB3}
}

@article{boussioux2024crowdless,
  title={The crowdless future? Generative AI and creative problem-solving},
  author={Boussioux, L{\'e}onard and Lane, Jacqueline N and Zhang, Miaomiao and Jacimovic, Vladimir and Lakhani, Karim R},
  journal={Organization Science},
  volume={35},
  number={5},
  pages={1589--1607},
  year={2024},
  publisher={INFORMS}
}

@inproceedings{macgyver,
    title = "{M}ac{G}yver: Are Large Language Models Creative Problem Solvers?",
    author = "Tian, Yufei  and
      Ravichander, Abhilasha  and
      Qin, Lianhui  and
      Le Bras, Ronan  and
      Marjieh, Raja  and
      Peng, Nanyun  and
      Choi, Yejin  and
      Griffiths, Thomas  and
      Brahman, Faeze",
    editor = "Duh, Kevin  and
      Gomez, Helena  and
      Bethard, Steven",
    booktitle = "Proceedings of the 2024 Conference of the North American Chapter of the Association for Computational Linguistics: Human Language Technologies (Volume 1: Long Papers)",
    month = jun,
    year = "2024",
    address = "Mexico City, Mexico",
    publisher = "Association for Computational Linguistics",
    url = "https://aclanthology.org/2024.naacl-long.297/",
    doi = "10.18653/v1/2024.naacl-long.297",
    pages = "5303--5324"
}

@inproceedings{
  llm_for_research_idea,
  title={Can {LLM}s Generate Novel Research Ideas? A Large-Scale Human Study with 100+ {NLP} Researchers},
  author={Chenglei Si and Diyi Yang and Tatsunori Hashimoto},
  booktitle={The Thirteenth International Conference on Learning Representations},
  year={2025},
  url={https://openreview.net/forum?id=M23dTGWCZy}
}

@inproceedings{
  judgebench,
  title={JudgeBench: A Benchmark for Evaluating {LLM}-Based Judges},
  author={Sijun Tan and Siyuan Zhuang and Kyle Montgomery and William Yuan Tang and Alejandro Cuadron and Chenguang Wang and Raluca Popa and Ion Stoica},
  booktitle={The Thirteenth International Conference on Learning Representations},
  year={2025},
  url={https://openreview.net/forum?id=G0dksFayVq}
}

@misc{llm_as_a_judge_opportunities_challenges,
    title={From Generation to Judgment: Opportunities and Challenges of LLM-as-a-judge},
    author={Dawei Li and Bohan Jiang and Liangjie Huang and Alimohammad Beigi and Chengshuai Zhao and Zhen Tan and Amrita Bhattacharjee and Yuxuan Jiang and Canyu Chen and Tianhao Wu and Kai Shu and Lu Cheng and Huan Liu},
    year={2024},
    eprint={2411.16594},
    archivePrefix={arXiv},
    primaryClass={cs.AI}
}

@misc{llms_as_judges_comprehensive_survey_llm_based,
      title={LLMs-as-Judges: A Comprehensive Survey on LLM-based Evaluation Methods}, 
      author={Haitao Li and Qian Dong and Junjie Chen and Huixue Su and Yujia Zhou and Qingyao Ai and Ziyi Ye and Yiqun Liu},
      year={2024},
      eprint={2412.05579},
      archivePrefix={arXiv},
      primaryClass={cs.CL},
      url={https://arxiv.org/abs/2412.05579}, 
}

@book{holyoak2005cambridge,
  title={The Cambridge Handbook of Thinking and Reasoning},
  author={Holyoak, K.J. and Morrison, R.G.},
  isbn={9780521824170},
  lccn={2004016029},
  series={Cambridge Handbooks in Psychology},
  url={https://books.google.com/books?id=znbkHaC8QeMC},
  year={2005},
  publisher={Cambridge University Press}
}

@inproceedings{kwon2023efficient,
  title={Efficient Memory Management for Large Language Model Serving with PagedAttention},
  author={Woosuk Kwon and Zhuohan Li and Siyuan Zhuang and Ying Sheng and Lianmin Zheng and Cody Hao Yu and Joseph E. Gonzalez and Hao Zhang and Ion Stoica},
  booktitle={Proceedings of the ACM SIGOPS 29th Symposium on Operating Systems Principles},
  year={2023}
}

@BOOK{Boden1994-uh,
  title     = "Dimensions of Creativity",
  editor    = "Boden, Margaret A",
  publisher = "The MIT Press",
  month     =  jun,
  year      =  1994,
  language  = "en"
}

@ARTICLE{Olson2021-xq,
  title     = "Naming unrelated words predicts creativity",
  author    = "Olson, Jay A and Nahas, Johnny and Chmoulevitch, Denis and
               Cropper, Simon J and Webb, Margaret E",
  journal   = "Proc. Natl. Acad. Sci. U. S. A.",
  publisher = "Proceedings of the National Academy of Sciences",
  volume    =  118,
  number    =  25,
  pages     = "e2022340118",
  month     =  jun,
  year      =  2021,
  language  = "en"
}

@MISC{Guilford2012-kh,
  title     = "Alternate Uses",
  author    = "Guilford, J P and Christensen, Paul R and Merrifield, Philip R
               and Wilson, Robert C",
  publisher = "American Psychological Association (APA)",
  month     =  jun,
  year      =  2012,
  note      = "Title of the publication associated with this dataset: PsycTESTS
               Dataset"
}

@ARTICLE{Sternberg1991-qe,
  title     = "An investment theory of creativity and its development",
  author    = "Sternberg, R and Lubart, T",
  journal   = "Human Development",
  publisher = "S. Karger AG",
  volume    =  34,
  number    =  1,
  pages     = "1--31",
  month     =  jun,
  year      =  1991,
  language  = "en"
}

@BOOK{Finke1992-ke,
  title     = "Creative Cognition: Theory, research, and applications",
  author    = "Finke, Ronald A and Ward, Thomas B and Smith, Steven M",
  publisher = "The MIT Press",
  month     =  oct,
  year      =  1992,
  language  = "en"
}

@ARTICLE{Alabbasi2022-jz,
  title     = "What do educators need to know about the Torrance Tests of
               Creative Thinking: A comprehensive review",
  author    = "Alabbasi, Ahmed M Abdulla and Paek, Sue Hyeon and Kim, Daehyun
               and Cramond, Bonnie",
  journal   = "Front. Psychol.",
  publisher = "Frontiers Media SA",
  volume    =  13,
  pages     =  1000385,
  month     =  oct,
  year      =  2022,
  language  = "en"
}

@ARTICLE{Hu2002-nn,
  title     = "A scientific creativity test for secondary school students",
  author    = "Hu, Weiping and Adey, Philip",
  journal   = "Int. J. Sci. Educ.",
  publisher = "Informa UK Limited",
  volume    =  24,
  number    =  4,
  pages     = "389--403",
  month     =  apr,
  year      =  2002,
  language  = "en"
}

@INPROCEEDINGS{Gomez-Rodriguez2023-xp,
  title     = "A confederacy of models: A comprehensive evaluation of {LLMs} on
               creative writing",
  author    = "Gómez-Rodríguez, Carlos and Williams, Paul",
  booktitle = "Findings of the Association for Computational Linguistics: EMNLP
               2023",
  publisher = "Association for Computational Linguistics",
  address   = "Stroudsburg, PA, USA",
  pages     = "14504--14528",
  month     =  dec,
  year      =  2023
}

@INPROCEEDINGS{Tian2024-at,
  title     = "Are large language models capable of generating human-level
               narratives?",
  author    = "Tian, Yufei and Huang, Tenghao and Liu, Miri and Jiang, Derek and
               Spangher, Alexander and Chen, Muhao and May, Jonathan and Peng,
               Nanyun",
  booktitle = "Proceedings of the 2024 Conference on Empirical Methods in
               Natural Language Processing",
  publisher = "Association for Computational Linguistics",
  address   = "Stroudsburg, PA, USA",
  pages     = "17659--17681",
  month     =  nov,
  year      =  2024
}

@INPROCEEDINGS{Mohammad2018-gc,
  title     = "Obtaining reliable human ratings of valence, arousal, and
               dominance for 20,000 English words",
  author    = "Mohammad, Saif",
  booktitle = "Proceedings of the 56th Annual Meeting of the Association for
               Computational Linguistics (Volume 1: Long Papers)",
  publisher = "Association for Computational Linguistics",
  address   = "Stroudsburg, PA, USA",
  pages     = "174--184",
  year      =  2018
}

@INPROCEEDINGS{Tian2024-py,
  title     = "{MacGyver}: Are large language models creative problem solvers?",
  author    = "Tian, Yufei and Ravichander, Abhilasha and Qin, Lianhui and Le
               Bras, Ronan and Marjieh, Raja and Peng, Nanyun and Choi, Yejin
               and Griffiths, Thomas and Brahman, Faeze",
  booktitle = "Proceedings of the 2024 Conference of the North American Chapter
               of the Association for Computational Linguistics: Human Language
               Technologies (Volume 1: Long Papers)",
  publisher = "Association for Computational Linguistics",
  address   = "Stroudsburg, PA, USA",
  pages     = "5303--5324",
  year      =  2024
}

@ARTICLE{Smith2007-zs,
  title     = "Modeling the determinants and effects of creativity in
               advertising",
  author    = "Smith, Robert E and MacKenzie, Scott B and Yang, Xiaojing and
               Buchholz, Laura M and Darley, William K",
  journal   = "Mark. Sci.",
  publisher = "Institute for Operations Research and the Management Sciences
               (INFORMS)",
  volume    =  26,
  number    =  6,
  pages     = "819--833",
  month     =  nov,
  year      =  2007,
  language  = "en"
}

@ARTICLE{Rosengren2020-ij,
  title     = "A meta-analysis of when and how advertising creativity works",
  author    = "Rosengren, Sara and Eisend, Martin and Koslow, Scott and Dahlen,
               Micael",
  journal   = "J. Mark.",
  publisher = "SAGE Publications",
  volume    =  84,
  number    =  6,
  pages     = "39--56",
  month     =  nov,
  year      =  2020,
  language  = "en"
}

@ARTICLE{Said-Metwaly2017-eq,
  title    = "Approaches to Measuring Creativity: A Systematic Literature Review",
  author   = "Said-Metwaly, Sameh and Van den Noortgate, Wim and Kyndt, Eva",
  journal  = "Creativity. Theories – Research - Applications",
  volume   =  4,
  number   =  2,
  pages    = "238--275",
  month    =  dec,
  year     =  2017
}

@ARTICLE{El-Murad2004-ue,
  title     = "The Definition and Measurement of Creativity: What Do We Know?",
  author    = "El-Murad, Jaafar and West, Douglas C",
  journal   = "J. Advert. Res.",
  publisher = "World Advertising Research Center",
  volume    =  44,
  number    =  02,
  pages     = "188--201",
  month     =  jun,
  year      =  2004
}

@INPROCEEDINGS{Yiming2025-xs,
  title     = "{NoveltyBench}: Evaluating Creativity and Diversity in Language
               Models",
  author    = "Zhang, Yiming and Diddee, Harshita and Holm, Susan and Liu,
               Hanchen and Liu, Xinyue and Samuel, Vinay and Wang, Barry and
               Ippolito, Daphne",
  booktitle = "Second Conference on Language Modeling",
  month     =  aug,
  year      =  2025
}

@ARTICLE{N-Lane2024-hp,
  title    = "The Narrative {AI} Advantage? A Field Experiment on Generative
              {AI}-Augmented Evaluations of Early-Stage Innovations",
  author   = "N. Lane, Jacqueline and Boussioux, Leonard and Ayoubi, Charles and
              Chen, Ying Hao and Lin, Camila and Spens, Rebecca and Wagh, Pooja
              and Wang, Pei-Hsin",
  journal  = "Social Science Research Network",
  month    =  aug,
  year     =  2024
}

@inproceedings{Zheng2023-pm,
author = {Zheng, Lianmin and Chiang, Wei-Lin and Sheng, Ying and Zhuang, Siyuan and Wu, Zhanghao and Zhuang, Yonghao and Lin, Zi and Li, Zhuohan and Li, Dacheng and Xing, Eric P. and Zhang, Hao and Gonzalez, Joseph E. and Stoica, Ion},
title = {Judging LLM-as-a-judge with MT-bench and Chatbot Arena},
year = {2023},
publisher = {Curran Associates Inc.},
address = {Red Hook, NY, USA},
booktitle = {Proceedings of the 37th International Conference on Neural Information Processing Systems},
articleno = {2020},
numpages = {29},
location = {New Orleans, LA, USA},
series = {NIPS '23}
}

@INPROCEEDINGS{Pennington2014-ea,
  title     = "{GloVe}: Global Vectors for Word Representation",
  author    = "Pennington, Jeffrey and Socher, Richard and Manning, Christopher
               D",
  booktitle = "Proceedings of the 2014 Conference on Empirical Methods in
               Natural Language Processing (EMNLP)",
  pages     = "1532--1543",
  month     =  oct,
  year      =  2014
}

@inproceedings{Zhang2019-ab,
title={BERTScore: Evaluating Text Generation with BERT},
author={Tianyi Zhang* and Varsha Kishore* and Felix Wu* and Kilian Q. Weinberger and Yoav Artzi},
booktitle={International Conference on Learning Representations},
year={2020},
url={https://openreview.net/forum?id=SkeHuCVFDr}
}

@INPROCEEDINGS{Jung2024-xz,
  title     = "Trust or Escalate: {LLM} Judges with Provable Guarantees for
               Human Agreement",
  author    = "Jung, Jaehun and Brahman, Faeze and Choi, Yejin",
  booktitle = "The Thirteenth International Conference on Learning
               Representations",
  month     =  oct,
  year      =  2024
}

@ARTICLE{DeepSeek-AI2024-mv,
  title         = "{DeepSeek}-{V3} Technical Report",
  author        = "{DeepSeek-AI} and Liu, Aixin and Feng, Bei and Xue, Bing and
                   Wang, Bingxuan and Wu, Bochao and Lu, Chengda and Zhao,
                   Chenggang and Deng, Chengqi and Zhang, Chenyu and Ruan, Chong
                   and Dai, Damai and Guo, Daya and Yang, Dejian and Chen, Deli
                   and Ji, Dongjie and Li, Erhang and Lin, Fangyun and Dai,
                   Fucong and Luo, Fuli and Hao, Guangbo and Chen, Guanting and
                   Li, Guowei and Zhang, H and Bao, Han and Xu, Hanwei and Wang,
                   Haocheng and Zhang, Haowei and Ding, Honghui and Xin, Huajian
                   and Gao, Huazuo and Li, Hui and Qu, Hui and Cai, J L and
                   Liang, Jian and Guo, Jianzhong and Ni, Jiaqi and Li, Jiashi
                   and Wang, Jiawei and Chen, Jin and Chen, Jingchang and Yuan,
                   Jingyang and Qiu, Junjie and Li, Junlong and Song, Junxiao
                   and Dong, Kai and Hu, Kai and Gao, Kaige and Guan, Kang and
                   Huang, Kexin and Yu, Kuai and Wang, Lean and Zhang, Lecong
                   and Xu, Lei and Xia, Leyi and Zhao, Liang and Wang, Litong
                   and Zhang, Liyue and Li, Meng and Wang, Miaojun and Zhang,
                   Mingchuan and Zhang, Minghua and Tang, Minghui and Li,
                   Mingming and Tian, Ning and Huang, Panpan and Wang, Peiyi and
                   Zhang, Peng and Wang, Qiancheng and Zhu, Qihao and Chen,
                   Qinyu and Du, Qiushi and Chen, R J and Jin, R L and Ge, Ruiqi
                   and Zhang, Ruisong and Pan, Ruizhe and Wang, Runji and Xu,
                   Runxin and Zhang, Ruoyu and Chen, Ruyi and Li, S S and Lu,
                   Shanghao and Zhou, Shangyan and Chen, Shanhuang and Wu,
                   Shaoqing and Ye, Shengfeng and Ye, Shengfeng and Ma, Shirong
                   and Wang, Shiyu and Zhou, Shuang and Yu, Shuiping and Zhou,
                   Shunfeng and Pan, Shuting and Wang, T and Yun, Tao and Pei,
                   Tian and Sun, Tianyu and Xiao, W L and Zeng, Wangding and
                   Zhao, Wanjia and An, Wei and Liu, Wen and Liang, Wenfeng and
                   Gao, Wenjun and Yu, Wenqin and Zhang, Wentao and Li, X Q and
                   Jin, Xiangyue and Wang, Xianzu and Bi, Xiao and Liu, Xiaodong
                   and Wang, Xiaohan and Shen, Xiaojin and Chen, Xiaokang and
                   Zhang, Xiaokang and Chen, Xiaosha and Nie, Xiaotao and Sun,
                   Xiaowen and Wang, Xiaoxiang and Cheng, Xin and Liu, Xin and
                   Xie, Xin and Liu, Xingchao and Yu, Xingkai and Song, Xinnan
                   and Shan, Xinxia and Zhou, Xinyi and Yang, Xinyu and Li,
                   Xinyuan and Su, Xuecheng and Lin, Xuheng and Li, Y K and
                   Wang, Y Q and Wei, Y X and Zhu, Y X and Zhang, Yang and Xu,
                   Yanhong and Xu, Yanhong and Huang, Yanping and Li, Yao and
                   Zhao, Yao and Sun, Yaofeng and Li, Yaohui and Wang, Yaohui
                   and Yu, Yi and Zheng, Yi and Zhang, Yichao and Shi, Yifan and
                   Xiong, Yiliang and He, Ying and Tang, Ying and Piao, Yishi
                   and Wang, Yisong and Tan, Yixuan and Ma, Yiyang and Liu,
                   Yiyuan and Guo, Yongqiang and Wu, Yu and Ou, Yuan and Zhu,
                   Yuchen and Wang, Yuduan and Gong, Yue and Zou, Yuheng and He,
                   Yujia and Zha, Yukun and Xiong, Yunfan and Ma, Yunxian and
                   Yan, Yuting and Luo, Yuxiang and You, Yuxiang and Liu, Yuxuan
                   and Zhou, Yuyang and Wu, Z F and Ren, Z Z and Ren, Zehui and
                   Sha, Zhangli and Fu, Zhe and Xu, Zhean and Huang, Zhen and
                   Zhang, Zhen and Xie, Zhenda and Zhang, Zhengyan and Hao,
                   Zhewen and Gou, Zhibin and Ma, Zhicheng and Yan, Zhigang and
                   Shao, Zhihong and Xu, Zhipeng and Wu, Zhiyu and Zhang,
                   Zhongyu and Li, Zhuoshu and Gu, Zihui and Zhu, Zijia and Liu,
                   Zijun and Li, Zilin and Xie, Ziwei and Song, Ziyang and Gao,
                   Ziyi and Pan, Zizheng",
  journal       = "arXiv [cs.CL]",
  month         =  dec,
  year          =  2024,
  archivePrefix = "arXiv",
  primaryClass  = "cs.CL"
}

@ARTICLE{DeepSeek-AI2025-wm,
  title         = "{DeepSeek}-{R1}: Incentivizing Reasoning Capability in {LLMs}
                   via Reinforcement Learning",
  author        = "{DeepSeek-AI} and Guo, Daya and Yang, Dejian and Zhang,
                   Haowei and Song, Junxiao and Zhang, Ruoyu and Xu, Runxin and
                   Zhu, Qihao and Ma, Shirong and Wang, Peiyi and Bi, Xiao and
                   Zhang, Xiaokang and Yu, Xingkai and Wu, Yu and Wu, Z F and
                   Gou, Zhibin and Shao, Zhihong and Li, Zhuoshu and Gao, Ziyi
                   and Liu, Aixin and Xue, Bing and Wang, Bingxuan and Wu,
                   Bochao and Feng, Bei and Lu, Chengda and Zhao, Chenggang and
                   Deng, Chengqi and Zhang, Chenyu and Ruan, Chong and Dai,
                   Damai and Chen, Deli and Ji, Dongjie and Li, Erhang and Lin,
                   Fangyun and Dai, Fucong and Luo, Fuli and Hao, Guangbo and
                   Chen, Guanting and Li, Guowei and Zhang, H and Bao, Han and
                   Xu, Hanwei and Wang, Haocheng and Ding, Honghui and Xin,
                   Huajian and Gao, Huazuo and Qu, Hui and Li, Hui and Guo,
                   Jianzhong and Li, Jiashi and Wang, Jiawei and Chen, Jingchang
                   and Yuan, Jingyang and Qiu, Junjie and Li, Junlong and Cai, J
                   L and Ni, Jiaqi and Liang, Jian and Chen, Jin and Dong, Kai
                   and Hu, Kai and Gao, Kaige and Guan, Kang and Huang, Kexin
                   and Yu, Kuai and Wang, Lean and Zhang, Lecong and Zhao, Liang
                   and Wang, Litong and Zhang, Liyue and Xu, Lei and Xia, Leyi
                   and Zhang, Mingchuan and Zhang, Minghua and Tang, Minghui and
                   Li, Meng and Wang, Miaojun and Li, Mingming and Tian, Ning
                   and Huang, Panpan and Zhang, Peng and Wang, Qiancheng and
                   Chen, Qinyu and Du, Qiushi and Ge, Ruiqi and Zhang, Ruisong
                   and Pan, Ruizhe and Wang, Runji and Chen, R J and Jin, R L
                   and Chen, Ruyi and Lu, Shanghao and Zhou, Shangyan and Chen,
                   Shanhuang and Ye, Shengfeng and Wang, Shiyu and Yu, Shuiping
                   and Zhou, Shunfeng and Pan, Shuting and Li, S S and Zhou,
                   Shuang and Wu, Shaoqing and Ye, Shengfeng and Yun, Tao and
                   Pei, Tian and Sun, Tianyu and Wang, T and Zeng, Wangding and
                   Zhao, Wanjia and Liu, Wen and Liang, Wenfeng and Gao, Wenjun
                   and Yu, Wenqin and Zhang, Wentao and Xiao, W L and An, Wei
                   and Liu, Xiaodong and Wang, Xiaohan and Chen, Xiaokang and
                   Nie, Xiaotao and Cheng, Xin and Liu, Xin and Xie, Xin and
                   Liu, Xingchao and Yang, Xinyu and Li, Xinyuan and Su,
                   Xuecheng and Lin, Xuheng and Li, X Q and Jin, Xiangyue and
                   Shen, Xiaojin and Chen, Xiaosha and Sun, Xiaowen and Wang,
                   Xiaoxiang and Song, Xinnan and Zhou, Xinyi and Wang, Xianzu
                   and Shan, Xinxia and Li, Y K and Wang, Y Q and Wei, Y X and
                   Zhang, Yang and Xu, Yanhong and Li, Yao and Zhao, Yao and
                   Sun, Yaofeng and Wang, Yaohui and Yu, Yi and Zhang, Yichao
                   and Shi, Yifan and Xiong, Yiliang and He, Ying and Piao,
                   Yishi and Wang, Yisong and Tan, Yixuan and Ma, Yiyang and
                   Liu, Yiyuan and Guo, Yongqiang and Ou, Yuan and Wang, Yuduan
                   and Gong, Yue and Zou, Yuheng and He, Yujia and Xiong, Yunfan
                   and Luo, Yuxiang and You, Yuxiang and Liu, Yuxuan and Zhou,
                   Yuyang and Zhu, Y X and Xu, Yanhong and Huang, Yanping and
                   Li, Yaohui and Zheng, Yi and Zhu, Yuchen and Ma, Yunxian and
                   Tang, Ying and Zha, Yukun and Yan, Yuting and Ren, Z Z and
                   Ren, Zehui and Sha, Zhangli and Fu, Zhe and Xu, Zhean and
                   Xie, Zhenda and Zhang, Zhengyan and Hao, Zhewen and Ma,
                   Zhicheng and Yan, Zhigang and Wu, Zhiyu and Gu, Zihui and
                   Zhu, Zijia and Liu, Zijun and Li, Zilin and Xie, Ziwei and
                   Song, Ziyang and Pan, Zizheng and Huang, Zhen and Xu, Zhipeng
                   and Zhang, Zhongyu and Zhang, Zhen",
  journal       = "arXiv [cs.CL]",
  month         =  jan,
  year          =  2025,
  archivePrefix = "arXiv",
  primaryClass  = "cs.CL"
}

@MISC{Kavukcuoglu2025-wr,
  title        = "Gemini 2.5: Our most intelligent {AI} model",
  author       = "Kavukcuoglu, Koray",
  booktitle    = "Google",
  month        =  mar,
  year         =  2025,
  howpublished = "\url{https://blog.google/technology/google-deepmind/gemini-model-thinking-updates-march-2025/}",
  note         = "Accessed: 2025-4-30",
  language     = "en"
}

@ARTICLE{Qwen2024-rw,
  title         = "{Qwen2}.5 Technical Report",
  author        = "{Qwen} and Yang, An and Yang, Baosong and Zhang, Beichen and
                   Hui, Binyuan and Zheng, Bo and Yu, Bowen and Li, Chengyuan
                   and Liu, Dayiheng and Huang, Fei and Wei, Haoran and Lin,
                   Huan and Yang, Jian and Tu, Jianhong and Zhang, Jianwei and
                   Yang, Jianxin and Yang, Jiaxi and Zhou, Jingren and Lin,
                   Junyang and Dang, Kai and Lu, Keming and Bao, Keqin and Yang,
                   Kexin and Yu, Le and Li, Mei and Xue, Mingfeng and Zhang, Pei
                   and Zhu, Qin and Men, Rui and Lin, Runji and Li, Tianhao and
                   Tang, Tianyi and Xia, Tingyu and Ren, Xingzhang and Ren,
                   Xuancheng and Fan, Yang and Su, Yang and Zhang, Yichang and
                   Wan, Yu and Liu, Yuqiong and Cui, Zeyu and Zhang, Zhenru and
                   Qiu, Zihan",
  journal       = "arXiv [cs.CL]",
  month         =  dec,
  year          =  2024,
  archivePrefix = "arXiv",
  primaryClass  = "cs.CL"
}

@ARTICLE{Hui2024-qw,
  title         = "{Qwen2}.5-Coder Technical Report",
  author        = "Hui, Binyuan and Yang, Jian and Cui, Zeyu and Yang, Jiaxi and
                   Liu, Dayiheng and Zhang, Lei and Liu, Tianyu and Zhang,
                   Jiajun and Yu, Bowen and Lu, Keming and Dang, Kai and Fan,
                   Yang and Zhang, Yichang and Yang, An and Men, Rui and Huang,
                   Fei and Zheng, Bo and Miao, Yibo and Quan, Shanghaoran and
                   Feng, Yunlong and Ren, Xingzhang and Ren, Xuancheng and Zhou,
                   Jingren and Lin, Junyang",
  journal       = "arXiv [cs.CL]",
  month         =  sep,
  year          =  2024,
  archivePrefix = "arXiv",
  primaryClass  = "cs.CL"
}

@INPROCEEDINGS{Groeneveld2024-pt,
  title     = "{OLMo}: Accelerating the science of language models",
  author    = "Groeneveld, Dirk and Beltagy, Iz and Walsh, Evan and Bhagia,
               Akshita and Kinney, Rodney and Tafjord, Oyvind and Jha, Ananya
               and Ivison, Hamish and Magnusson, Ian and Wang, Yizhong and
               Arora, Shane and Atkinson, David and Authur, Russell and Chandu,
               Khyathi and Cohan, Arman and Dumas, Jennifer and Elazar, Yanai
               and Gu, Yuling and Hessel, Jack and Khot, Tushar and Merrill,
               William and Morrison, Jacob and Muennighoff, Niklas and Naik,
               Aakanksha and Nam, Crystal and Peters, Matthew and Pyatkin,
               Valentina and Ravichander, Abhilasha and Schwenk, Dustin and
               Shah, Saurabh and Smith, William and Strubell, Emma and
               Subramani, Nishant and Wortsman, Mitchell and Dasigi, Pradeep and
               Lambert, Nathan and Richardson, Kyle and Zettlemoyer, Luke and
               Dodge, Jesse and Lo, Kyle and Soldaini, Luca and Smith, Noah and
               Hajishirzi, Hannaneh",
  booktitle = "Proceedings of the 62nd Annual Meeting of the Association for
               Computational Linguistics (Volume 1: Long Papers)",
  publisher = "Association for Computational Linguistics",
  address   = "Stroudsburg, PA, USA",
  pages     = "15789--15809",
  year      =  2024
}

@ARTICLE{Grattafiori2024-yn,
  title         = "The Llama 3 herd of models",
  author        = "Grattafiori, Aaron and Dubey, Abhimanyu and Jauhri, Abhinav
                   and Pandey, Abhinav and Kadian, Abhishek and Al-Dahle, Ahmad
                   and Letman, Aiesha and Mathur, Akhil and Schelten, Alan and
                   Vaughan, Alex and Yang, Amy and Fan, Angela and Goyal,
                   Anirudh and Hartshorn, Anthony and Yang, Aobo and Mitra,
                   Archi and Sravankumar, Archie and Korenev, Artem and
                   Hinsvark, Arthur and Rao, Arun and Zhang, Aston and
                   Rodriguez, Aurelien and Gregerson, Austen and Spataru, Ava
                   and Roziere, Baptiste and Biron, Bethany and Tang, Binh and
                   Chern, Bobbie and Caucheteux, Charlotte and Nayak, Chaya and
                   Bi, Chloe and Marra, Chris and McConnell, Chris and Keller,
                   Christian and Touret, Christophe and Wu, Chunyang and Wong,
                   Corinne and Ferrer, Cristian Canton and Nikolaidis, Cyrus and
                   Allonsius, Damien and Song, Daniel and Pintz, Danielle and
                   Livshits, Danny and Wyatt, Danny and Esiobu, David and
                   Choudhary, Dhruv and Mahajan, Dhruv and Garcia-Olano, Diego
                   and Perino, Diego and Hupkes, Dieuwke and Lakomkin, Egor and
                   AlBadawy, Ehab and Lobanova, Elina and Dinan, Emily and
                   Smith, Eric Michael and Radenovic, Filip and Guzmán,
                   Francisco and Zhang, Frank and Synnaeve, Gabriel and Lee,
                   Gabrielle and Anderson, Georgia Lewis and Thattai, Govind and
                   Nail, Graeme and Mialon, Gregoire and Pang, Guan and
                   Cucurell, Guillem and Nguyen, Hailey and Korevaar, Hannah and
                   Xu, Hu and Touvron, Hugo and Zarov, Iliyan and Ibarra, Imanol
                   Arrieta and Kloumann, Isabel and Misra, Ishan and Evtimov,
                   Ivan and Zhang, Jack and Copet, Jade and Lee, Jaewon and
                   Geffert, Jan and Vranes, Jana and Park, Jason and Mahadeokar,
                   Jay and Shah, Jeet and van der Linde, Jelmer and Billock,
                   Jennifer and Hong, Jenny and Lee, Jenya and Fu, Jeremy and
                   Chi, Jianfeng and Huang, Jianyu and Liu, Jiawen and Wang, Jie
                   and Yu, Jiecao and Bitton, Joanna and Spisak, Joe and Park,
                   Jongsoo and Rocca, Joseph and Johnstun, Joshua and Saxe,
                   Joshua and Jia, Junteng and Alwala, Kalyan Vasuden and
                   Prasad, Karthik and Upasani, Kartikeya and Plawiak, Kate and
                   Li, Ke and Heafield, Kenneth and Stone, Kevin and El-Arini,
                   Khalid and Iyer, Krithika and Malik, Kshitiz and Chiu,
                   Kuenley and Bhalla, Kunal and Lakhotia, Kushal and
                   Rantala-Yeary, Lauren and van der Maaten, Laurens and Chen,
                   Lawrence and Tan, Liang and Jenkins, Liz and Martin, Louis
                   and Madaan, Lovish and Malo, Lubo and Blecher, Lukas and
                   Landzaat, Lukas and de Oliveira, Luke and Muzzi, Madeline and
                   Pasupuleti, Mahesh and Singh, Mannat and Paluri, Manohar and
                   Kardas, Marcin and Tsimpoukelli, Maria and Oldham, Mathew and
                   Rita, Mathieu and Pavlova, Maya and Kambadur, Melanie and
                   Lewis, Mike and Si, Min and Singh, Mitesh Kumar and Hassan,
                   Mona and Goyal, Naman and Torabi, Narjes and Bashlykov,
                   Nikolay and Bogoychev, Nikolay and Chatterji, Niladri and
                   Zhang, Ning and Duchenne, Olivier and Çelebi, Onur and
                   Alrassy, Patrick and Zhang, Pengchuan and Li, Pengwei and
                   Vasic, Petar and Weng, Peter and Bhargava, Prajjwal and
                   Dubal, Pratik and Krishnan, Praveen and Koura, Punit Singh
                   and Xu, Puxin and He, Qing and Dong, Qingxiao and Srinivasan,
                   Ragavan and Ganapathy, Raj and Calderer, Ramon and Cabral,
                   Ricardo Silveira and Stojnic, Robert and Raileanu, Roberta
                   and Maheswari, Rohan and Girdhar, Rohit and Patel, Rohit and
                   Sauvestre, Romain and Polidoro, Ronnie and Sumbaly, Roshan
                   and Taylor, Ross and Silva, Ruan and Hou, Rui and Wang, Rui
                   and Hosseini, Saghar and Chennabasappa, Sahana and Singh,
                   Sanjay and Bell, Sean and Kim, Seohyun Sonia and Edunov,
                   Sergey and Nie, Shaoliang and Narang, Sharan and Raparthy,
                   Sharath and Shen, Sheng and Wan, Shengye and Bhosale, Shruti
                   and Zhang, Shun and Vandenhende, Simon and Batra, Soumya and
                   Whitman, Spencer and Sootla, Sten and Collot, Stephane and
                   Gururangan, Suchin and Borodinsky, Sydney and Herman, Tamar
                   and Fowler, Tara and Sheasha, Tarek and Georgiou, Thomas and
                   Scialom, Thomas and Speckbacher, Tobias and Mihaylov, Todor
                   and Xiao, Tong and Karn, Ujjwal and Goswami, Vedanuj and
                   Gupta, Vibhor and Ramanathan, Vignesh and Kerkez, Viktor and
                   Gonguet, Vincent and Do, Virginie and Vogeti, Vish and
                   Albiero, Vítor and Petrovic, Vladan and Chu, Weiwei and
                   Xiong, Wenhan and Fu, Wenyin and Meers, Whitney and Martinet,
                   Xavier and Wang, Xiaodong and Wang, Xiaofang and Tan,
                   Xiaoqing Ellen and Xia, Xide and Xie, Xinfeng and Jia, Xuchao
                   and Wang, Xuewei and Goldschlag, Yaelle and Gaur, Yashesh and
                   Babaei, Yasmine and Wen, Yi and Song, Yiwen and Zhang, Yuchen
                   and Li, Yue and Mao, Yuning and Coudert, Zacharie Delpierre
                   and Yan, Zheng and Chen, Zhengxing and Papakipos, Zoe and
                   Singh, Aaditya and Srivastava, Aayushi and Jain, Abha and
                   Kelsey, Adam and Shajnfeld, Adam and Gangidi, Adithya and
                   Victoria, Adolfo and Goldstand, Ahuva and Menon, Ajay and
                   Sharma, Ajay and Boesenberg, Alex and Baevski, Alexei and
                   Feinstein, Allie and Kallet, Amanda and Sangani, Amit and
                   Teo, Amos and Yunus, Anam and Lupu, Andrei and Alvarado,
                   Andres and Caples, Andrew and Gu, Andrew and Ho, Andrew and
                   Poulton, Andrew and Ryan, Andrew and Ramchandani, Ankit and
                   Dong, Annie and Franco, Annie and Goyal, Anuj and Saraf,
                   Aparajita and Chowdhury, Arkabandhu and Gabriel, Ashley and
                   Bharambe, Ashwin and Eisenman, Assaf and Yazdan, Azadeh and
                   James, Beau and Maurer, Ben and Leonhardi, Benjamin and
                   Huang, Bernie and Loyd, Beth and De Paola, Beto and
                   Paranjape, Bhargavi and Liu, Bing and Wu, Bo and Ni, Boyu and
                   Hancock, Braden and Wasti, Bram and Spence, Brandon and
                   Stojkovic, Brani and Gamido, Brian and Montalvo, Britt and
                   Parker, Carl and Burton, Carly and Mejia, Catalina and Liu,
                   Ce and Wang, Changhan and Kim, Changkyu and Zhou, Chao and
                   Hu, Chester and Chu, Ching-Hsiang and Cai, Chris and Tindal,
                   Chris and Feichtenhofer, Christoph and Gao, Cynthia and
                   Civin, Damon and Beaty, Dana and Kreymer, Daniel and Li,
                   Daniel and Adkins, David and Xu, David and Testuggine, Davide
                   and David, Delia and Parikh, Devi and Liskovich, Diana and
                   Foss, Didem and Wang, Dingkang and Le, Duc and Holland,
                   Dustin and Dowling, Edward and Jamil, Eissa and Montgomery,
                   Elaine and Presani, Eleonora and Hahn, Emily and Wood, Emily
                   and Le, Eric-Tuan and Brinkman, Erik and Arcaute, Esteban and
                   Dunbar, Evan and Smothers, Evan and Sun, Fei and Kreuk, Felix
                   and Tian, Feng and Kokkinos, Filippos and Ozgenel, Firat and
                   Caggioni, Francesco and Kanayet, Frank and Seide, Frank and
                   Florez, Gabriela Medina and Schwarz, Gabriella and Badeer,
                   Gada and Swee, Georgia and Halpern, Gil and Herman, Grant and
                   Sizov, Grigory and {Guangyi} and {Zhang} and
                   Lakshminarayanan, Guna and Inan, Hakan and Shojanazeri, Hamid
                   and Zou, Han and Wang, Hannah and Zha, Hanwen and Habeeb,
                   Haroun and Rudolph, Harrison and Suk, Helen and Aspegren,
                   Henry and Goldman, Hunter and Zhan, Hongyuan and Damlaj,
                   Ibrahim and Molybog, Igor and Tufanov, Igor and Leontiadis,
                   Ilias and Veliche, Irina-Elena and Gat, Itai and Weissman,
                   Jake and Geboski, James and Kohli, James and Lam, Janice and
                   Asher, Japhet and Gaya, Jean-Baptiste and Marcus, Jeff and
                   Tang, Jeff and Chan, Jennifer and Zhen, Jenny and
                   Reizenstein, Jeremy and Teboul, Jeremy and Zhong, Jessica and
                   Jin, Jian and Yang, Jingyi and Cummings, Joe and Carvill, Jon
                   and Shepard, Jon and McPhie, Jonathan and Torres, Jonathan
                   and Ginsburg, Josh and Wang, Junjie and Wu, Kai and U, Kam
                   Hou and Saxena, Karan and Khandelwal, Kartikay and Zand,
                   Katayoun and Matosich, Kathy and Veeraraghavan, Kaushik and
                   Michelena, Kelly and Li, Keqian and Jagadeesh, Kiran and
                   Huang, Kun and Chawla, Kunal and Huang, Kyle and Chen, Lailin
                   and Garg, Lakshya and A, Lavender and Silva, Leandro and
                   Bell, Lee and Zhang, Lei and Guo, Liangpeng and Yu, Licheng
                   and Moshkovich, Liron and Wehrstedt, Luca and Khabsa, Madian
                   and Avalani, Manav and Bhatt, Manish and Mankus, Martynas and
                   Hasson, Matan and Lennie, Matthew and Reso, Matthias and
                   Groshev, Maxim and Naumov, Maxim and Lathi, Maya and
                   Keneally, Meghan and Liu, Miao and Seltzer, Michael L and
                   Valko, Michal and Restrepo, Michelle and Patel, Mihir and
                   Vyatskov, Mik and Samvelyan, Mikayel and Clark, Mike and
                   Macey, Mike and Wang, Mike and Hermoso, Miquel Jubert and
                   Metanat, Mo and Rastegari, Mohammad and Bansal, Munish and
                   Santhanam, Nandhini and Parks, Natascha and White, Natasha
                   and Bawa, Navyata and Singhal, Nayan and Egebo, Nick and
                   Usunier, Nicolas and Mehta, Nikhil and Laptev, Nikolay
                   Pavlovich and Dong, Ning and Cheng, Norman and Chernoguz,
                   Oleg and Hart, Olivia and Salpekar, Omkar and Kalinli, Ozlem
                   and Kent, Parkin and Parekh, Parth and Saab, Paul and Balaji,
                   Pavan and Rittner, Pedro and Bontrager, Philip and Roux,
                   Pierre and Dollar, Piotr and Zvyagina, Polina and
                   Ratanchandani, Prashant and Yuvraj, Pritish and Liang, Qian
                   and Alao, Rachad and Rodriguez, Rachel and Ayub, Rafi and
                   Murthy, Raghotham and Nayani, Raghu and Mitra, Rahul and
                   Parthasarathy, Rangaprabhu and Li, Raymond and Hogan,
                   Rebekkah and Battey, Robin and Wang, Rocky and Howes, Russ
                   and Rinott, Ruty and Mehta, Sachin and Siby, Sachin and
                   Bondu, Sai Jayesh and Datta, Samyak and Chugh, Sara and Hunt,
                   Sara and Dhillon, Sargun and Sidorov, Sasha and Pan, Satadru
                   and Mahajan, Saurabh and Verma, Saurabh and Yamamoto, Seiji
                   and Ramaswamy, Sharadh and Lindsay, Shaun and Lindsay, Shaun
                   and Feng, Sheng and Lin, Shenghao and Zha, Shengxin Cindy and
                   Patil, Shishir and Shankar, Shiva and Zhang, Shuqiang and
                   Zhang, Shuqiang and Wang, Sinong and Agarwal, Sneha and
                   Sajuyigbe, Soji and Chintala, Soumith and Max, Stephanie and
                   Chen, Stephen and Kehoe, Steve and Satterfield, Steve and
                   Govindaprasad, Sudarshan and Gupta, Sumit and Deng, Summer
                   and Cho, Sungmin and Virk, Sunny and Subramanian, Suraj and
                   Choudhury, Sy and Goldman, Sydney and Remez, Tal and Glaser,
                   Tamar and Best, Tamara and Koehler, Thilo and Robinson,
                   Thomas and Li, Tianhe and Zhang, Tianjun and Matthews, Tim
                   and Chou, Timothy and Shaked, Tzook and Vontimitta, Varun and
                   Ajayi, Victoria and Montanez, Victoria and Mohan, Vijai and
                   Kumar, Vinay Satish and Mangla, Vishal and Ionescu, Vlad and
                   Poenaru, Vlad and Mihailescu, Vlad Tiberiu and Ivanov,
                   Vladimir and Li, Wei and Wang, Wenchen and Jiang, Wenwen and
                   Bouaziz, Wes and Constable, Will and Tang, Xiaocheng and Wu,
                   Xiaojian and Wang, Xiaolan and Wu, Xilun and Gao, Xinbo and
                   Kleinman, Yaniv and Chen, Yanjun and Hu, Ye and Jia, Ye and
                   Qi, Ye and Li, Yenda and Zhang, Yilin and Zhang, Ying and
                   Adi, Yossi and Nam, Youngjin and {Yu} and {Wang} and Zhao, Yu
                   and Hao, Yuchen and Qian, Yundi and Li, Yunlu and He, Yuzi
                   and Rait, Zach and DeVito, Zachary and Rosnbrick, Zef and
                   Wen, Zhaoduo and Yang, Zhenyu and Zhao, Zhiwei and Ma, Zhiyu",
  journal       = "arXiv [cs.AI]",
  month         =  jul,
  year          =  2024,
  archivePrefix = "arXiv",
  primaryClass  = "cs.AI"
}

@ARTICLE{Jiang2024-uo,
  title         = "Mixtral of Experts",
  author        = "Jiang, Albert Q and Sablayrolles, Alexandre and Roux, Antoine
                   and Mensch, Arthur and Savary, Blanche and Bamford, Chris and
                   Chaplot, Devendra Singh and Casas, Diego de las and Hanna,
                   Emma Bou and Bressand, Florian and Lengyel, Gianna and Bour,
                   Guillaume and Lample, Guillaume and Lavaud, Lélio Renard and
                   Saulnier, Lucile and Lachaux, Marie-Anne and Stock, Pierre
                   and Subramanian, Sandeep and Yang, Sophia and Antoniak,
                   Szymon and Scao, Teven Le and Gervet, Théophile and Lavril,
                   Thibaut and Wang, Thomas and Lacroix, Timothée and Sayed,
                   William El",
  journal       = "arXiv [cs.LG]",
  month         =  jan,
  year          =  2024,
  archivePrefix = "arXiv",
  primaryClass  = "cs.LG"
}

@ARTICLE{Jiang2023-xi,
  title         = "Mistral {7B}",
  author        = "Jiang, Albert Q and Sablayrolles, Alexandre and Mensch,
                   Arthur and Bamford, Chris and Chaplot, Devendra Singh and
                   Casas, Diego de las and Bressand, Florian and Lengyel, Gianna
                   and Lample, Guillaume and Saulnier, Lucile and Lavaud, Lélio
                   Renard and Lachaux, Marie-Anne and Stock, Pierre and Scao,
                   Teven Le and Lavril, Thibaut and Wang, Thomas and Lacroix,
                   Timothée and Sayed, William El",
  journal       = "arXiv [cs.CL]",
  month         =  oct,
  year          =  2023,
  archivePrefix = "arXiv",
  primaryClass  = "cs.CL"
}

@inproceedings{pennington-etal-2014-glove,
    title = "{G}lo{V}e: Global Vectors for Word Representation",
    author = "Pennington, Jeffrey  and
      Socher, Richard  and
      Manning, Christopher",
    editor = "Moschitti, Alessandro  and
      Pang, Bo  and
      Daelemans, Walter",
    booktitle = "Proceedings of the 2014 Conference on Empirical Methods in Natural Language Processing ({EMNLP})",
    month = oct,
    year = "2014",
    address = "Doha, Qatar",
    publisher = "Association for Computational Linguistics",
    url = "https://aclanthology.org/D14-1162/",
    doi = "10.3115/v1/D14-1162",
    pages = "1532--1543"
}

@INPROCEEDINGS{Lu2024-io,
  title     = "{LLM} Discussion: Enhancing the Creativity of Large Language
               Models via Discussion Framework and Role-Play",
  author    = "Lu, Li-Chun and Chen, Shou-Jen and Pai, Tsung-Min and Yu,
               Chan-Hung and Lee, Hung-Yi and Sun, Shao-Hua",
  booktitle = "First Conference on Language Modeling",
  month     =  aug,
  year      =  2024
}

@ARTICLE{Peeperkorn2024-ig,
  title    = "Is Temperature the Creativity Parameter of Large Language Models?",
  author   = "Peeperkorn, Max and Kouwenhoven, Tom and Brown, Daniel and
              Jordanous, Anna",
  journal  = "CoRR",
  month    =  jan,
  year     =  2024
}

@INPROCEEDINGS{Chakrabarty2024-lj,
  title     = "Creativity support in the age of large language models: An
               empirical study involving professional writers",
  author    = "Chakrabarty, Tuhin and Padmakumar, Vishakh and Brahman, Faeze and
               Muresan, Smaranda",
  booktitle = "Creativity and Cognition",
  publisher = "ACM",
  address   = "New York, NY, USA",
  month     =  jun,
  year      =  2024,
  language  = "en"
}

@INPROCEEDINGS{Padmakumar2023-ry,
  title     = "Does Writing with Language Models Reduce Content Diversity?",
  author    = "Padmakumar, Vishakh and He, He",
  booktitle = "The Twelfth International Conference on Learning Representations",
  month     =  oct,
  year      =  2023
}

@INPROCEEDINGS{McLaughlin2024-jb,
  title     = "{AidanBench}: Evaluating Novel Idea Generation on Open-Ended
               Questions",
  author    = "McLaughlin, Aidan and Uppuluri, Anuja and Campbell, James",
  booktitle = "Language Gamification - NeurIPS 2024 Workshop",
  month     =  dec,
  year      =  2024
}

@INPROCEEDINGS{He2019-mq,
  title     = "Pun Generation with Surprise",
  author    = "He, He and Peng, Nanyun and Liang, Percy",
  editor    = "Burstein, Jill and Doran, Christy and Solorio, Thamar",
  booktitle = "Proceedings of the 2019 Conference of the North {A}merican
               Chapter of the Association for Computational Linguistics: Human
               Language Technologies, Volume 1 (Long and Short Papers)",
  publisher = "Association for Computational Linguistics",
  address   = "Minneapolis, Minnesota",
  pages     = "1734--1744",
  month     =  jun,
  year      =  2019
}

@ARTICLE{Lu2025-sn,
  title         = "Rethinking creativity evaluation: A critical analysis of
                   existing creativity evaluations",
  author        = "Lu, Li-Chun and Liu, Miri and Lu, Pin-Chun and Tian, Yufei
                   and Sun, Shao-Hua and Peng, Nanyun",
  journal       = "arXiv [cs.CL]",
  month         =  aug,
  year          =  2025,
  archivePrefix = "arXiv",
  primaryClass  = "cs.CL"
}

@ARTICLE{He2025-ij,
  title         = "What shapes a creative machine mind? Comprehensively
                   benchmarking creativity in foundation models",
  author        = "He, Zicong and Zhang, Boxuan and Liu, Weihao and Tang,
                   Ruixiang and Cheng, Lu",
  journal       = "arXiv [cs.AI]",
  month         =  oct,
  year          =  2025,
  archivePrefix = "arXiv",
  primaryClass  = "cs.AI"
}

@ARTICLE{Fang2025-im,
  title         = "Creation-{MMBench}: Assessing context-aware creative
                   intelligence in {MLLM}",
  author        = "Fang, Xinyu and Chen, Zhijian and Lan, Kai and Ma, Lixin and
                   Ding, Shengyuan and Liang, Yingji and Zhao, Xiangyu and Wen,
                   Farong and Zhang, Zicheng and Zhang, Guofeng and Duan,
                   Haodong and Chen, Kai and Lin, Dahua",
  journal       = "arXiv [cs.CV]",
  month         =  mar,
  year          =  2025,
  archivePrefix = "arXiv",
  primaryClass  = "cs.CV"
}

@TECHREPORT{ttct-current,
  title     = "Torrance tests of creative thinking interpretive manual",
  author    = "Torrance, E Paul",
  publisher = "Scholastic Testing Service",
  address   = "Bensenville",
  language  = "en",
  url = "https://www.ststesting.com/gift/TTCT_InterpMOD.2018.pdf"
}

@article{torrance1974torrance,
  title={Torrance tests of creative thinking},
  author={Torrance, E Paul},
  journal={Educational and psychological measurement},
  year={1974}
}

@MISC{ttct-training,
  title        = "Torrance Tests of Creative Thinking ({TTCT}) Training - Mary
                  Frances Early College of Education",
  howpublished = "\url{https://coe.uga.edu/outreach/programs/ttct/}",
  note         = "Accessed: 2026-4-1",
  language     = "en"
}

@article{runco2012standard,
  title={The standard definition of creativity},
  author={Runco, Mark A and Jaeger, Garrett J},
  journal={Creativity research journal},
  volume={24},
  number={1},
  pages={92--96},
  year={2012},
  publisher={Taylor \& Francis}
}

@ARTICLE{Verga2024-us,
  title         = "Replacing judges with juries: Evaluating {LLM} generations
                   with a Panel of diverse models",
  author        = "Verga, Pat and Hofstatter, Sebastian and Althammer, Sophia
                   and Su, Yixuan and Piktus, Aleksandra and Arkhangorodsky,
                   Arkady and Xu, Minjie and White, Naomi and Lewis, Patrick",
  journal       = "arXiv [cs.CL]",
  month         =  apr,
  year          =  2024,
  archivePrefix = "arXiv",
  primaryClass  = "cs.CL"
}

@ARTICLE{Li2025-nv,
  title         = "Jointly reinforcing diversity and quality in language model
                   generations",
  author        = "Li, Tianjian and Zhang, Yiming and Yu, Ping and Saha,
                   Swarnadeep and Khashabi, Daniel and Weston, Jason and
                   Lanchantin, Jack and Wang, Tianlu",
  journal       = "arXiv [cs.CL]",
  month         =  sep,
  year          =  2025,
  archivePrefix = "arXiv",
  primaryClass  = "cs.CL"
}
